\documentclass[12pt]{article} 
\usepackage{amsmath,amsthm,amssymb,bm,mathrsfs,}
\usepackage{graphicx,subfig}
\usepackage[linesnumbered, ruled, lined]{algorithm2e}
\usepackage{caption,enumerate,natbib,listings,setspace}
\usepackage[OT1]{fontenc}
\usepackage[colorlinks,linkcolor=red,anchorcolor=blue,citecolor=blue,hypertexnames=false]{hyperref}
\usepackage[margin=1in]{geometry}
\usepackage[protrusion=false,expansion=true]{microtype}
\usepackage[title]{appendix}
\usepackage{bbm}
\usepackage{comment}
\theoremstyle{plain}
\usepackage{smile}
\usepackage{xr}
\usepackage{color}

\title{\Large{\textbf{Pessimistic Causal Reinforcement Learning with Mediators for Confounded Offline Data}} }

\author
{
Danyang Wang\thanks{Department of Statistics, Purdue University, Email: wang4589@purdue.edu},
Chengchun Shi\thanks{Department of Statistics, London School of Economics and Political Science, Email: c.shi7@lse.ac.uk},
Shikai Luo\thanks{ByteDance, Email: shadow.luo@bytedance.com},
Will Wei Sun\thanks{Daniels School of Business, Purdue University. Email: sun244@purdue.edu. Corresponding author.}
}

\date{}
\begin{document} 

\maketitle

\begin{abstract}
\noindent
In real-world scenarios, datasets collected from randomized experiments are often constrained by size, due to limitations in time and budget. As a result, leveraging large observational datasets becomes a more attractive option for achieving high-quality policy learning. However, most existing offline reinforcement learning (RL) methods depend on two key assumptions—unconfoundedness and positivity—which frequently do not hold in observational data contexts. Recognizing these challenges, we propose a novel policy learning algorithm, PESsimistic CAusal Learning (PESCAL). We utilize the mediator variable based on front-door criterion to remove the confounding bias; additionally, we adopt the pessimistic principle to address the distributional shift between the action distributions induced by candidate policies, and the behavior policy that generates the observational data. Our key observation is that, by incorporating auxiliary variables that mediate the effect of actions on system dynamics, it is sufficient to learn a lower bound of the mediator distribution function, instead of the Q-function, to partially mitigate the issue of distributional shift. This insight significantly simplifies our algorithm, by circumventing the challenging task of sequential uncertainty quantification for the estimated Q-function. Moreover, we provide theoretical guarantees for the algorithms we propose, and demonstrate their efficacy through simulations, as well as real-world experiments utilizing offline datasets from a leading ride-hailing platform.
\end{abstract}


\newpage
\baselineskip=24pt

\section{Introduction}
\label{sec:introduction}
In reinforcement learning (RL), an agent interacts with the environment at each time step to obtain a reward, and continually updates its policy to maximize the expected total (discounted in infinite horizon case) reward \citep{sutton2018reinforcement}.
The adoption of RL in solving real-world problems has been largely benefited from the advancements in deep learning (e.g., \cite{lecun2015deep}) over the past decade. Since 2013, deep RL has achieved remarkable success in fields including video games, natural language processing and game playing, where it is cheap to generate numerous data for policy learning (e.g., \cite{mnih2013playing, mnih2015human, silver2016mastering, berner2019dota, openai2023gpt4}). 
However, in many other applications such as finance, healthcare, ride-sharing, video-sharing and autonomous driving, it is often unrealistic, infeasible, or unethical to continuously interact with the environment to update the optimal policy in real time. Meanwhile, datasets collected from randomized experiments are often
limited in size due to time or budget constraints. This motivates us to use large observational offline datasets to conduct high-quality policy learning \citep{levine2020offline,chang2021mitigating,shi2022pessimistic,qi2022offline,kallus2022stateful}. 

\begin{figure}
\centering
\subfloat[Standard MDP]{\includegraphics[width=0.47\linewidth]{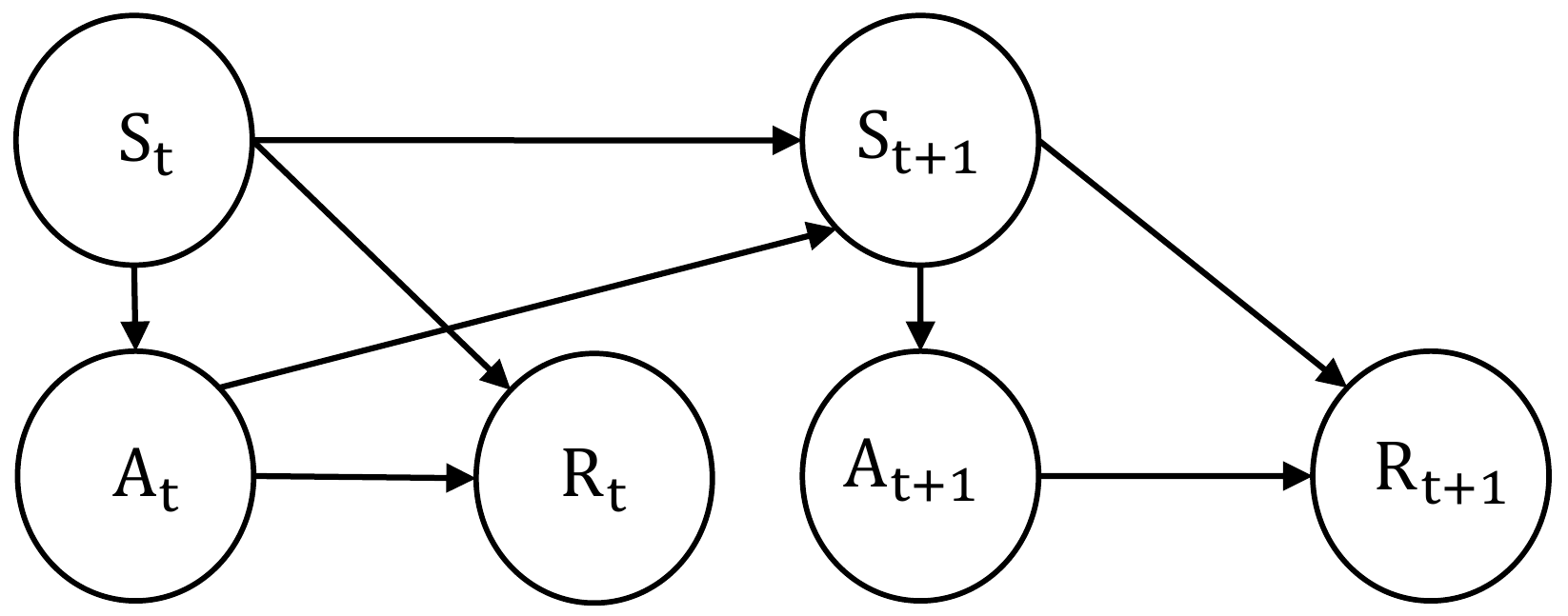}\label{img: online_mdp}}
\subfloat[MDP with unobserved confounder]{\includegraphics[width=0.53\linewidth]{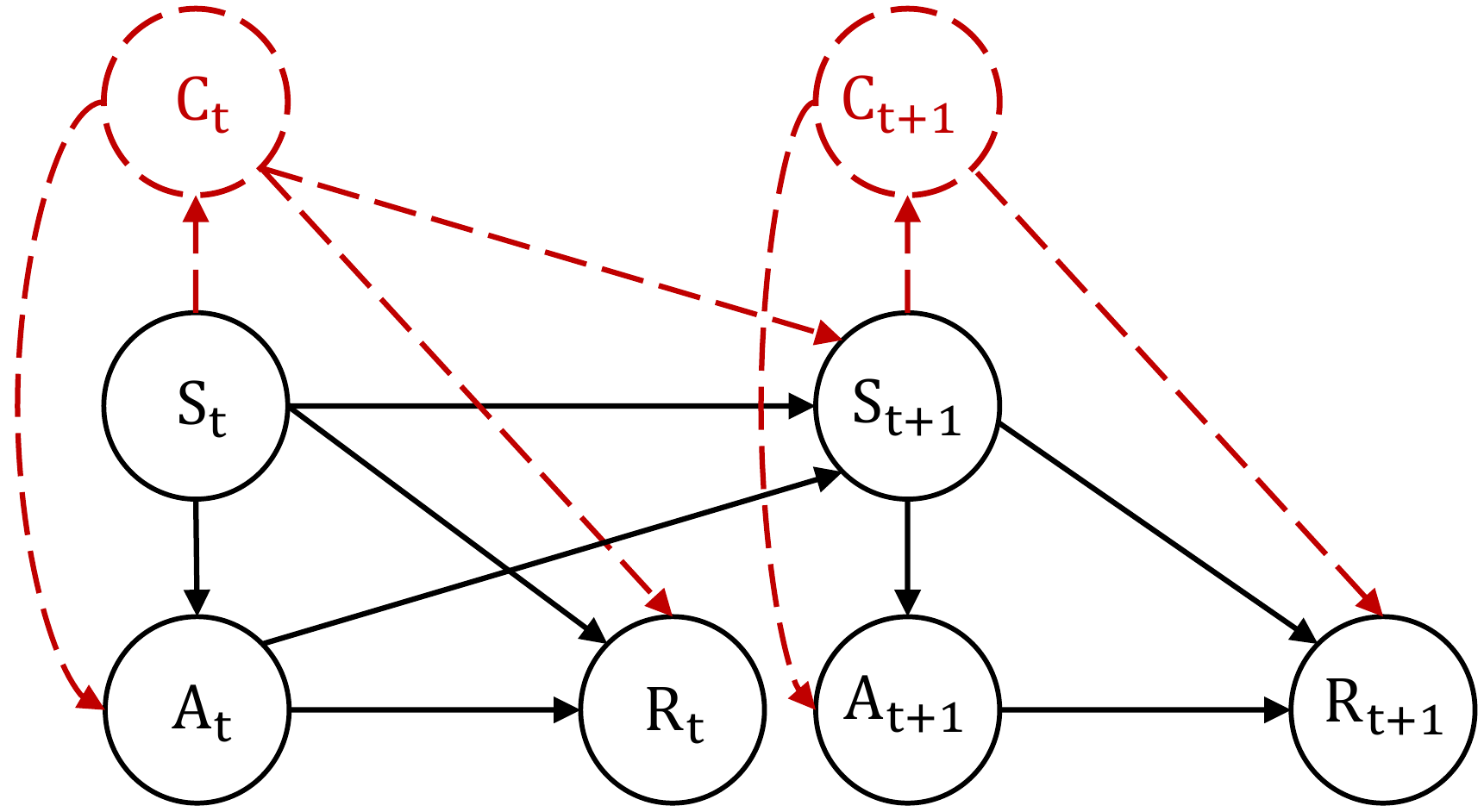}\label{img: cfd_mdp}}
\caption{Causal relationships in Markov Decision Process. (a) standard MDP; (b) MDP with unobserved confounder. $S, A, R \text{ and } C$ represent state, action, reward and confounder, respectively. Arrows denote causal relationship. Solid lines indicate observed variables (or relationships), while dotted lines indicate unobserved variables (or relationships).}
\label{img: online}
\end{figure}

Most existing state-of-the-art offline policy learning algorithms rely on the following two crucial assumptions: 1) unconfoundedness: no unmeasured variables exist that confound the action-reward or action-next-state associations (see Figure \ref{img: online_mdp}); 2) positivity (also known as the full coverage assumption in the machine learning literature): all the actions have been sufficiently explored in the offline dataset conditioning on the states \citep[e.g.,][]{yan2022model}. Both assumptions are likely to be violated in a variety of applications. On one hand, data collected from robotics, healthcare and ride-sharing may not contain all relevant confounding variables \citep[see e.g.,][]{siciliano2008springer,kleinberg2018human,namkoong2020off,shi2022off}. As an example, doctors may prescribe a specific drug to a patient based on communications or visual observations which are not fully recorded. As another example, a robotic arm may slow down its operation based on a perceived obstruction, e.g., an unexpected object on the conveyor belt, which is not detected by the sensor. Figure \ref{img: cfd_mdp} describes the causal relationship in a Markov Decision Process (MDP) with unobserved confounder.
On the other hand, distributional shift is commonly seen in offline observational data \citep{levine2020offline, jin2021pessimism, shi2022pessimistic}. In particular, when the behavior policy is close to deterministic, certain state-action pairs are explored less frequently in the offline data, which makes the data distribution inadequate to cover the state-action distribution induced by some candidate policy, invalidating the positivity assumption.

A motivating example of our work is from the application of learning an optimal passenger-side subsidizing policy in large-scale ride-sharing platforms, such as Uber, Lyft and Didi. When a passenger opens a ride-sharing application on their smartphone and enters their destination, the platform needs to decide whether to send a specific coupon to subsidize this ride. Sending the coupon not only increases the likelihood that the passenger books this particular ride, but also encourages them to use the app more frequently in the future. Yet, the challenge of confoundedness and distributional shift presents a substantial obstacle to optimal policy learning. These issues arise because the behavior policy is a mixture of two types of policies: estimated automated policy and human interventions. The estimated automated policy is designed to optimize the company's rewards, following a deterministic approach that prescribes a singular action for each specific state. However, this deterministic nature presents a limitation -- it leads to under-coverage in the collected dataset, violating the positivity assumption. Meanwhile, human interventions are aimed at better balancing demand and supply in areas with extreme weather, major sports events, or unexpected public transportation strikes. These interventions are necessary to mitigate the effects of unexpected disruptions that are often not recorded in the system, resulting in a confounded dataset.

\subsection{Contributions}
We summarize our contributions as follows. First, offline policy learning without additional online data collection is essential in real-world sequential decision making. Nonetheless, most existing works require either positivity or confoundedness (see Section \ref{sec:relatedwork} for a review of the literature), yielding sub-optimal policies when \textbf{both} assumptions are simultaneously violated.  To our knowledge, this is the first paper to study offline RL in confounded mediated Markov decision processes (M2DP, see Figure \ref{img: m2dp} for details) in the presence of both unmeasured confounding and distributional shift. Second, we propose an original pessimistic causal learning algorithm to maximize the expected cumulative reward over time. In particular, our algorithm properly utilizes the auxiliary mediator variable in M2DP that mediates the effect of actions on system dynamics to remove the confounding bias, and adopts the pessimistic principle \citep{jin2021pessimism} to learn a ``conservative" policy which avoids taking actions that are less explored in the observational data. Existing works adopted pessimism by sequentially learning a lower bound of the Q-function \citep[see e.g.,][]{jin2021pessimism,zhou2023optimizing} at each iteration. Our proposal is not a simple combination of existing causal RL and pessimistic RL algorithms. A key innovation of our proposal is to learn a lower bound of the distribution function of mediator variable instead. This partially alleviates the issue of distributional shift while avoiding the task of sequential uncertainty quantification for the estimated Q-function, which is challenging as the Q-function itself is parameter-dependent at each iteration (see Section \ref{sec:reviewpessimism} for details). Additionally, we show that pointwise uncertainty quantification for the mediator distribution function is adequate to ensure the resulting policy's consistency, eliminating the need for the demanding task of uniformly quantifying uncertainty across the state space. Finally, we develop a theoretical framework for policy learning in M2DP by introducing a mediated optimal Q-function, establishing the Bellman optimality equation for this Q-function and obtaining the closed-form expression of the optimal policy based on this function. In addition, we establish the regret bound of the proposed optimal policy without assuming unconfoundedness and positivity.

\subsection{Related Works}\label{sec:relatedwork}
There is a vast body of literature on RL within the field of computer science \citep[see e.g.,][]{sutton2018reinforcement,levine2020offline}. In statistics, RL is closely related to a growing literature on estimating dynamic treatment regimes (DTRs) in precision medicine \citep{DTRs,kosorok2019precision,tsiatis2019dynamic}; as well as treatment in general \citep{Murphy,gest,qian2011performance,zhang2013robust,zhao2015new,ertefaie2018constructing,shi2018high,wang2018quantile,zhang2018interpretable,luckett2019estimating,qi2020multi,mo2021learning,nie2021quasi,zhou2022estimating}. Nonetheless, most works in these fields require unconfoundedness and positivity. 

There is a growing interest in developing causal policy learning algorithms that allow unmeasured confounding \citep{lu2018deconfounding,kallus2018confounding,chen2021estimating,liao2021instrumental,li2021causal,wang2021provably,tan2022offline,wang2022blessing,bruns2023robust} and pessimistic policy learning algorithms without the positivity assumption \citep{kumar2020conservative,yu2020mopo,chang2021mitigating,jin2021pessimism,rashidinejad2021bridging,xie2021bellman,jin2022policy,shi2022pessimistic,yan2022efficacy,ueharapessimistic,zhou2023optimizing}. However, these papers have only addressed either confoundedness or distributional shift, which could lead to suboptimal policies when both issues coexist.

There are a few recent proposals for offline policy learning without unconfoundedness and positivity. In particular, \citet{fu2022offline} developed an instrumental variable (IV) approach for value identification and employed minimax optimization to leverage pessimism for policy optimization. However, since IV is not available in our ride-sharing application, this method is not applicable in our setting. \citet{lu2022pessimism} developed a P3O algorithm by integrating proximal RL designed for value identification in confounded partially observable Markov decision processes \citep[see e.g.,][]{tennenholtz2020off} and pessimistic policy learning. However, proximal RL is not applicable in our application where the behavior policy depends on both the observed and unobserved confounders. 

\subsection{Organization of the Paper}
In the following, we first introduce the confounded mediated MDP (M2DP), our problem setting, and previous uncertain quantification method in Section \ref{sec:preliminary}. In Section \ref{sec:alg&thm} we derive the optimal policy based on confounded M2DP, and introduce two frameworks of algorithms, PESCAL and CAL, with and without the integration of pessimism, respectively. We further study the convergence of the algorithms in terms of the regret in Section \ref{sec:theory}. Finally, in Section \ref{sec:simu_exper} and Section \ref{sec:real_exper}, we assess the effectiveness of PESCAL and CAL in comparison to state-of-the-art offline RL algorithm using both simulated datasets and real-world datasets.

\section{Preliminaries}\label{sec:preliminary}
We first introduce the confounded M2DP, a special Markov decision process (MDP) with unmeasured confounders \citep[MDPUC,][]{zhang2016markov}, featuring the presence of a certain auxiliary mediator variable that mediates the effect of actions on the system dynamics. We next introduce the pessimistic principle, since it is closely related to our proposal, and discuss the challenges of developing pessimistic-type RL algorithms.
\subsection{Confounded MDP with Mediator}
\label{sec:model}
We denote the M2DP as $\mdp$, represented by the tuple $(\cS, \cC, \cA, \cM, p_s, p_c, p_m, p_r, \gamma, \rho_0)$, where the first four elements $\cS$, $\cC$, $\cA$, and $\cM$ denote the spaces of state, unmeasured confounders, action, and mediator, respectively. We focus on the setting where both $\cA$ and $\cM$ are discrete, finite spaces, whereas $\cS$ and $\cC$ are allowed to be either discrete or continuous. The next four elements represent the conditional probability mass (or density) distributions of the state, unmeasured confounders, mediator and reward; see the details below. Finally, $\gamma \in [0,1)$ denotes the discount factor that balances the trade-off between the immediate and future rewards whereas $\rho_0$ denotes the initial state distribution. Throughout the paper, we use capital letters to represent  random variables and lowercase letters to represent their realizations.

\begin{figure}[htb]
\centering
\subfloat[Offline Confounded M2DP]{\includegraphics[width=0.52\linewidth]{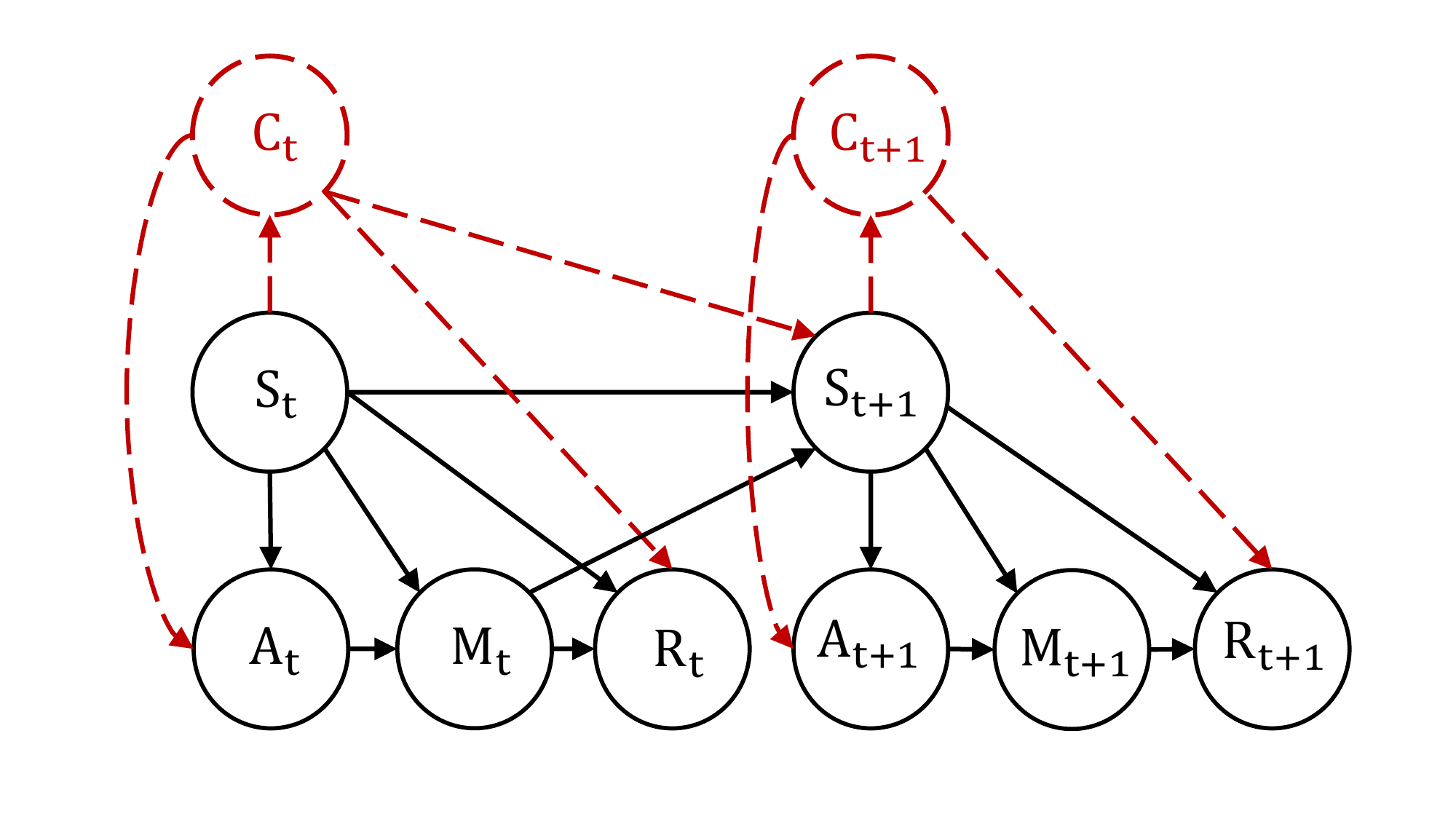}\label{img: medi}}
\subfloat[Online Unconfounded M2DP]{\includegraphics[width=0.5\linewidth]{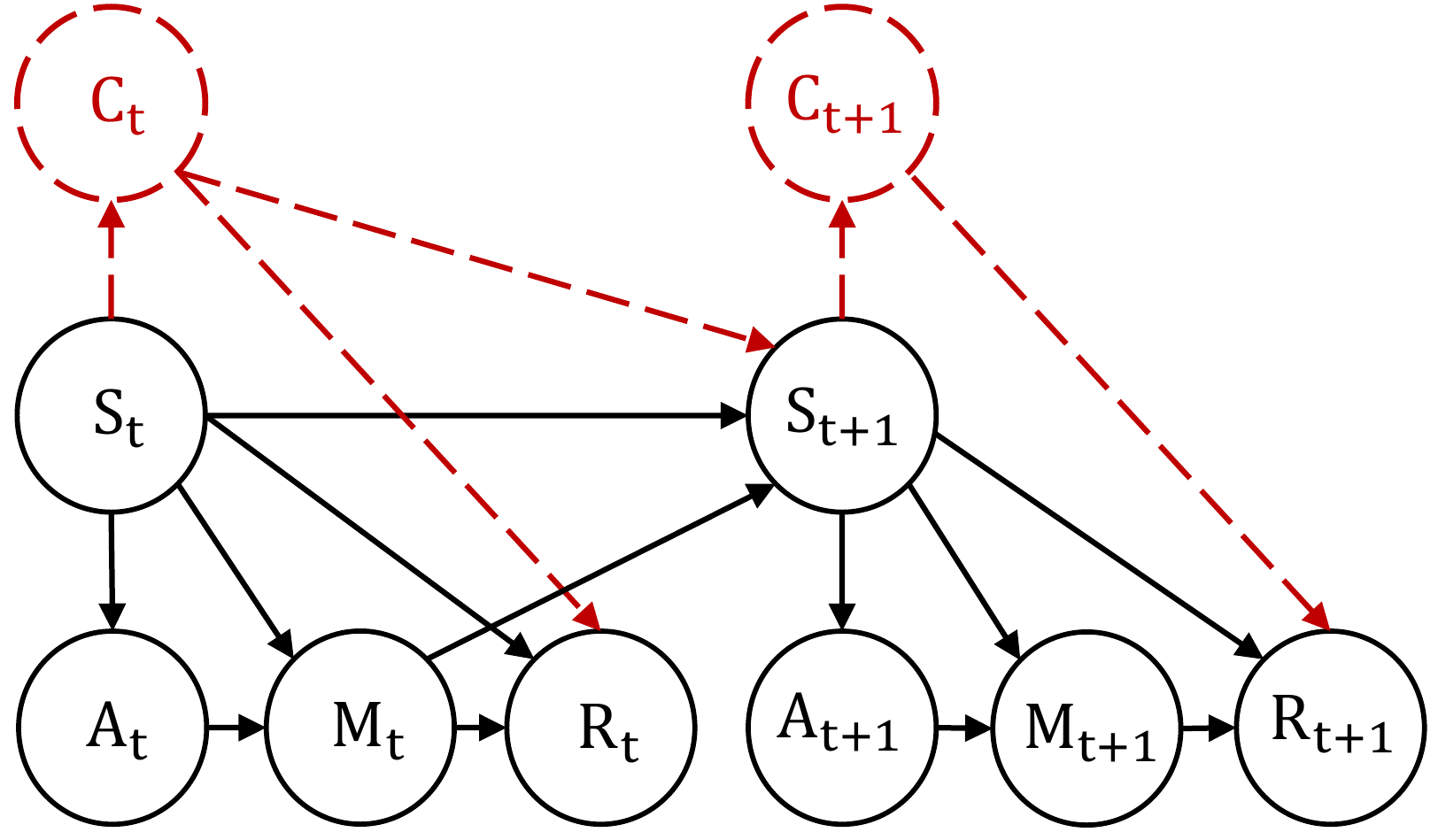}\label{img:online_cmdp}}
\caption{Mediated Markov Decision Processes (M2DPs): (a) We assume that the offline dataset generating process follows an offline confounded M2DP, as illustrated in the diagram; (b) Online unconfounded M2DP. Again, dotted lines indicate that the relationship (or variable) is not directly observable. We aim to learn an optimal policy $\pi^*$ that can be executed online without the confounding effect, as shown in (b), while training on a static dataset $\cD$ generated by the process described in (a).}
\label{img: m2dp}
\end{figure}

The data are generated as follows. At the initial time, the environment arrives at a state $S_0\sim \rho_0(\bullet)$ with some latent confounders $C_0\sim p_c(\bullet|S_0)$. Next, the agent chooses an action $A_0$ according to certain behavior policy $p_a(\bullet|S_0,C_0)=\mathbb{P}(A_0=\bullet|S_0,C_0)$ that depends on both $S_0$ and $C_0$. The environment then generates a mediator $M_0\sim p_m(\bullet|S_0,C_0,A_0)$, provides an immediate reward $R_0\sim p_r(\bullet|S_0,C_0,A_0,M_0)$ to the agent and arrives at a new state $S_1\sim p_s(\bullet|S_0,C_0,A_0,M_0)$\footnote{We implicitly assume that the reward and future state are conditionally independent given other variables. However, this condition is unnecessary since the optimal policy depends on the data generating process only through $p_s$, $p_c$, $p_b$, $p_m$, $p_r$, and remains the same regardless of whether the reward and next state are conditionally independent or not.}. This procedure repeats until termination. According to the data generating process, $C_t$ is indeed an unobserved confounder that influences both the action, and the reward and next state. This data generating process essentially requires the data trajectory to follow a Markov chain. It implies the so-called ``memoryless unmeasured confounding" assumption which requires $C_t$ to be independent of $\{C_j\}_{j<t}$ and past observed variables given $S_t$. This memoryless assumption has been frequently imposed in the RL literature to simplify the analysis \citep[see e.g.,][]{kallus2020confounding,liao2021instrumental,wang2021provably,fu2022offline,xu2022instrumental}. Besides, the M2DP will automatically assign a value $p_{m^-}$ to small $p_m(m|s,c,a)$, when $p_m(m|s,c,a)<p_{m^-}$, where $p_{m^-}$ is a threshold that is small enough, e.g., $1\times10^{-5}$. Following the standard assumption in MDP \citep{sutton2018reinforcement,yang2020function}, we assume the reward is bounded, specifically, there exists some constant $R_{\max}<\infty$ such that $\max_t|R_t|\le R_{\max}$ almost surely. In addition, we assume the mediators in $\mdp$ satisfy the following assumption. See Figure \ref{img: medi} for a graphical visualization of the data generating process under Assumption \ref{asm:frontdoor_crit}. 

\begin{assumption}[Front-door Criterion \citep{pearl2009causality}]
\label{asm:frontdoor_crit}
A set of variables $M_t$ is said to satisfy the frond-door criterion if
(i) $M_t$ intercepts every directed path from $A_t$ to $S_{t+1}$ or $R_t$;
(ii) Conditioning on $S_t$, no path between $A_t$ and $M_t$ has an incoming arrow into $A_t$;
(iii) conditioning on $S_t$, $A_t$ blocks every path between $M_t$ and $S_{t+1}$ or $R_t$ that has an incoming arrow into $M_t$.
\end{assumption}

As depicted in Figure \ref{img: medi}, Assumption \ref{asm:frontdoor_crit}(i) assumes that the mediator fully reflects the influence of the action (blocking all outgoing arrows from the action to the reward and next state). In other words, there is no direct effect from the action to the reward or the next state. Assumption \ref{asm:frontdoor_crit}(ii) and (iii) preclude the existence of any incoming arrow from the mediator to the action, or from the immediate reward (or next state) to the mediator. It guarantees that the mediator is independent of unobserved confounders. As such, $p_m(M_t|S_t,C_t,A_t)$ can be simplified to  $p_m(M_t|S_t,A_t)$. These assumptions are applicable in our ride-sharing application. In particular, in the subsidizing example, the final discount applied to each ride corresponds to the mediator, which is based on whether the coupon is being sent or not (action) and other subsidizing policies. As such, the mediator is independent of unobserved confounders (e.g., extreme weather, transportation strikes). While customers can observe the final discount applied to their ride on the application, they are usually not aware of which subsidizing strategy resulted in the discount. Therefore, it is reasonable to assume that the policy will only affect passenger behavior through the mediator.

Under Assumption \ref{asm:frontdoor_crit}, an \textbf{online} policy $\pi=\{\pi_t\}_{t\ge0}$ is a sequence of conditional probability mass functions on $\mathcal{A}$. Following policy $\pi$, the distribution of the resulting data trajectory is:
\begin{eqnarray}\label{eqn:counterdistribution}
\begin{split}
    S_0\sim \rho_0(\bullet), \quad C_0\sim p_c(\bullet|S_0), \quad A_0\sim \pi_0(\bullet|S_0), \quad M_0\sim p_m(\bullet|S_0,A_0), \\  R_0\sim p_r(\bullet|S_0,C_0,A_0,M_0),\quad S_1\sim p_s(\bullet|S_0,C_0,A_0,M_0),\quad \cdots
\end{split}    
\end{eqnarray}
See Figure \ref{img:online_cmdp} for a graphical representation of \eqref{eqn:counterdistribution}. 

In order to define the expected return and the optimal policy, we next introduce intervention, outlined by \citet{pearl2009causality}. A deterministic intervention on the action $A_t$ assigns certain value $a$ to $A_t$, regardless of the influence from other variables (e.g., $S_t$ \& $C_t$). We represent the deterministic intervention on $A$ as $do(A = a)$ and denote it as $do(a)$. A stochastic intervention assigns a stochastic distribution $\pi_t$ (which is dependent only on $S_t$) to $A_t$ such that the distribution of $A_t$ satisfies $\mathbb{P}(A_t=a)=\pi(a|S_t)$ and is independent of $C_t$, where we denote it as $do(X\sim P)$. 

Analogues to value functions in standard MDP, we define value functions in M2DP. Without loss of generality, assume infinite horizon for now. State value function $V^\pi(s)$ is the expected return starting from a given state $s$, and acting following online policy $\pi$ onwards:
\(V^\pi(s)=\underset{t\geq0}{\sum}\gamma^t\E^\pi(R_t|S_0=s, do(A_{0:\infty} \sim \pi))\); State-action value function $q^\pi(s,a)$ is the expected return when we start in state $s$, take action $a$, and then act according to online policy $\pi$: $q^\pi(s,a)=\underset{t\geq0}{\sum}\gamma^t\E^\pi(R_t|S_0=s,A_0=a,do(A_{1:\infty} \sim \pi))$; In addition, we also define the mediated state-action value function (\cite{shi2022off}, also referred to as the mediated Q-function), $Q^\pi(s,a,m)$, which is the expected return when we start in state $s$, take action $a$, observe mediator $m$, and then act according to online policy $\pi$: $$Q^\pi(s,a,m)=\underset{t\geq0}{\sum}\gamma^t\E^\pi(R_t|S_0=s,A_0=a,M_0=m, do(A_{1:\infty} \sim \pi)).$$
Note that all the value functions are defined online without effect from unobserved confounder, as indicated by interventions. Based on boundedness of $R_t$, by definition, $V^\pi, q^\pi, Q^\pi\le V_{\max}:=\frac{R_{\max}}{1-\gamma}$, $\forall \pi, s, a, m$. Throughout this paper, we also assume $\sup_{(s,a,m)} |\widehat{Q}(s,a,m)|\le V_{\max}$ for any estimation $\widehat{Q}$ of $Q$, which is a standard assumption in RL literature \citep[see e.g.,][]{sutton2018reinforcement, liao2021instrumental,fu2022offline}. The resulting expected discounted cumulative return is given by
\begin{eqnarray*}
    J(\pi)=\sum_{t\ge 0} \gamma^t \E^{\pi}(R_t).
\end{eqnarray*}
In definitions above, $\E^{\pi}$ denotes the expectation assuming the trajectory is generated following \eqref{eqn:counterdistribution}. 
Given an offline dataset $\cD$ with unobserved confounders, our goal is to learn an optimal policy $\pi^*$, which we can execute online without the effect of confounders, that maximizes $J(\pi)$.

Due to the presence of unmeasured confounders, accurately determining $J(\pi)$ and thus the optimal policy is typically unattainable without extra assumptions. To grasp this challenge, observe that $\E^{\pi} (R_t|S_t, do(A_t\sim \pi_t))$ generally does not equate to $\E(R_t|S_t, A_t\sim \pi_t)$. This distinction arises due to the expectations being computed based on distinct distribution functions. Specifically, the former is taken with respect to the intervention distribution under an unconfounded M2DP where $C_t$ does not affect $A_t$ (see Figure \ref{img:online_cmdp}) whereas the latter is taken with respect to the offline data distribution generated by a confounded M2DP where $C_t$ has a direct impact on $A_t$ (see Figure \ref{img: medi}). 

We incorporate the front-door adjustment criterion to address this challenge. Under Assumption \ref{asm:frontdoor_crit}, it follows from the front-door adjustment formula that
\begin{eqnarray}\label{eqn:frontdoor}
    \E^{\pi}(R_t|S_t)=\sum_{a,\tilde{a},m} p_m(m|S_t,a) \pi_t(a|S_t)p_b(\tilde{a}|S_t) \E(R_t|M_t=m,A_t=\tilde{a},S_t),
\end{eqnarray}
where $p_b(a|s)=\mathbb{P}(A_t=a|S_t=s)$ denotes the conditional probability mass function used to collect the offline dataset. Notice that the right-hand-side does not involve the unobserved $C_t$ and depends on the observed data distribution only. This allows us to consistently identify the 
interventional distribution. We adopt this strategy to identify the optimal policy in Section \ref{sec:optimalpolicy}. 

To conclude this section, we remark that although Assumption \ref{asm:frontdoor_crit} might appear to be strong, the front-door adjustment criteria, as discussed in \citep{pearl2009causality, imbens2015causal, fulcher2020robust}, is commonly adopted in the RL literature to study the effects of unobserved confounders \citep{liao2021instrumental, kallus2020confounding, fu2022offline, wang2021provably, xu2022instrumental}. Meanwhile, when Assumption \ref{asm:frontdoor_crit} is violated, our proposed policy will asymptotically maximize the long-term natural indirect effect \citep[the effect on the cumulative reward through the mediator; see e.g.,][]{fulcher2020robust}, instead of the sum of both the direct and indirect effects. Besides, we observe that fulfilling the memoryless assumption is not a prerequisite for ensuring the large sample properties of our algorithm. In particular, the ``memoryless unmeasured confounding'' assumption can be relaxed by adopting a ``high-order memoryless unmeasured confounding"
assumption or by applying certain completeness assumptions \citep{shi2022off}, allowing the unobserved confounder to depend on a specific length of past transitions.

\subsection{The Pessimistic Principle and Uncertainty Quantification}\label{sec:reviewpessimism}
\begin{figure}[h!]
    \centering
    \includegraphics[width=0.5\linewidth]{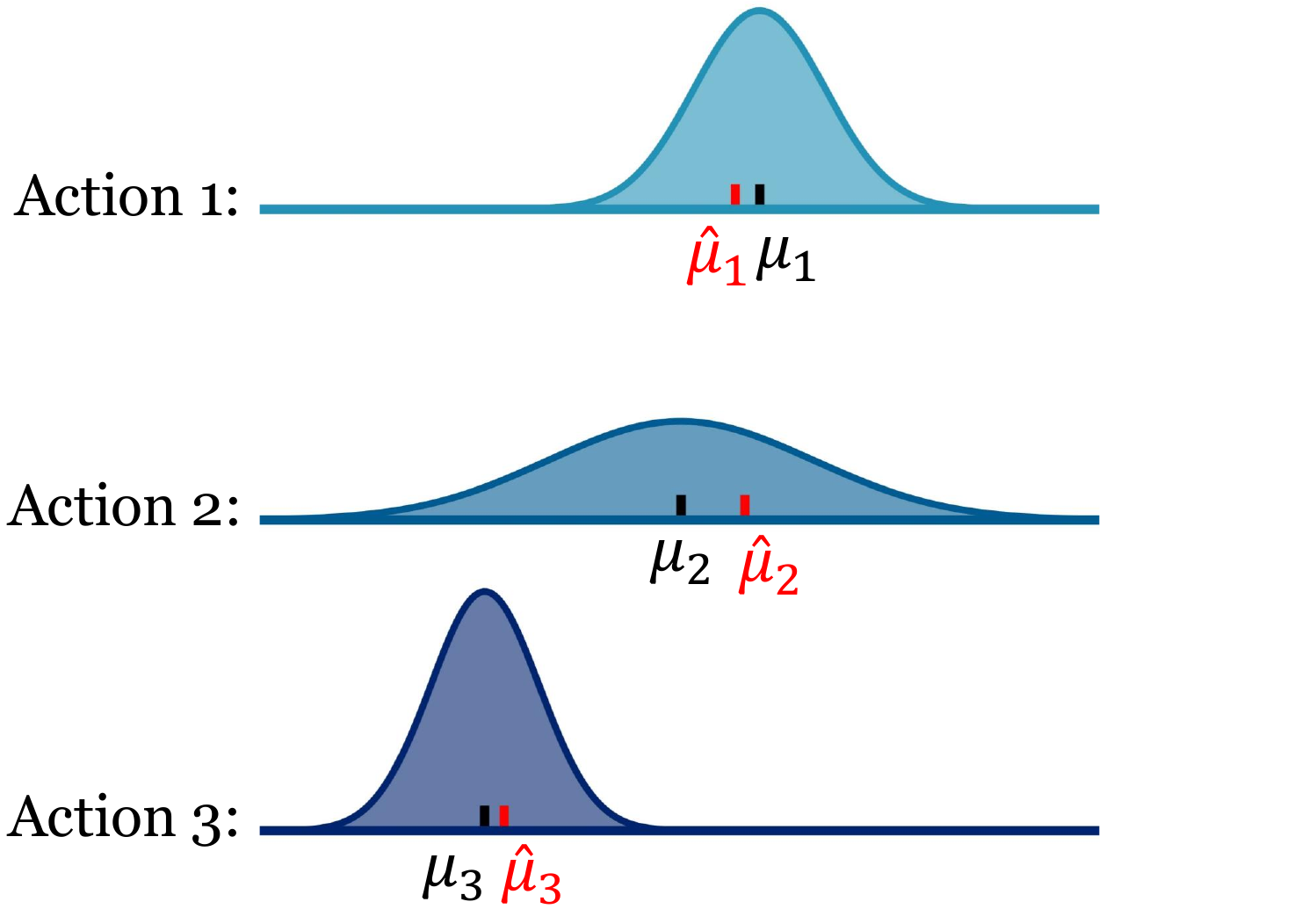}
    \caption{Example distributions of estimated expected reward for policy learning in a three-armed bandit. The oracle expected rewards for the three arms $a_1$, $a_2$ and $a_3$ are given by $\mu_1>\mu_2>\mu_3$. However, due to the limited sample size for the second arm, the estimated expected reward $\hat{\mu}_2$, is subject to a significant variance. Consequently, $\hat{\mu}_2>\hat{\mu}_1>\hat{\mu}_3$ occurs with a high probability, as illustrated in the graph.}
    \label{img:mab}
\end{figure}

As commented earlier, when the behavior policy is nearly deterministic, some state-action pairs might not be sufficiently explored in the offline data, making their Q-functions exceedingly difficult to learn. This can lead to high variances of the Q-function estimator and ultimately suboptimal decision-making; see Figure \ref{img:mab} for an illustration in an offline multi-armed bandit setting (a special case of MDP without state information). 

To elaborate, suppose the offline data is generated from an unconfounded MDP, consisting of state-action-reward triplets over time. Existing Q-learning type algorithms learn an optimal Q-function, e.g., the expected discounted cumulative return starting from an initial state-action pair, and compute the estimated optimal policy as the greedy policy with respect to the estimated Q-function. However, the consistency of this policy typically requires the data distribution to cover the state-action distribution induced by all possible policies \citep[see e.g.,][]{chen2019information}. 

Pessimistic Q-learning algorithms address this issue by sequentially penalizing the Q-function estimator and calculating the greedy policy with respect to this penalized estimator \citep[see e.g.,][]{jin2021pessimism,shi2022pessimistic,zhou2023optimizing}. In particular, at the $k$th iteration, these algorithms propose to learn a lower bound $\widehat{q}_L^{(k)}$ of the following Q-function, 
\begin{eqnarray*}
    q^{(k)}(s,a)=\E \left[R_t+\gamma \max_a \widehat{q}_L^{(k-1)}(S_{t+1},a)|S_t=s,A_t=a,\widehat{q}_L^{(k-1)}\right],
\end{eqnarray*}
where $\widehat{q}_L^{(k-1)}$ denotes the lower bound at the last iteration such that the event $\cap_{s,a} \{q^{(k)}(s,a)\ge \widehat{q}_L^{(k)}(s,a)\}$ holds with large probability. However, this is a challenging task since (i) $q^{(k)}$ itself is parameter-dependent and relies on the estimated lower bound at the last iteration; (ii) $\widehat{q}_L^{(k)}$ is required to lower bound $q^{(k)}$ uniformly over the state space. 

To address the first challenge, one can employ data splitting to use distinct data trajectories to learn the lower bound at different iterations. This guarantees that the Q-function target $\widehat{q}_L^{(k-1)}$ is independent of the data to be used to produce $\widehat{q}_L^{(k)}$ so that existing methods are applicable to quantify the uncertainty of an estimated $q^{(k)}$ (denoted by $\widehat{q}^{(k)}$) in order to derive $\widehat{q}_L^{(k)}$. However, this would result in a loss of efficiency due to the usage of only a fraction of data at each iteration. Alternatively, one may consider a loose bound by setting the difference $\widehat{q}^{(k)}-\widehat{q}^{(k)}_L$ to be much larger than the standard deviation of $\widehat{q}^{(k)}$. However, the regret of the resulting policy is typically proportional to this difference \citep{jin2021pessimism}. 

To tackle the second challenge, one can consider the simpler task of learning a pointwise lower bound. However, it remains unclear whether the resulting algorithm is theoretically guaranteed. In the next section, we will introduce a newly proposed policy design that naturally handles these challenges in the considered confounded M2DP problem.

\section{Pessimistic Causal Learning}
\label{sec:alg&thm}
We propose a theoretical framework for policy learning in a confounded M2DP. In particular, we derive the Bellman optimality equation, and establish its connection with the optimal policy in a confounded M2DP. We next present a pessimistic causal Q-learning algorithm for policy learning with unmeasured confounding and distributional shift. We refer to our proposal as \textbf{PESCAL}, short for  \textbf{PES}simistic \textbf{CA}usal \textbf{L}earning. 
\subsection{The Optimal Policy in A Confounded M2DP}\label{sec:optimalpolicy}
Recall that the optimal policy $\pi^*$ is the argmax of the expected return $J(\pi)$. Similar to \eqref{eqn:frontdoor}, we can show that 
\begin{eqnarray*}
J(\pi)=\sum_{a,\tilde{a},m} \E_{S_0\sim\rho_0} [p_m(m|S_0,a) \pi_0(a|S_0)p_b(\tilde{a}|S_0)Q^{\pi}(S_0,\tilde{a},m)],
\end{eqnarray*}
and can thus be consistently identified under Assumption \ref{asm:frontdoor_crit} \citep{shi2022off}. 
We define the mediated optimal Q-function by $Q^*(s,a,m)=\sup_{\pi} Q^{\pi}(s,a,m)$ where the supremum is taken over the class of all policies. 

We next introduce three important classes of policies: Markov policies, stationary policies and deterministic policies. A Markov policy $\pi=\{\pi_t\}_t$ is a type of policy where each conditional distribution $\pi$ depends on the history $H_t$ (all the variables observed up to time $t$) only through the current state vector $S_t$, e.g., $\pi_t(a|H_t)=\pi_t(a|S_t)$ almost surely, for any $a$ and $t$. In contrast, a stationary policy is a special case of a Markov policy where these conditional distributions are stationary over time, with $\pi_0=\pi_1=\cdots=\pi_t=\cdots$. With a slight abuse of notation, we will use $\pi$ to denote this stationary conditional distribution. A deterministic policy $\pi$ is such that each conditional distribution $\pi_t$ is degenerate, e.g., for each $s_t$, there exists some action $a$ such that $\pi_t(a|s_t)=1$. We will use $\pi_t(s_t)$ as a shorthand to denote this action in following text. Our theoretical findings are summarized as follows. 
\begin{theorem}[Existence and uniqueness of optimal policy]\label{thm1}
    Under Assumption \ref{asm:frontdoor_crit}, (i) there exists an unique optimal deterministic stationary policy $\pi^*$ whose $J(\pi^*)$ is no worse than any other policies; (ii) $\pi^*$ is greedy with respect to a weighted average of the $Q^{*}$. In particular,
    \begin{eqnarray}\label{eqn:optimalpolicy}
        \pi^*(s)=\argmax_{a} \left[\sum_{\tilde{a},m} p_m(m|s,a) p_b(\tilde{a}|s)Q^*(s,\tilde{a},m)\right];
    \end{eqnarray}
    (iii) $Q^*$ satisfies the following Bellman optimality equation: 
    \begin{eqnarray}\label{eqn:Belloptimal}
        Q^*(S_t,A_t,M_t)=\E \left[\left.R_t+\gamma \max_{a} \sum_{\tilde{a},m} p_m(m|S_{t+1},a) p_b(\tilde{a}|S_{t+1})Q^*(S_{t+1},\tilde{a},m)\right|S_t,A_t,M_t\right].
    \end{eqnarray}
\end{theorem}
The detailed proof is provided in Supplementary \ref{proof:val_iter}. We make a few remarks. First, it is well-known that the optimal policy is stationary in unconfounded MDPs \citep{puterman2014markov}. Theorem \ref{thm1}(i) extends this result to confounded M2DPs under memoryless unmeasured confounding. Second, Theorem \ref{thm1}(ii) and (iii) form the basis of our proposed Q-learning algorithms (see Section \ref{sec:alg}), which derive the optimal policy by learning the mediated optimal Q-function that solves the Bellman optimality equation.

\subsection{Pessimistic Causal Q-learning}\label{sec:alg}
We first introduce a \textbf{CA}usal \textbf{L}earning (CAL) algorithm which requires the positivity assumption and present its pseudocode in Algorithm \ref{alg:cal}. The proposed algorithm effectively integrates the fitted Q-iteration \citep{riedmiller2005neural} designed for unconfounded MDPs with the front-door adjustment to handle unmeasured confounding. Given an offline dataset $\cD$, we organize it into state-action-mediator-reward-next state tuples, denoted by $\{(S_{(i)},A_{(i)},M_{(i)},R_{(i)},S_{(i)}'):1\le i\le N\}$ where $N$ denotes the total number of observations (transition tuples). We begin by applying existing supervised learning algorithms to learn $p_m$ and $p_b$ from the observed data. Notice that both functions do not involve the unmeasured confounders and can thus be consistently estimated. Let $\widehat{p}_m$ and $\widehat{p}_b$ denote the resulting estimators. We next 
iteratively update the mediated optimal Q-function to solve the Bellman optimality equation \eqref{eqn:Belloptimal}. Specifically, we initialize $\widehat{Q}^{(0)}$ as a zero function. During the $k$th iteration, it computes $\widehat{Q}^{(k)}$ by solving the following supervised learning,
\begin{eqnarray*}
\begin{split}
    \widehat{Q}^{(k)}=\argmin_{Q} \sum_{i=1}^{N}\left\{R_{(i)}+\gamma \max_a \sum_{\tilde{a},m} \widehat{p}_m(m|S_{(i)}',a) \widehat{p}_b(\tilde{a}|S_{(i)}')\widehat{Q}^{(k-1)}(S_{(i)}',\tilde{a},m)-Q(S_{(i)},A_{(i)},M_{(i)}) \right\}^2, 
\end{split}    
\end{eqnarray*}
with $\{R_{(i)}+\gamma \max_a \sum_{\tilde{a},m} \widehat{p}_m(m|S_{(i)}',a) \widehat{p}_b(\tilde{a}|S_{(i)}')\widehat{Q}^{(k-1)}(S_{(i)}',\tilde{a},m)\}_{i}$ as the supervised learning targets and $\{(S_{(i)},A_{(i)},M_{(i)})\}_{i}$ as the predictors. For a sufficiently large iteration number $K$ (e.g., 20), the estimated optimal policy is derived from \eqref{eqn:optimalpolicy}, given by
\begin{eqnarray}\label{eqn:estoptimalpolicy}
    \widehat{\pi}^{(K)}(s)=\argmax_a \Big[ \underbrace{\sum_{\tilde{a},m}\widehat{p}_m(m|s,a)\widehat{p}_b(\tilde{a}|s)\widehat{Q}^{(K)}(s,\tilde{a},m)}_{\widehat{q}^{(K)}(s,a)} \Big]. 
\end{eqnarray}
Statistical property of this estimated policy is summarized in Theorem \ref{thm:converge}. 

\begin{algorithm}[t]
\caption{CAL Algorithm}
\label{alg:cal}
\DontPrintSemicolon
\SetKwInOut{Input}{Input}
\Input{An offline dataset $\cD=\{(S_{(i)},A_{(i)},M_{(i)},R_{(i)},S_{(i)}'):1\le i\le N\}$.}Learn $p_b$ and $p_m$ based on the entire dataset $\cD$. Denote the estimators by $\widehat{p}_b$ and $\widehat{p}_m$. \\
Initialize $\widehat{Q}^{(0)}$ to be a zero function. \\
\For{$k=1,2,\ldots, K$}{
    For each $1\le i\le N$, compute the supervised learning target $Y_{(i)}=R_{(i)}+\gamma \max_a \left[\sum_{\tilde{a},m} \widehat{p}_m(m|S_{(i)}',a) \widehat{p}_b(\tilde{a}|S_{(i)}')\widehat{Q}^{(k-1)}(S_{(i)}',\tilde{a},m)\right]$. 
    \;
    Update $\widehat{Q}^{(k)} \gets \underset{Q}{\argmin} \underset{i}{\sum}  \left[Y_{(i)}-Q(S_{(i)},A_{(i)},M_{(i)})\right]^2$.\;
}
Compute the optimal policy according to \eqref{eqn:estoptimalpolicy}. \\
\SetKwInOut{Output}{Output}
\Output{$\widehat{\pi}^{(K)}$.}
\end{algorithm}

We next extend the CAL algorithm to handle distributional shift issue. We draw inspiration from existing pessimistic Q-learning methods, which seek to quantify the uncertainty of the learned Q-functions $\{\widehat{q}^{(k)}\}_k$. However, as commented in Section \ref{sec:reviewpessimism}, this is challenging in the sense that the supervised learning target involves parameter-dependent nuisance functions $\widehat{p}_m$, $\widehat{p}_b$ and $\{\widehat{Q}^{(k-1)}\}_k$, and we need to ensure that the uncertainty quantification is uniform over the entire state space. 

\textbf{Intuition Behind Penalizing $\hat{p}_m$.} As an illustrative example, let us revisit the multi-armed bandit (MAB) setting (see Figure \ref{img:mab}), which is a simplified RL problem. By applying the front-door assumption, as outlined in Assumption \ref{asm:frontdoor_crit}, and eliminating the dependence on the state, rewards depend solely on the mediator and the action. For convenience, let us assume the action space is $\{-1,0,1\}$ where $1$ represents the best arm. The estimated value function for an action $a$ in this MAB setting is:
\begin{align}\label{eq:mabq}
\hat{Q}(a)=\sum_{m}\underbrace{\sum_{\tilde{a}}\widehat{p}_b(\tilde{a})r(\tilde{a},m)}_{\widehat{R}(m)}\widehat{p}_m(m|a).
\end{align}
When the inferior actions (-1 and 1) are not adequately covered, their estimated values suffer from large variances and can be larger than the value under the optimal action, marked as $1$, with a non-negligible probability, resulting in suboptimal decision-making. To address this challenge, we note that $\tilde{a}$ is a pseudo action variable independent of $a$, the sum, $\sum_{\tilde{a}} \widehat{p}_b(\tilde{a})r(\tilde{a},m)$ (denoted by $\widehat{R}(m)$), is constant across all actions, as shown in \eqref{eq:mabq}. Consequently, the last term $\widehat{p}_m(m|a)$ which depends on $a$, becomes crucial in determining the optimal action. This inspires us to penalize  $\widehat{p}_m(m|a)$ to incorporate the pessimistic principle. When an action $a$ is less explored, the uncertainty of $\widehat{p}_m(m|a)$ is usually large. Penalizing $\widehat{p}_m$ with its uncertainty effectively avoids overfitting value functions, allowing us to correctly identify the optimal action.

We now extend our observation from MAB to RL setting. To illustrate our main idea, let $q^*(s,a)$ denote $\sum_{\tilde{a},m} p_m(m|s,a)p_b(\tilde{a}|s)Q^*(s,\tilde{a},m)$, the population limit of $\widehat{q}^{(K)}(s,a)$. A closer look at the proof of Theorem \ref{thm:converge} for Algorithm \ref{alg:cal} reveals that the regret of $\widehat{\pi}^{(K)}$ depends crucially on the difference between $\widehat{q}^{(K)}$ and $q^*$. Assuming that $p_b$ is known for now, this difference can be decomposed into the sum of
\begin{eqnarray*}
    &&\sum_{\tilde{a},m} p_m(m|s,a)p_b(\tilde{a}|s)[\widehat{Q}^{(K)}(s,\tilde{a},m)-Q^*(s,\tilde{a},m)]\\
    &+&\sum_{\tilde{a},m}[\widehat{p}_m(m|s,a)-p_m(m|s,a)]p_b(\tilde{a}|s)\widehat{Q}^{(K)}(s,\tilde{a},m).
\end{eqnarray*}
Notice that the first line can be upper bounded by $\max_m \sum_{\tilde{a}} p_b(\tilde{a}|s) |Q^*(s,\tilde{a},m)-\widehat{Q}^{(K)}(s,\tilde{a},m)|$. This indicates that, given a particular state $s$, an action $\tilde{a}$ that has not been explored as much will result in a larger loss $|Q^*(s,\tilde{a},m)-\widehat{Q}^{(K)}(s,\tilde{a},m)|$. However, this loss is mitigated by the presence of the weight $p_b(\tilde{a}|s)$, which reflects how frequently the action $\tilde{a}$ occurs in the offline data for that state $s$. As such, it is unnecessary to learn a ``conservative" $\widehat{Q}^{(K)}$ that lower bounds $Q^*$ to ensure that the first line is small. 

Regarding the second line, it depends on $a$ only through $\widehat{p}_m(m|s,a)-p_m(m|s,a)$. Thus, if this action occurs less frequently and $\widehat{p}_m$ overestimates $p_m$, it would lead to a large loss. To address this, when $\widehat{Q}^{(K)}$ is non-negative, it is reasonable to replace $\widehat{p}_m$ with its lower bound $\widehat{p}_m-\Delta$ for some uncertainty quantifier $\Delta$ to ensure the non-positivity of the second line. 

Based on the above discussion, the pessimistic causal Q-learning algorithm can be succinctly described. First, we apply a causal Q-learning algorithm (e.g., the proposed CAL algorithm described in Algorithm \ref{alg:cal}) to learn an estimated Q-function $\widehat{Q}$. Then we transform the Q-function by subtracting the minimum value, e.g., $\widetilde{Q}(s,a,m)= \widehat{Q}(s,a,m)-\inf_{(s',a',m')} \widehat{Q}(s',a',m')$. Note that based on definition of $Q$ and boundedness of $R_t$, a convenient choice of $\inf_{(s',a',m')} \widehat{Q}(s',a',m')$ would be $-V_{\max}$. Next, we learn a pointwise uncertainty quantifier $\Delta$ that satisfies the following assumption.
\begin{assumption}[Pointwise Uncertainty Quantification]\label{asump:uncertainty}
    There exists some $\alpha>0$ such that for any $s,a$ and $m$, $\mathbb{P}\left[|\widehat{p}_m(m|s,a)-p_m(m|s,a)|\le \Delta(s,a,m)\right]\ge 1-\alpha$. 
\end{assumption}
\noindent We require $\alpha$ to decay to zero in order to guarantee the consistency of the resulting estimated optimal policy (see Theorem \ref{thm:pess/pessm} for details). When the model for $p_m$ is correctly specified, the Delta method can be utilized to estimate the standard deviation of $\widehat{p}_m$ and to derive $\Delta$. 
Finally, we derive the optimal policy via
\begin{eqnarray}\label{eqn:pessiestoptimalpolicy}
    \widetilde{\pi}^*(s)=\argmax_a \left[ \sum_{\tilde{a},m}\{\widehat{p}_m(m|s,a)-\Delta(s,a,m)\} \widehat{p}_b(\tilde{a}|s)\widetilde{Q}(s,\tilde{a},m) \right]. 
\end{eqnarray}
See Algorithm \ref{alg:favi+pes} for a summary.

\begin{algorithm}[t]
\caption{PESCAL Algorithm} \label{alg:favi+pes}
\DontPrintSemicolon
\SetKwInOut{Input}{Input}
\Input{An offline dataset $\cD=\{(S_{(i)},A_{(i)},M_{(i)},R_{(i)},S_{(i)}'):1\le i\le N\}$, and a significance level $\alpha$.} Learn $p_b$ and $p_m$ based on the entire dataset $\cD$. Denote the estimators by $\widehat{p}_b$ and $\widehat{p}_m$.\\
Apply a base causal Q-learning algorithm (e.g., Algorithm \ref{alg:cal}) to learn $\widehat{Q}$.\\
Learn an uncertainty quantifier $\Delta$ that upper bounds $|\widehat{p}_m-p_m|$ with probability at least $1-\alpha$. \\
Transform the Q-function by subtracting the minimum value, e.g., for any $s,a,m$, $\widetilde{Q}(s,a,m)\leftarrow \widehat{Q}(s,a,m)-\inf_{(s',a',m')} \widehat{Q}(s',a',m')$. \\
Compute the optimal policy according to \eqref{eqn:pessiestoptimalpolicy}. \\
\SetKwInOut{Output}{Output}
\Output{$\widetilde{\pi}$.}
\end{algorithm}

Note that the stopping criterion for Algorithm \ref{alg:favi+pes} is implicitly stated in line 2, where Algorithm \ref{alg:cal} specifies that the $Q$ function update should be repeated $K$ times. We make a few remarks: 1) as commented earlier, the proposed algorithm circumvents the challenging task of quantifying the uncertainty of $\widehat{Q}$; 2) we show in Section \ref{sec:theory} that pointwise uncertainty quantification for $\widehat{p}_m$ across the state space is sufficient to ensure the consistency of $\widetilde{\pi}$, thereby eliminating the need for uniformly quantifying uncertainty. 
Overall, both 1) and 2) enhance the practicality of our algorithm; 3) as commented in the introduction, our proposal only \textit{partially} alleviates the distributional shift. To elaborate, we impose the following assumption: 
\begin{assumption}[Coverage]\label{assump:coverage}
    (i) $c_1=\sup_{s,\pi} \rho^{\pi}(s)/\rho^{\cD}(s) <\infty$; (ii)  $c_2=\sup_{s,a} p_b^{-1}(a|s)<\infty$ (for Theorem \ref{thm:converge} only); (iii) $c_3=\sup_s p_b^{-1}(\pi^*(s)|s)<\infty$ (for Theorem \ref{thm:pess/pessm} only). The supremum in (i) is taken over the class of stationary policies, $\rho^{\pi}(s)$ denotes the $\gamma$-discounted visitation distribution $(1-\gamma)\sum_{t\ge 0}\gamma^t \rho_t^{\pi}(s)$ where $\rho_t^{\pi}$ denotes the counterfactual probability mass or density function of $S_t$ assuming the system follows $\pi$, and $\rho^{\cD}(s)$ denotes the mixture of individual state distributions in the dataset $\cD$\footnote{In other words, $\rho^{\cD}=n^{-1}\sum_{i=1}^n \rho^{(i)}$ where $\rho^{(i)}$ denotes the probability mass or density function of the state variable in the $i$th trajectory.}. 
\end{assumption}
\noindent Assumption \ref{assump:coverage}(i) requires $\rho^{\cD}$ to have good coverage of the state distribution induced by all policies, as a trade-off for avoiding the challenging task of sequential uncertainty quantification of $\{\widehat{Q}^{(k)}\}_k$. In contrast, existing pessimistic Q-learning algorithms tailored for environments without mediators require the single policy coverage assumption, e.g., Assumption \ref{assump:coverage}(iii) and $\sup_s \rho^{\pi^*}(s)/\rho^{\cD}(s) <\infty$ to hold. We also remark that Assumption \ref{assump:coverage}(i) is likely to hold in settings with weak treatment effects. Specifically, in such contexts, different policies exert similar influences on state transitions, resulting in comparable $\gamma$-discounted state visitations. Meanwhile, ride-sharing applications, where the magnitude of the treatment effect typically falls within a modest range of 0.5\% to 2\%, offer common instances of these weak treatment effects \citep{xu2018large,tang2019deep}. 

Moreover, we would like to clarify that non-pessimistic policy learning algorithms, including the proposed CAL algorithm, require both Assumption \ref{assump:coverage}(i) and (ii) to hold, as demonstrated in Theorem \ref{thm:converge}. Essentially, they require the data distribution to not deviate too much from the state-action distribution induced by any possible policies. The proposed PESCAL, on the other hand, only requires Assumption \ref{assump:coverage}(i) and (iii) to be valid, as shown in Theorem \ref{thm:pess/pessm}. This means that it only requires the behavior policy to cover the optimal policy instead of any policies. 

Finally, to conclude this section, we make two remarks about the proposed algorithm. First, the proposed PESCAL is general in that it can incorporate any causal Q-learning algorithm as base algorithm. Based on the proposed CAL algorithm, there is an option to adopt the target network technique \citep{mnih2015human} to alleviate training instability. At each iteration, we sample a random minibatch of transitions $\{(s_i,a_i,m_i,r_i,s_i'):1\le i\le n\}$ from $\mathcal{D}$, computes the target  
\begin{eqnarray*}\label{eqn:target}
    y_i=r_i+\gamma \max_a \sum_{\tilde{a},m} \widehat{p}_m(m|s_i,a) \widehat{p}_b(a'|s_i)\widehat{Q}_{\tilde{\theta}}(s_i,a',m),
\end{eqnarray*}
based on the target network $\widehat{Q}_{\tilde{\theta}}$, performs a gradient descent step to update the parameter $\theta$ in the Q-network
\begin{eqnarray*}\label{eqn:gradientdescent}
    \theta \leftarrow\theta-\frac{\eta}{n}\sum_{i=1}^n [y_i-\widehat{Q}_{\theta}(s_i,a_i,m_i)]\frac{\partial \widehat{Q}_{\theta}(s_i,a_i,m_i)}{\partial \theta},
\end{eqnarray*}
for some learning rate $\eta>0$, and updates the target network parameter $\tilde{\theta}$ by delayed behavior, such as setting $\tilde{\theta}$ to $\theta$ every $T_{target}$ steps. Second, in the base causal Q-learning algorithm, an alternative to using $\widehat{p}_m$ is employing its lower bound $\widehat{p}_m-\Delta$ to produce the initial Q-estimator $\widehat{Q}$. This strategy further addresses the distributional shift problem in $\widehat{Q}$.

\section{Theoretical Guarantees}\label{sec:theory}
We first study the convergence behavior of CAL described in Algorithm \ref{alg:cal}, where the pessimistic principle is not employed. To begin with, we impose the following assumptions. Let $\cQ$, $\mathcal{P}_b$ and $\mathcal{P}_m$ denote the function spaces of the estimated $Q$-function, $\widehat{p}_b$ and $\widehat{p}_m$ respectively. Let $\nu$ be a measure on a space $X$, and function $f:X \rightarrow \mathbb {R}$. For $x\in X$, we define $\|f\|_{\nu}=\E_{x\sim \nu}(|f(x)|^2)$. In addition, define the Bellman optimality operator $\bello$ such that for any mediated Q-function, $\bello Q(s,a,m)$ equals
\begin{eqnarray*}
    \E \left[\left.R_t+\gamma \max_{a}\sum_{\tilde{a},m} Q(S_{t+1},\tilde{a},m)p_b(\tilde{a}|S_{t+1})p_m(m|S_{t+1},a)\right|S_t=s,A_t=a,M_t=m\right].
\end{eqnarray*}
\begin{assumption}[$\beta$-mixing]\label{assump:mixing}
The process that generates offline dataset $\mathcal{D}$ is stationary and exponentially $\beta$-mixing (see e.g., \cite{bradley2005basic} for a detailed definition) with $\beta$-mixing coefficients $\{\beta(\lambda)\}_{\lambda\geq 0}$, where $\beta(\lambda)\le \kappa \rho^{\lambda}$ for some constants $\kappa>0$, $0<\rho<1$.
\end{assumption}
\begin{assumption}[Completeness] \label{asm:completeness}
    For any $Q \in \cQ$, we have $\bello Q \in \cQ$. (When this assumption is violated, we define the error $\varepsilon_{\cQ}^2:=\inf_{Q_1\in \cQ}\sup_{Q_2\in \cQ} \|Q_1-\mathcal{B}^* Q_2\|_{\mu}^2$.)
\end{assumption}
\begin{assumption}[Finite Hypothesis Class]\label{asm:finite}
    $\cQ$, $\mathcal{P}_b$ and $\mathcal{P}_m$ are finite hypothesis classes, e.g., their cardinalities satisfy $|\cQ|+|\mathcal{P}_b|+|\mathcal{P}_m|<\infty$. 
\end{assumption}

Assumption \ref{assump:mixing} is a commonly used assumption adopted in existing RL literature \citep{dedecker2002maximal, antos2008learning, luckett2019estimating, shi2022multi, shi2022statistical}. The $\beta$-mixing condition is more relaxed than the i.i.d. assumption used in existing literature \citep{sutton2018reinforcement, chen2019information, uehara2020minimax,zhu2023robust}. Assumption \ref{asm:completeness} essentially requires the space of the estimated Q-functions $\cQ$ to be closed under the Bellman optimality operator and is hence referred to as the completeness assumption. It is again, widely imposed in the literature \citep{munos2008finite,chen2019information,uehara2021finite}. Notice that the completeness assumption intrinsically ensures the realizability assumption $Q^* \in \cQ$, via an application of the Banach fixed point theorem. It is worth mentioning that when Assumption \ref{asm:completeness} is violated, our proof still holds with a slight modification. In Theorem \ref{thm:thm2full} of Supplementary \ref{proof:convergence}, we provide a general theory without this completeness assumption. Third, the finite hypothesis class assumption (Assumption \ref{asm:finite}) is commonly employed in the machine learning literature to simplify the theoretical analysis \citep[see e.g.,][]{chen2019information}. It can be relaxed by requiring $\cQ$, $\mathcal{P}_b$ and $\mathcal{P}_m$ to be VC-classes  \citep{munos2008finite,uehara2021finite}. 

In the following theorems, we utilize the Kullback–Leibler (KL) divergence measure \citep{kullback1951information} to quantify the estimation errors of $\widehat{p}_b$ and $\widehat{p}_m$. For two discrete probability mass functions $p_1$ and $p_2$, the KL divergence is given by
    \begin{align*}
        \textrm{KL}(p_1\|p_2)&= \sum_{x\in \cX} p_1(x)\ln\left( \frac{p_1(x)}{p_2(x)}\right).
    \end{align*}
When $p_1$ and $p_2$ are probability density functions, the KL divergence can be similarly defined by replacing the sum with the integral, e.g., $\textrm{KL}(p_1\|p_2)=\int_{-\infty}^{\infty}p_1(x)\ln\left( \frac{p_1(x)}{p_2(x)}\right)dx$. We are now ready to present the convergence behavior of CAL, where detailed proof is provided in Supplementary \ref{proof:convergence}.

\begin{theorem}[Convergence of CAL]\label{thm:converge}
Under Assumptions \ref{asm:frontdoor_crit}, \ref{assump:coverage}(i),(ii), and \ref{assump:mixing} -- \ref{asm:finite}, where $\lambda$ in Assumption \ref{assump:mixing} is set such that $\lambda\propto\ln(N)$. Given a dataset $\mathcal{D}$ of size $N$, for some constant $C>0$, the learned policy at the $K$th iteration satisfies 
\begin{align*}
\E&[J(\pi^*)-J(\hat{\pi}_K)]\le \frac{C\sqrt{c_1}\gamma^K V_{\max}}{1-\gamma}+\frac{C\sqrt{c_1}}{(1-\gamma)^2}\Big[V_{\max}N^{-1/2}\ln(N)\sqrt{\ln(|\cQ|^2|\cP_b||\cP_m|)}\\
    &+V_{\max}\Big(\sqrt{\E \E_{S\sim \rho^{\cD}}\textrm{KL}(p_b(\bullet|S)\|\hat{p}_b(\bullet|S))}+\sqrt{c_2\E \E_{S\sim \rho^{\cD},A\sim p_b}\textrm{KL}(p_m(\bullet|S,A)\|\hat{p}_m(\bullet|S,A))}\Big)\Big].
\end{align*}
\end{theorem}
\noindent
In the last line, the first expectation in the first term is taken with respect to $\hat{p}_b$, whereas the first expectation in the second term is taken with respect to $\hat{p}_m$. The second expectations in these two terms are take with respect to the state, or state-action distribution, respectively.

According to Theorem \ref{thm:converge}, the error bound decreases with the number of iterations $K$, the sample size $N$, the KL divergence measures; and increases with $c_1$ and $c_2$ which quantify the degree of data coverage. The first term of the error bound represents the initialization bias (difference between $Q^*$ and the initial Q-estimator) which decays to zero exponentially fast with respect to the number of iterations. The subsequent terms delineate how estimation errors in the mediated Q-function $\widehat{Q}$, $\widehat{p}_b$ and $\widehat{p}_m$ influence the regret bound. When maximum likelihood estimation is employed to derive $\widehat{p}_b$ and $\widehat{p}_m$, and under certain regularity conditions, the two KL divergence measures are of the order $\sqrt{\ln(N)/N}$. 

We next establish the main statistical properties of our proposed PESCAL policy.

\begin{theorem}[Convergence of PESCAL]\label{thm:pess/pessm}
Suppose Assumptions \ref{asm:frontdoor_crit}, \ref{asump:uncertainty}, and \ref{assump:coverage}(i),(iii) hold and the initial Q-estimator $\widehat{Q}$ is bounded by $V_{\max}$. Then for some $C>0$,
\begin{eqnarray*}
    \E[J(\pi^*)-J(\tilde{\pi})]\le \frac{C}{1-\gamma}\sqrt{c_1\E\|Q^*-\hat{Q}\|^2_{\rho^{\mathcal{D}}\times p_b\times p_m}}+\frac{CV_{\max}}{1-\gamma}\Big[\sqrt{c_1\E\E_{S\sim \rho^{\mathcal{D}}}\textrm{KL}(p_b(\bullet|S)\|\hat{p}_b(\bullet|S))}\\
    +\sum_m \sqrt{c_1c_3\E\E_{S\sim \rho^{\mathcal{D}},A\sim p_b} \Delta^2(S,A,m)}+\alpha(|\mathcal{M}|+|\mathcal{A}|)\Big]. 
\end{eqnarray*}
\end{theorem}
Proof is detailed in Supplementary \ref{proof:pessim}. Compared to the results in Theorem \ref{thm:converge}, it can be seen that the error bound in Theorem \ref{thm:pess/pessm} depends on $c_3$ instead of $c_2$, demonstrating the usefulness of employing the pessimistic principle. In practice, we recommend to fit a simple parametric model, such as a logistic regression, to $p_m$. The uncertainty quantifier $\Delta$ can be set at 1.96 times the standard deviation of $\widehat{p}_m$, which can be derived using the Delta method.

\section{Simulation}\label{sec:simu_exper}
In this section, we demonstrate the effectiveness of the proposed PESCAL in Algorithm \ref{alg:favi+pes} using a toy synthetic data experiment. Let $\text{sig}(x)=\frac{e^x}{1+e^x}$ denote the sigmoid function. The simulated data is generated as follows:

\begin{enumerate}
\itemsep-0.5em 
\item Generate initial state $S_0 \in \{0,1\}$ with $\mathbb{P}(S_0=0)=\mathbb{P}(S_0=1)=0.5$.
\item Generate the unobserved confounder $C_t \in \{-1,1\}$ according to $\mathbb{P}(C_t=1|S_t)=\text{sig}(0.1S_t)$ and $\mathbb{P}(C_t=-1|S_t)=1-\mathbb{P}(C_t=1|S_t)$.
\item Generate the action $A_t \in \{-1,0,1\}$ according to $\mathbb{P}(A_t=-1|S_t,C_t)=\mathbb{P}(A_t=1|S_t,C_t)=0.5\times \text{sig}(S_t+2C_t)$ and $\mathbb{P}(A_t=0|S_t,C_t)=1-\text{sig}(S_t+2C_t)$.
\item Generate the mediator $M_t \in \{0,1\}$ according to $\mathbb{P}(M_t=0|S_t,A_t)=\text{sig}(0.1S_t+A_t)$ and $\mathbb{P}(M_t=1|S_t,A_t)=1-\mathbb{P}(M_t=0|S_t,A_t)$.
\item Generate the reward $R_t \in \{-1,1\}$ according to $\mathbb{P}(R_t=1|S_t,C_t,A_t,M_t)=\text{sig}(2C_t+0.1S_t+2M_t)$ and $\mathbb{P}(R_t=-1|S_t,C_t,A_t,M_t)=1-\mathbb{P}(R_t=1|S_t,C_t,A_t,M_t)$.
\item Generate the next state $S_{t+1} \in \{0,1\}$ according to $\mathbb{P}(S_{t+1}=1|S_t,C_t,A_t,M_t)=\text{sig}(2C_t+0.1S_t+2M_t)$ and $\mathbb{P}(S_{t+1}=0|S_t,C_t,A_t,M_t)=1-\mathbb{P}(S_{t+1}=1|S_t,C_t,A_t,M_t)$.
\end{enumerate}

\begin{figure}[h]
    \centering
    \includegraphics[width=0.25\linewidth]{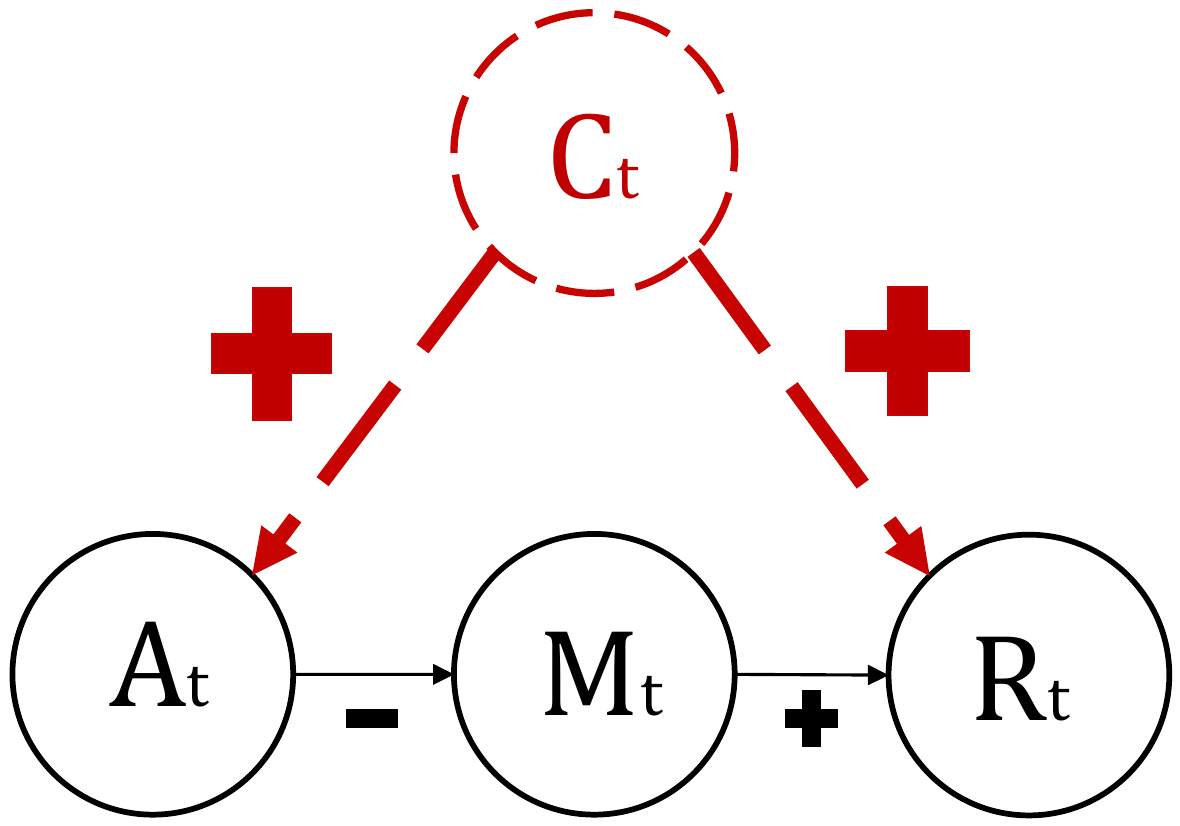}
    \caption{Confounding relationship in synthetic data conditional on state. This figure depicts the impact of an unobserved confounder ($C_t$) on both action ($A_t$) and reward ($R_t$). Larger ``$+$'' or ``$-$'' symbols indicate stronger positive or negative correlations, respectively. This plot highlights how confounding can mislead policy learning, and how the mediator variable ($M_t$) elucidates the genuine relationship between action and reward.}
    \label{img:syn}
\end{figure}

The depicted causal framework in Figure \ref{img:syn} illustrates the relationship among the confounder, action, mediator, and reward in the designed experiment, with thicker lines indicating more pronounced effects. This diagram effectively illustrates the effect of unmeasured confounder. Specifically, the unobserved confounder exhibits a strong positive correlation with both the action and the reward. Additionally, the action has a subtle negative impact on the mediator, whereas the mediator has a subtle positive influence on the reward. Consequently, under the underlying causal structure, a positive action is likely to produce a small reward. However, due to the confounder's positive influence on both the action and reward, actions might appear positively correlated with the rewards. This potential misalignment could result in suboptimal policy learning if we ignore unmeasured confounding issue. 

The complete dataset is composed of $5\times 10^4$ $\{(s, a, m, r, s')\}$ tuples, meets the full coverage assumption. To introduce partial coverage, we maintain a specific number of original tuples and remove those with suboptimal actions (where $a=0$ or $a=1$ in this case) from the remainder. Consequently, instances with actions equal to either 0 or 1 have limited data for value function estimation. This limitation can result in overestimating the value function, and consequently, a suboptimal policy.

In this example, the state, action and mediator are discrete random variables, and 
we estimate $p_b$ and $p_m$ using tabular methods. To implement PESCAL, we set the uncertainty quantifier $\Delta$ to 1.96 times the standard deviation of $\widehat{p}_m$, given by $\sqrt{\frac{\widehat{p}_m(m|s,a)(1-\widehat{p}_m(m|s,a))}{\mathbb{N}(s,a)}}$ where $\mathbb{N}(s,a)$ denotes the state-action counter. Since the environment lacks a hard stopping criterion, we set the maximum horizon length to 500 for policy evaluation. Given the simplicity of the training environment, as opposed to complex locomotion tasks (e.g., MuJoCo tasks \citep{todorov2012mujoco}), we observe that training results stabilize or exhibit a clear pattern after approximately $1 \times 10^4$ steps. Consequently, we set the number of training iterations to $1 \times 10^4$. Every 50 steps during training, we assess the current learned policy's performance based on the online discounted return, averaged over 10 trajectories with a horizon length of 500, with a moving window size of 50. Utilizing 100 random seeds, Figure \ref{fig:synresults} presents the results, where the solid lines represent the average and the shaded areas indicate the standard deviation across the random seeds.

\newcommand{\nh}{0.33\linewidth}
\newcommand{\hh}{\hspace*{0.1mm}}
\begin{figure}[htb]
\centering
\begin{tabular}{@{}c@{\hh}c@{\hh}c@{\hh}c@{}}
\rotatebox{90}{\makebox[\nh][c]{M2DP~~~~~~~~~~}}&
\includegraphics[width=\nh]{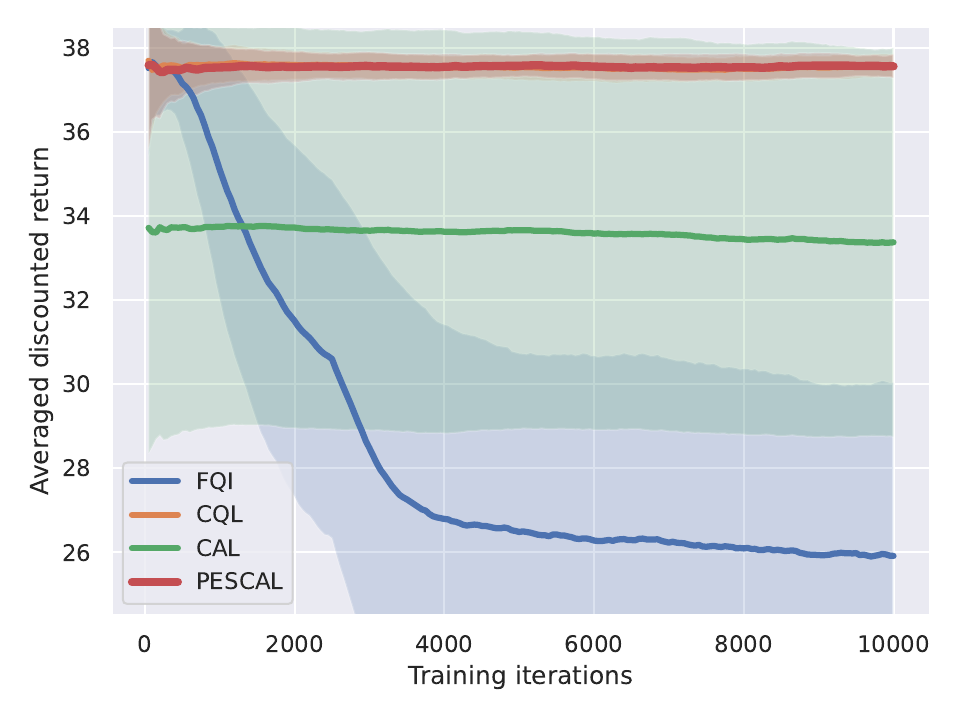}&
\includegraphics[width=\nh]{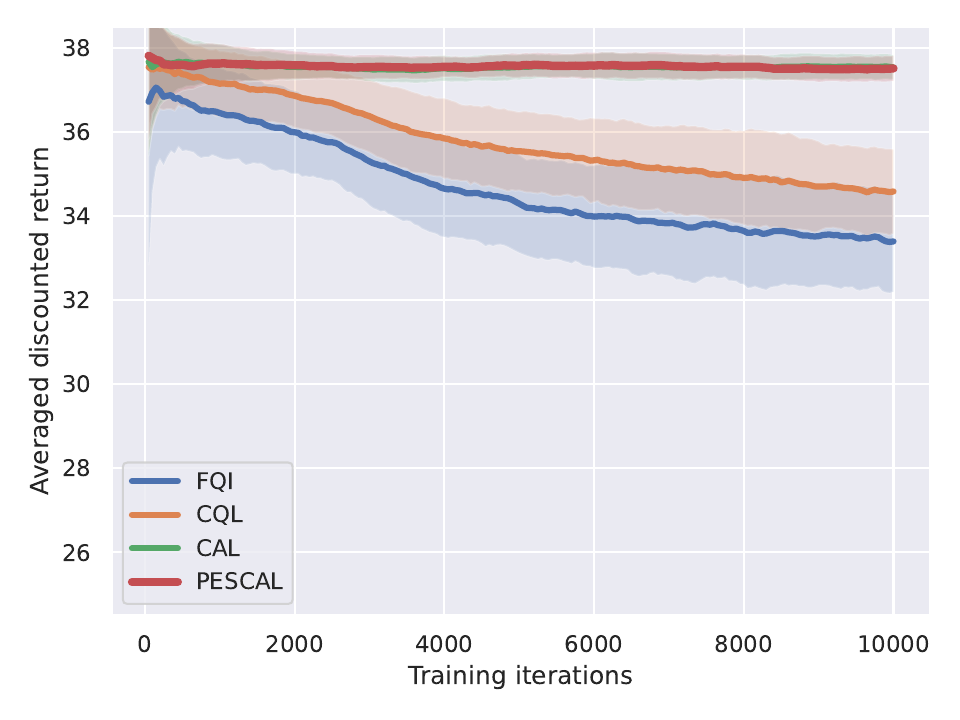}&
\includegraphics[width=\nh]{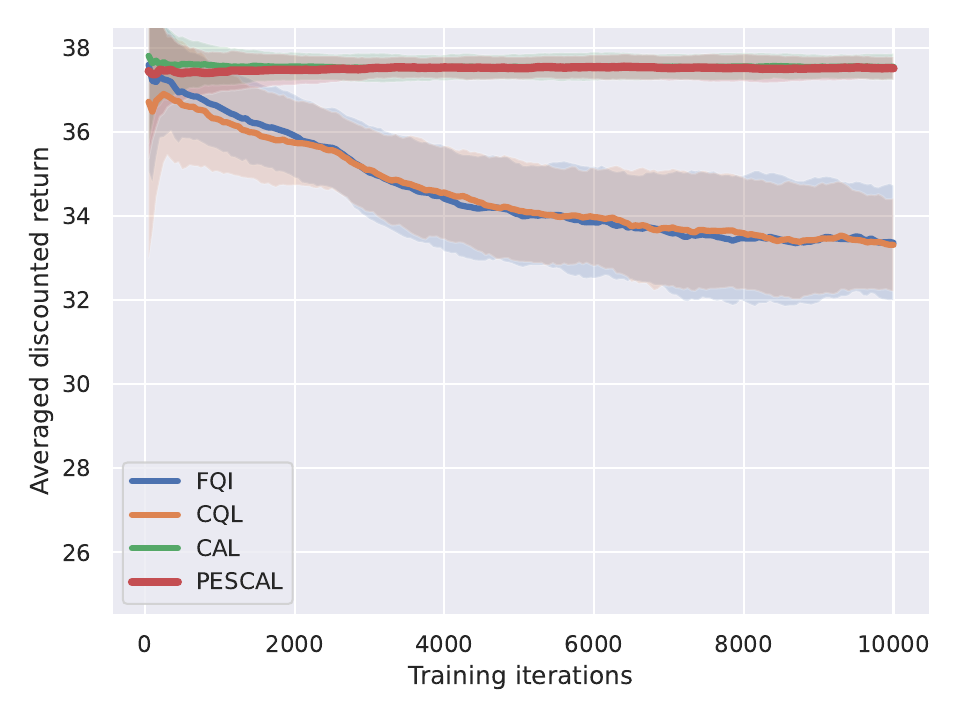}\vspace*{-15mm}\\
\rotatebox{90}{\makebox[\nh][c]{MDP~~~~~~~~~~}}&
\includegraphics[width=\nh]{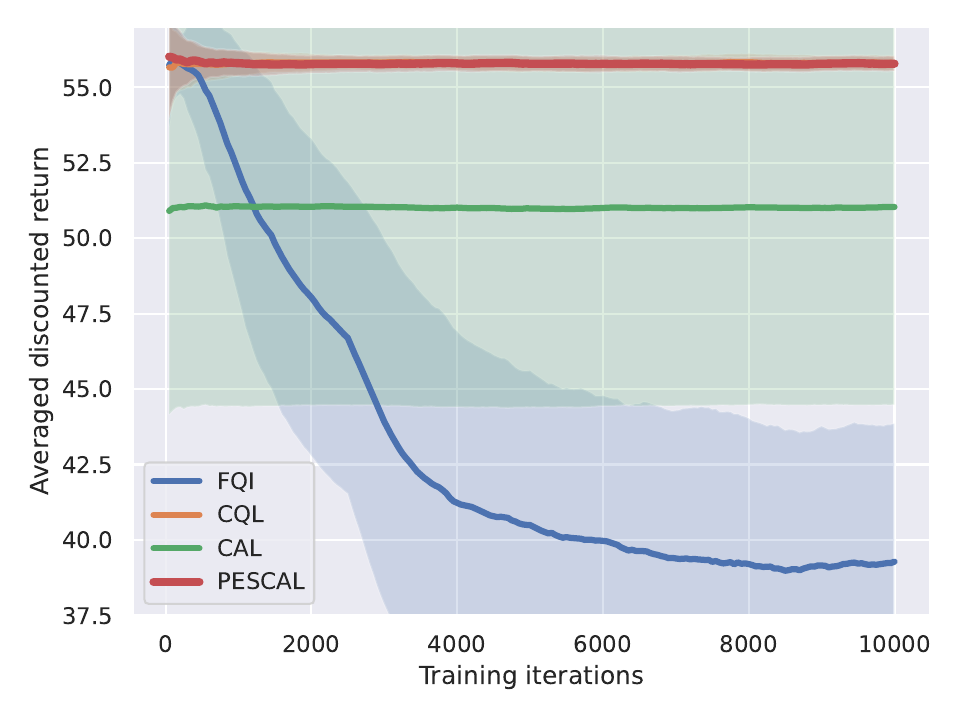}&
\includegraphics[width=\nh]{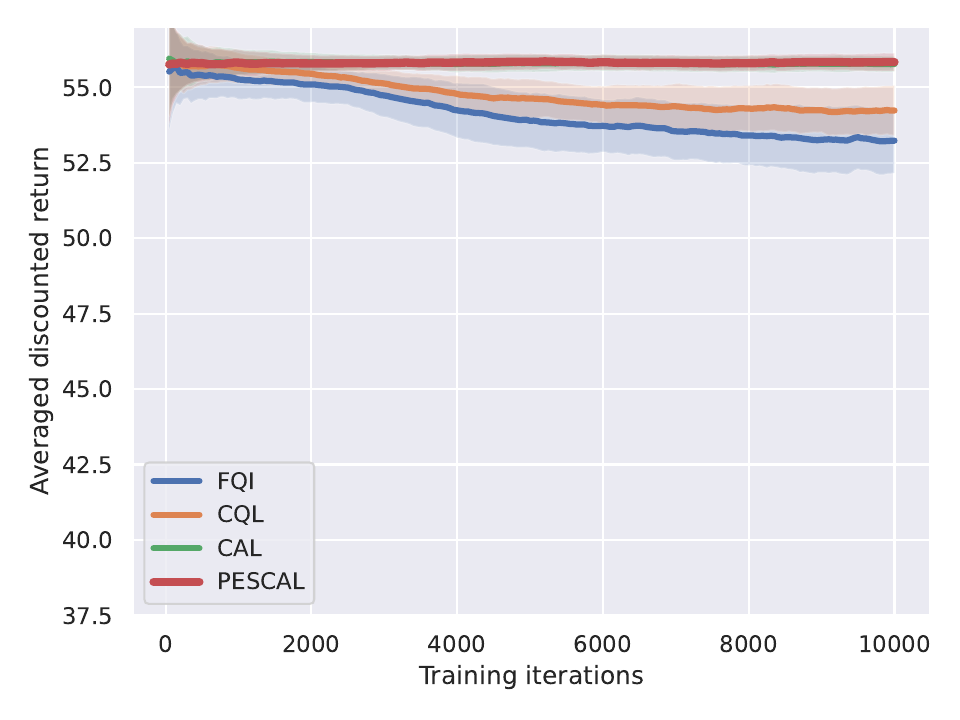}&
\includegraphics[width=\nh]{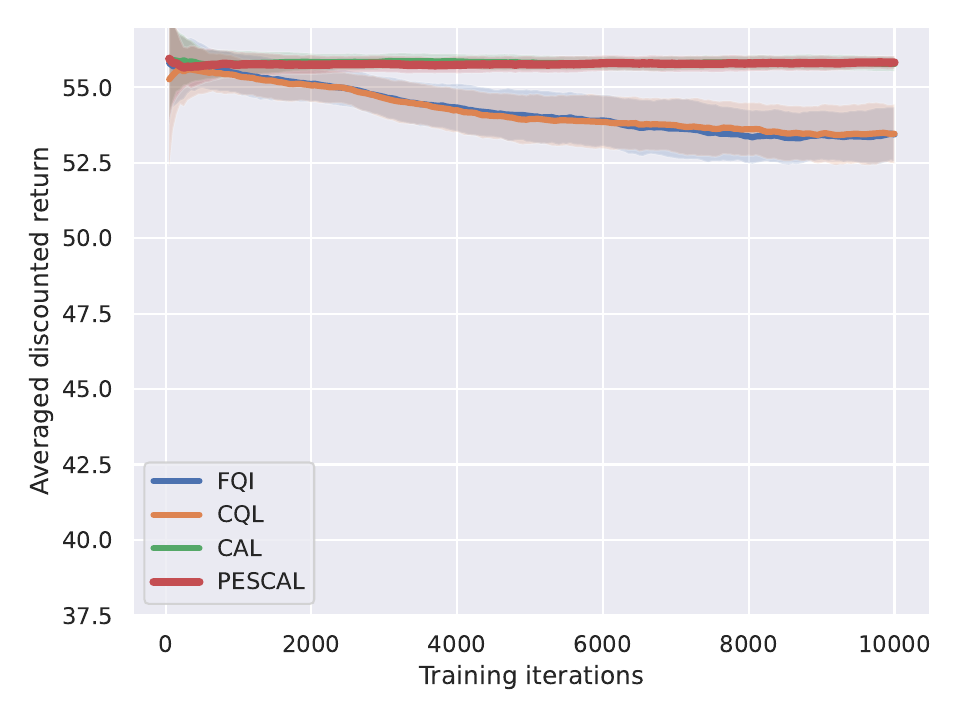}\\
&%
\small (a) Keeping 15 data points&
\small (b) Keeping half data points&
\small (c) Keeping all data points\\
\end{tabular}
\caption{\label{fig:synresults}%
Online return of the learned policy across training iterations in the synthetic data experiments. The top row, corresponding to the confounded M2DP as shown in Figure \ref{img: medi}, contrasts with the second row, which depicts a standard MDP free from confounding effects (Figure \ref{img: online_mdp}). The distinguishing feature between these two is the absence of a confounder in the second row; otherwise, both share the same environment. Furthermore, in cases (a) and (b), a certain number of original tuples are retained, while all tuples featuring suboptimal actions are eliminated from the remaining data. In contrast, case (c) retains all original data, which includes a total of 50,000 tuples.}
\end{figure}

We compare the proposed PESCAL algorithm in Algorithm \ref{alg:favi+pes} against the following baseline methods:
\begin{enumerate}
\itemsep-0.5em
\item The standard FQI algorithm which requires both positivity and unconfoundedness \citep{riedmiller2005neural}; 
\item A state-of-the-art pessimistic Q-learning algorithm CQL \citep{kumar2020conservative};
\item The proposed CAL algorithm without implementing the pessimistic principle.
\end{enumerate}

For the CQL algorithm \citep{kumar2020conservative}, we set CQL $\alpha$ to 0.1, which is the author's default implementation. The neural network used for predicting the $\hat{Q}(s,a,m)$ value comprises sequential layers: a linear layer with a dimension of 128, followed by ReLU, another linear layer with a dimension of 64, followed by ReLU, and a final linear layer. To ensure fairness in comparison, we maintain the same neural network structure across all algorithms.

The first row of Figure \ref{fig:synresults} presents results on confounded environment, while the second row presents results on unconfounded, standard environment. It can be seen from Figure \ref{fig:synresults} column (c) that when the full coverage assumption is satisfied, CAL and PESCAL demonstrate higher online returns from the learned policy, compared to FQI and CQL. Furthermore, CAL and PESCAL showcase superior stability, as evidenced by the narrower confidence interval of the online return. 

In Figure \ref{fig:synresults} columns (a) and (b), we intentionally retain only 15 and half of the original tuples, and delete those with suboptimal actions from the remainder data. As expected, the performance of FQI and CAL deteriorates as we discard more samples. Remarkably, PESCAL consistently delivers the best performance, confirming its ability to simultaneously tackle the challenges of unobserved confounders and under-coverage. Notably, CQL's performance seems to enhance as more samples are excluded. This phenomenon can be attributed to the inherent dynamics that strike a balance between unmeasured confounding and distributional shift. Specifically, transitioning from Figure \ref{fig:synresults} column (a) to column (c) amplifies the confounding issue, whereas moving in the reverse direction accentuates the under-coverage challenge. As the influence of under-coverage amplifies, CQL's performance witnesses a gradual ascent.

Notably, even in the absence of confounding, our proposed PESCAL consistently surpasses both CQL and FQI in performance (Figure \ref{fig:synresults} row 1). This enhanced performance can be credited to the incorporation of the mediator variable. This supplementary information bolsters PESCAL's effectiveness, demonstrating its effectiveness even without the presence of confounding effects.


\section{Real Data Application}\label{sec:real_exper}

In this section, we apply the PESCAL algorithm to a ride-hailing scenario. Our objective is to derive an optimal policy ($\pi^*$) for deciding whether a customer should be offered a 20\% off coupon ($A$). Notably, the action of offering this discount is susceptible to confounding ($C$) during data collection, due to factors like human interventions, as elaborated in Section \ref{sec:introduction}. Our choice of mediator ($M$) is the effective discount applied to the order. This mediator is calculated by the platform, and incorporates both the action taken and additional promotional strategies. However, it remains conditionally independent of any unmeasured variables that may contribute to unobserved confounding variable $C$. Customers see the final discount applied to their ride in the app, but remain unaware of the specific promotional strategy behind it. Therefore, it is a reasonable assumption that the action's impact on customer behavior is channeled solely through this final discount. The state $S$ is the user's information observed by the platform. While reward $R$ in general comes from two resources: 1) whether the action $A$ increases the chance a particular customer books a trip, and also increases the frequency the customer uses the app in the future; 2) whether the action helps to balance the supply and need between drivers and riders in a region. In this scenario, $R$ is quantified as a real number, pre-defined by the platform. From formulating this real world setting into an M2DP, the resulting reward and subsequent state are conditionally independent of the action, and any alternative promotional strategies when this mediator is accounted for. This setup aligns with the front-door criteria assumption, as specified in Assumption \ref{asm:frontdoor_crit}.

\begin{figure}[hbt]
\centering
\begin{tabular}{@{}c@{\hh}c@{\hh}c@{\hh}c@{}}
\rotatebox{90}{\makebox[\nh][c]{M2DP~~~~~~~~}}&
\includegraphics[width=\nh]{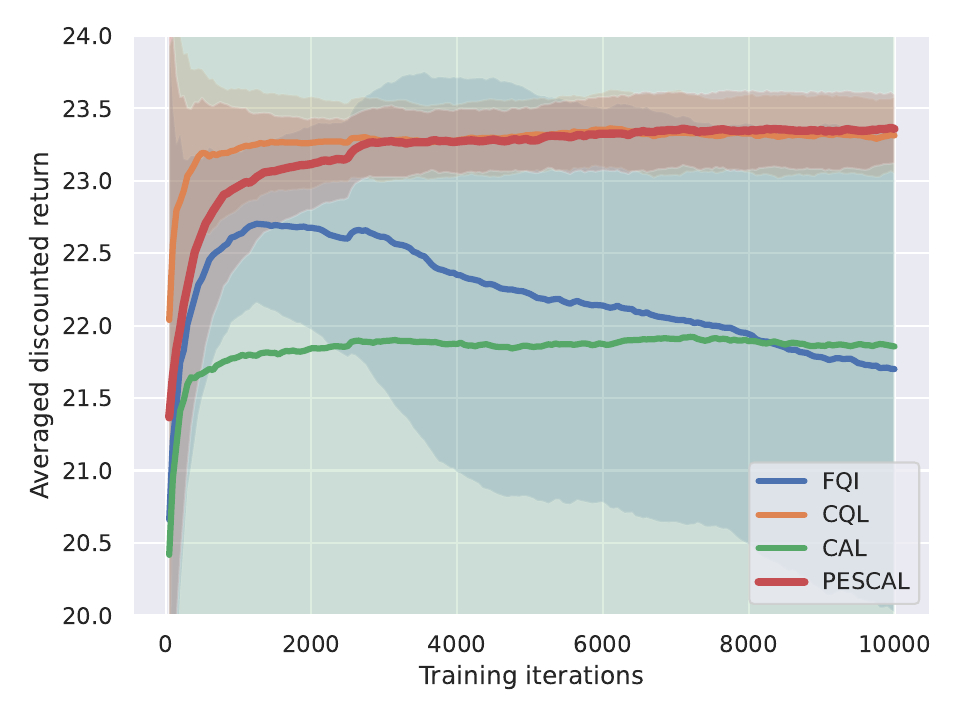}&
\includegraphics[width=\nh]{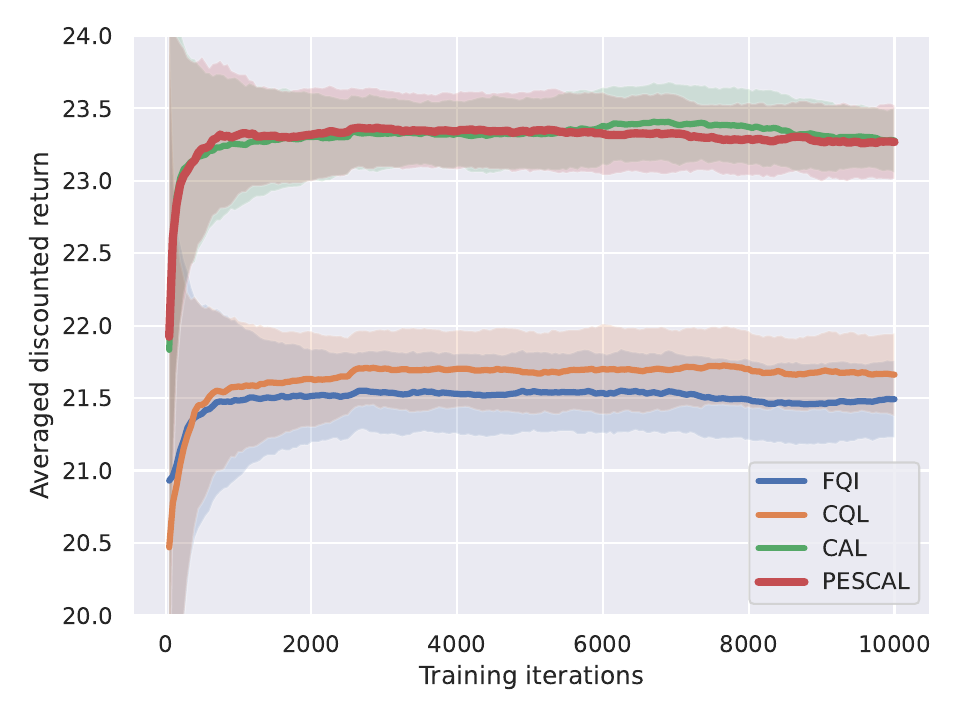}&
\includegraphics[width=\nh]{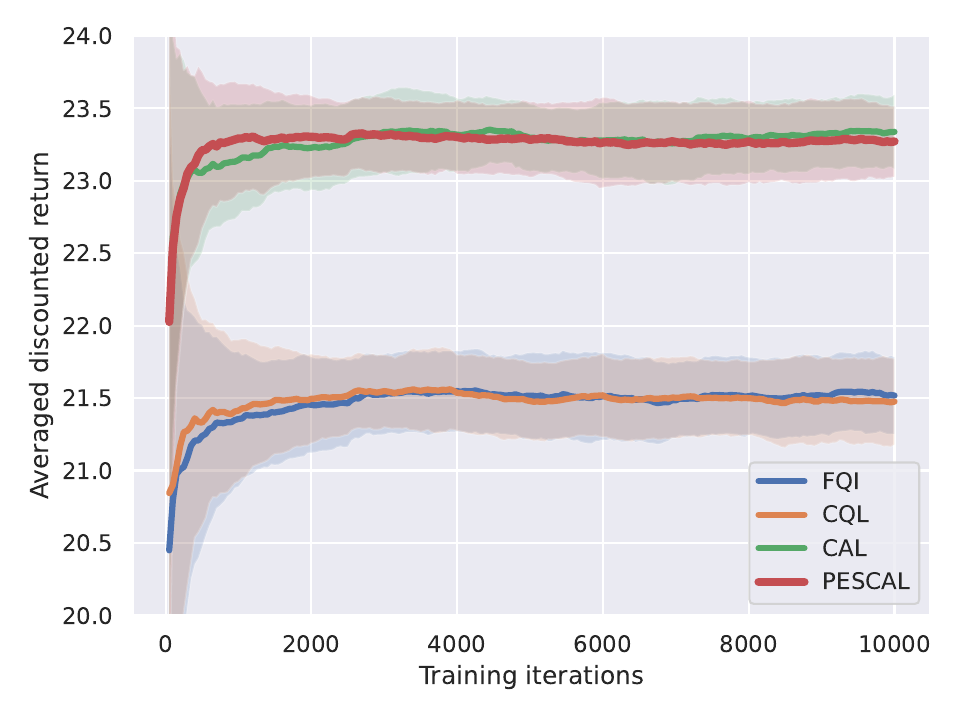}
\vspace*{-15mm}\\
\rotatebox{90}{\makebox[\nh][c]{MDP~~~~~~~~~}}&
\includegraphics[width=\nh]{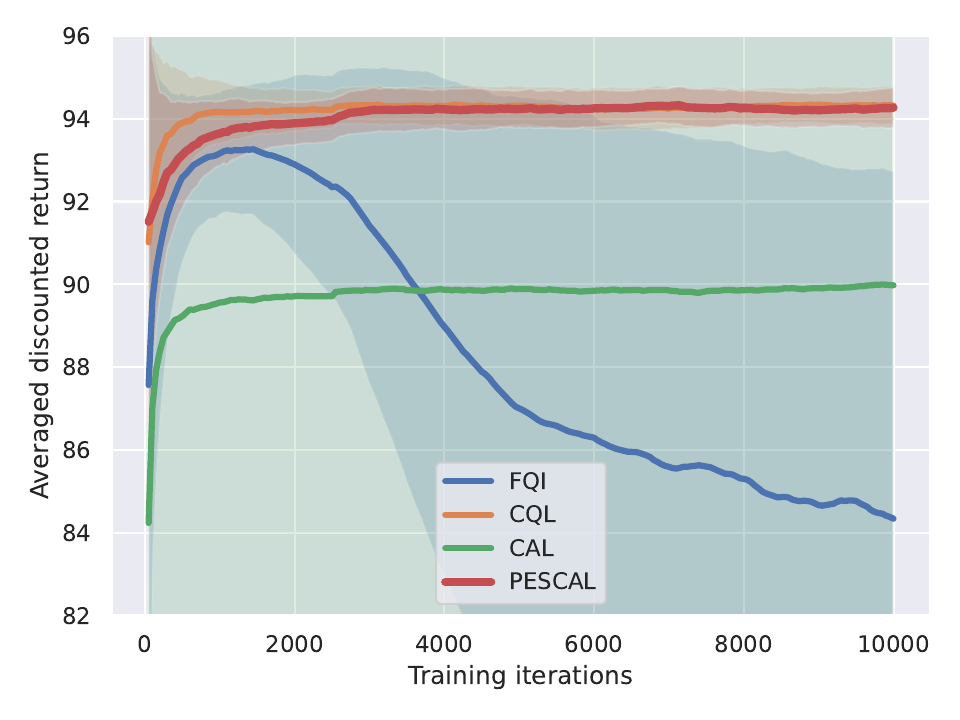}&
\includegraphics[width=\nh]{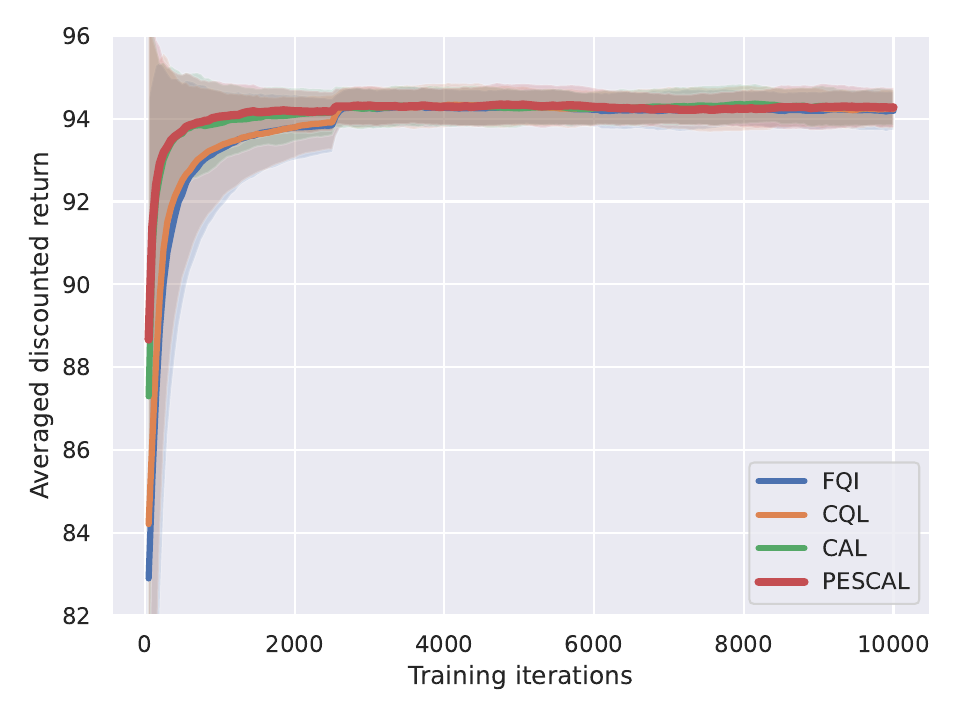}&
\includegraphics[width=\nh]{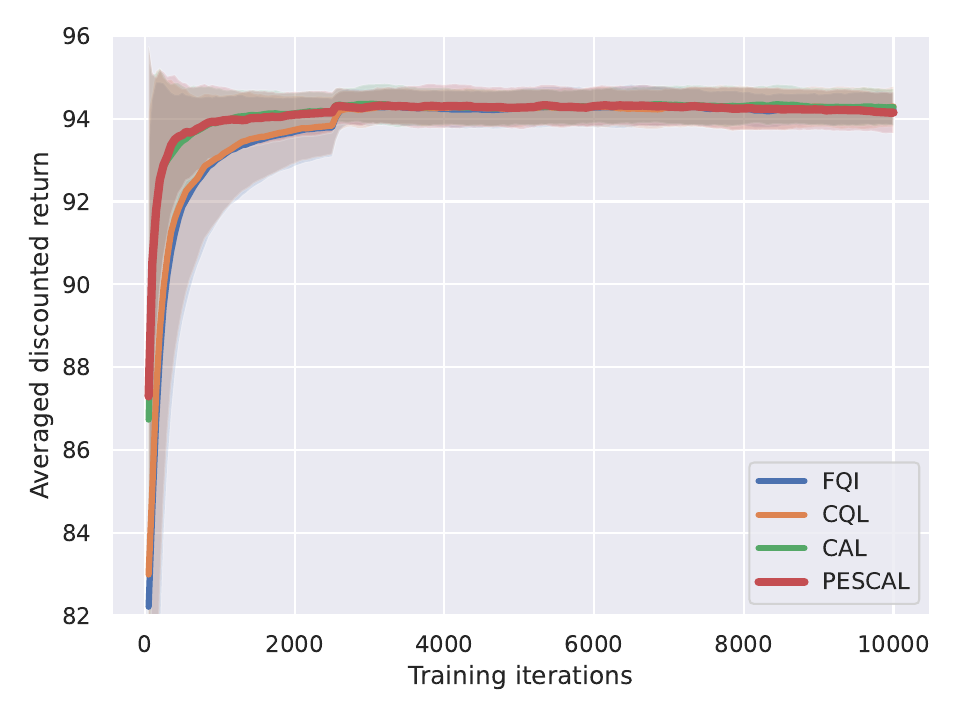}\\
&%
\small (a) Keeping 15 data&
\small (b) Keeping half data&
\small (c) Keeping all data\\
\end{tabular}
\caption{\label{fig:realresults}%
Real data results. The first row corresponds to the confounded M2DP (Figure \ref{img: medi}), while the second row pertains to a standard MDP, devoid of confounding effects (Figure \ref{img: online_mdp}). The sole environmental difference between the two rows is the absence of a confounder in the second row. Furthermore, columns (a) and (b) retain a selected number of original tuples, eliminating those with suboptimal actions from the remaining data. In contrast, column (c) maintains all original data, encompassing a total of 50,000 tuples.}
\end{figure}

We gather offline data by employing the real-data-based simulation environment developed by \cite{shi2022off}. The data generating process follows an M2DP framework, where the state distribution is characterized by a multivariate normal distribution, denoted as $N(\mu(S_t,A_t,M_t),\sigma^2)$. The conditional mean function $\mu$ is estimated through linear function approximation. Similarly, the expected reward, $\mathbb{E}[R_t|S_t,A_t,M_t]$, is also estimated via linear function approximation.

The dataset comprises 100 trajectories, each spanning a horizon of 500 steps, making there $5\times10^4$ tuples in total. Similar to the simulation experiments, we evaluate PESCAL's effectiveness by comparing its performance with that of FQI, CQL, and CAL in terms of their online average discounted return. Following the author's default settings \citep{kumar2020conservative}, we set the $\alpha$ parameter in CQL to 0.1. The multilayer perceptron used for predicting $\hat{Q}(s,a,m)$ consists of sequential layers: a linear layer with a dimension of 64, followed by a ReLU activation, another linear layer with a dimension of 32, also followed by ReLU, and a final linear layer. To ensure fairness in comparison, we maintain the same neural network structure across all four algorithms. For PESCAL, we set the uncertainty quantifier $\Delta$ to 1.96 times the standard deviation of $\widehat{p}_m$, which is approximated by the Delta method. We continue to set the total number of training steps to be $1\times 10^4$, with evaluations conducted every 50 steps, with a moving window of length 50. The results are averaged over 100 random seeds. In Figure \ref{fig:realresults}, the solid line represents the mean across the seeds, and the shaded area indicates the standard deviation across the 100 random seeds.

Similar to the synthetic data experiments, we run experiments in both confounded and unconfounded M2DP scenarios to assess our algorithms' performance, as shown in the first and second rows of Figure \ref{fig:realresults}, respectively. Our primary focus is on the confounded scenario. As depicted in Figure \ref{fig:realresults} column (c), with full data coverage, PESCAL and CAL stand out by achieving an approximately 8\% improvement over FQI and CQL. This margin represents a significant difference for a company with massive capacity. In Figure \ref{fig:realresults} columns (a) and (b), we evaluate the performance of our proposed algorithms under a partial coverage setting. 

Specifically, as the effects of the distributional shift intensify, the performance of CAL and FQI deteriorates. On the other hand, CQL's performance declines when the impact of the distributional shift is mitigated, and the confounding issue becomes more pronounced. Despite these variations, PESCAL consistently maintains a high level of performance. 

In the unconfounded M2DP scenario, when full data coverage is satisfied (column (c) in Figure \ref{fig:realresults}), the performances of all algorithms appear similar. However, as the distributional shift issue becomes more pronounced (columns (a) and (b) in Figure \ref{fig:realresults}), both PESCAL and CQL outperform CAL and FQI. This outcome highlights PESCAL's effectiveness in addressing both unobserved confounding and under-coverage challenges in real-world environments.

\section{Discussion}\label{sec:conc}
In this work, we introduce an offline RL algorithm tailored for datasets derived from confounded M2DPs. Utilizing an auxiliary mediator variable, we define a mediated Q-function, apply the front-door adjustment criteria, and demonstrate the value iteration's optimality under this framework. We present two novel online learning algorithms, CAL and PESCAL. CAL specifically addresses the issue of unobserved confounders, while PESCAL extends CAL by incorporating a pessimism strategy to address under-coverage problems. We provide theoretical guarantees regarding the convergence rates of both algorithms. Through experimental validation on synthetic and real-world datasets, we illustrate PESCAL's superior efficiency compared to CAL, as well as to two established offline RL algorithms, FQI and CQL. Notably, PESCAL stands out by simultaneously addressing the critical challenges of unobserved confounders and under-coverage.

Based on the theoretical insights presented in Section \ref{sec:theory}, PESCAL demonstrates significantly enhanced effectiveness in contexts where both the action and mediator spaces are finite and well-defined. However, as these spaces grow large or become infinite, discretizing the action and mediator variables is essential to unlock the algorithm's full performance potential. 
The development of effective learning algorithms for continuous control problems, particularly those influenced by unobserved confounders with continuous action and mediator variables, stands as a challenging frontier that deserves future research efforts.

\baselineskip=16pt
\bibliographystyle{asa}
\bibliography{ref_favi}

\begin{thebibliography}{82}
\newcommand{\enquote}[1]{``#1''}
\expandafter\ifx\csname natexlab\endcsname\relax\def\natexlab#1{#1}\fi

\bibitem[{Antos et~al.(2008)Antos, Szepesv{\'a}ri, and Munos}]{antos2008learning}
Antos, A., Szepesv{\'a}ri, C., and Munos, R. (2008), \enquote{Learning near-optimal policies with Bellman-residual minimization based fitted policy iteration and a single sample path,} \textit{Machine Learning}, 71, 89--129.

\bibitem[{Banach(1922)}]{banach1922operations}
Banach, S. (1922), \enquote{Sur les op{\'e}rations dans les ensembles abstraits et leur application aux {\'e}quations int{\'e}grales,} \textit{Fundamenta mathematicae}, 3, 133--181.

\bibitem[{Berner et~al.(2019)Berner, Brockman, Chan, Cheung, Debiak, Dennison, Farhi, Fischer, Hashme, Hesse, et~al.}]{berner2019dota}
Berner, C., Brockman, G., Chan, B., Cheung, V., Debiak, P., Dennison, C., Farhi, D., Fischer, Q., Hashme, S., Hesse, C., et~al. (2019), \enquote{Dota 2 with large scale deep reinforcement learning,} \textit{arXiv preprint arXiv:1912.06680}.

\bibitem[{Bernstein(1946)}]{bernstein1946theory}
Bernstein, S.~N. (1946), \textit{Theory of Probability (in Russian)}, Moscow.

\bibitem[{Bradley(2005)}]{bradley2005basic}
Bradley, R.~C. (2005), \enquote{Basic properties of strong mixing conditions. A survey and some open questions,} .

\bibitem[{Bruns-Smith and Zhou(2023)}]{bruns2023robust}
Bruns-Smith, D. and Zhou, A. (2023), \enquote{Robust Fitted-Q-Evaluation and Iteration under Sequentially Exogenous Unobserved Confounders,} \textit{arXiv preprint arXiv:2302.00662}.

\bibitem[{Chakraborty and Moodie(2013)}]{DTRs}
Chakraborty, B. and Moodie, E. E.~M. (2013), \textit{Statistical methods for dynamic treatment regimes}, New York: Springer.

\bibitem[{Chang et~al.(2021)Chang, Uehara, Sreenivas, Kidambi, and Sun}]{chang2021mitigating}
Chang, J., Uehara, M., Sreenivas, D., Kidambi, R., and Sun, W. (2021), \enquote{Mitigating covariate shift in imitation learning via offline data with partial coverage,} \textit{Advances in Neural Information Processing Systems}, 34, 965--979.

\bibitem[{Chen and Jiang(2019)}]{chen2019information}
Chen, J. and Jiang, N. (2019), \enquote{Information-theoretic considerations in batch reinforcement learning,} in \textit{International Conference on Machine Learning}, PMLR, pp. 1042--1051.

\bibitem[{Chen and Zhang(2021)}]{chen2021estimating}
Chen, S. and Zhang, B. (2021), \enquote{Estimating and improving dynamic treatment regimes with a time-varying instrumental variable,} \textit{arXiv preprint arXiv:2104.07822}.

\bibitem[{Dedecker and Louhichi(2002)}]{dedecker2002maximal}
Dedecker, J. and Louhichi, S. (2002), \enquote{Maximal inequalities and empirical central limit theorems,} in \textit{Empirical process techniques for dependent data}, Springer, pp. 137--159.

\bibitem[{Ertefaie and Strawderman(2018)}]{ertefaie2018constructing}
Ertefaie, A. and Strawderman, R.~L. (2018), \enquote{Constructing dynamic treatment regimes over indefinite time horizons,} \textit{Biometrika}, 105, 963--977.

\bibitem[{Fu et~al.(2022)Fu, Qi, Wang, Yang, Xu, and Kosorok}]{fu2022offline}
Fu, Z., Qi, Z., Wang, Z., Yang, Z., Xu, Y., and Kosorok, M.~R. (2022), \enquote{Offline reinforcement learning with instrumental variables in confounded markov decision processes,} \textit{arXiv preprint arXiv:2209.08666}.

\bibitem[{Fulcher et~al.(2020)Fulcher, Shpitser, Marealle, and Tchetgen~Tchetgen}]{fulcher2020robust}
Fulcher, I.~R., Shpitser, I., Marealle, S., and Tchetgen~Tchetgen, E.~J. (2020), \enquote{Robust inference on population indirect causal effects: the generalized front door criterion,} \textit{Journal of the Royal Statistical Society Series B: Statistical Methodology}, 82, 199--214.

\bibitem[{Imbens and Rubin(2015)}]{imbens2015causal}
Imbens, G.~W. and Rubin, D.~B. (2015), \textit{Causal inference in statistics, social, and biomedical sciences}, Cambridge University Press.

\bibitem[{Jin et~al.(2022)Jin, Ren, Yang, and Wang}]{jin2022policy}
Jin, Y., Ren, Z., Yang, Z., and Wang, Z. (2022), \enquote{Policy learning" without''overlap: Pessimism and generalized empirical Bernstein's inequality,} \textit{arXiv preprint arXiv:2212.09900}.

\bibitem[{Jin et~al.(2021)Jin, Yang, and Wang}]{jin2021pessimism}
Jin, Y., Yang, Z., and Wang, Z. (2021), \enquote{Is pessimism provably efficient for offline rl?} in \textit{International Conference on Machine Learning}, PMLR, pp. 5084--5096.

\bibitem[{Kakade and Langford(2002)}]{kakade2002approximately}
Kakade, S. and Langford, J. (2002), \enquote{Approximately optimal approximate reinforcement learning,} in \textit{Proceedings of the Nineteenth International Conference on Machine Learning}, pp. 267--274.

\bibitem[{Kallus and Zhou(2018)}]{kallus2018confounding}
Kallus, N. and Zhou, A. (2018), \enquote{Confounding-robust policy improvement,} \textit{Advances in neural information processing systems}, 31.

\bibitem[{Kallus and Zhou(2020)}]{kallus2020confounding}
--- (2020), \enquote{Confounding-robust policy evaluation in infinite-horizon reinforcement learning,} \textit{Advances in neural information processing systems}, 33, 22293--22304.

\bibitem[{Kallus and Zhou(2022)}]{kallus2022stateful}
--- (2022), \enquote{Stateful offline contextual policy evaluation and learning,} in \textit{International Conference on Artificial Intelligence and Statistics}, PMLR, pp. 11169--11194.

\bibitem[{Kleinberg et~al.(2018)Kleinberg, Lakkaraju, Leskovec, Ludwig, and Mullainathan}]{kleinberg2018human}
Kleinberg, J., Lakkaraju, H., Leskovec, J., Ludwig, J., and Mullainathan, S. (2018), \enquote{Human decisions and machine predictions,} \textit{The quarterly journal of economics}, 133, 237--293.

\bibitem[{Kosorok and Laber(2019)}]{kosorok2019precision}
Kosorok, M.~R. and Laber, E.~B. (2019), \enquote{Precision medicine,} \textit{Annual review of statistics and its application}, 6, 263--286.

\bibitem[{Kullback and Leibler(1951)}]{kullback1951information}
Kullback, S. and Leibler, R.~A. (1951), \enquote{On information and sufficiency,} \textit{The annals of mathematical statistics}, 22, 79--86.

\bibitem[{Kumar et~al.(2020)Kumar, Zhou, Tucker, and Levine}]{kumar2020conservative}
Kumar, A., Zhou, A., Tucker, G., and Levine, S. (2020), \enquote{Conservative q-learning for offline reinforcement learning,} \textit{Advances in Neural Information Processing Systems}, 33, 1179--1191.

\bibitem[{LeCun et~al.(2015)LeCun, Bengio, and Hinton}]{lecun2015deep}
LeCun, Y., Bengio, Y., and Hinton, G. (2015), \enquote{Deep learning,} \textit{nature}, 521, 436--444.

\bibitem[{Levine et~al.(2020)Levine, Kumar, Tucker, and Fu}]{levine2020offline}
Levine, S., Kumar, A., Tucker, G., and Fu, J. (2020), \enquote{Offline reinforcement learning: Tutorial, review, and perspectives on open problems,} \textit{arXiv preprint arXiv:2005.01643}.

\bibitem[{Li et~al.(2021)Li, Luo, and Zhang}]{li2021causal}
Li, J., Luo, Y., and Zhang, X. (2021), \enquote{Causal reinforcement learning: An instrumental variable approach,} \textit{arXiv preprint arXiv:2103.04021}.

\bibitem[{Liao et~al.(2021)Liao, Fu, Yang, Wang, Kolar, and Wang}]{liao2021instrumental}
Liao, L., Fu, Z., Yang, Z., Wang, Y., Kolar, M., and Wang, Z. (2021), \enquote{Instrumental variable value iteration for causal offline reinforcement learning,} \textit{arXiv preprint arXiv:2102.09907}.

\bibitem[{Lu et~al.(2018)Lu, Sch{\"o}lkopf, and Hern{\'a}ndez-Lobato}]{lu2018deconfounding}
Lu, C., Sch{\"o}lkopf, B., and Hern{\'a}ndez-Lobato, J.~M. (2018), \enquote{Deconfounding reinforcement learning in observational settings,} \textit{arXiv preprint arXiv:1812.10576}.

\bibitem[{Lu et~al.(2022)Lu, Min, Wang, and Yang}]{lu2022pessimism}
Lu, M., Min, Y., Wang, Z., and Yang, Z. (2022), \enquote{Pessimism in the face of confounders: Provably efficient offline reinforcement learning in partially observable markov decision processes,} \textit{arXiv preprint arXiv:2205.13589}.

\bibitem[{Luckett et~al.(2020)Luckett, Laber, Kahkoska, Maahs, Mayer-Davis, and Kosorok}]{luckett2019estimating}
Luckett, D.~J., Laber, E.~B., Kahkoska, A.~R., Maahs, D.~M., Mayer-Davis, E., and Kosorok, M.~R. (2020), \enquote{Estimating Dynamic Treatment Regimes in Mobile Health Using V-learning,} \textit{Journal of the American Statistical Association}, 115, 692.

\bibitem[{Mnih et~al.(2013)Mnih, Kavukcuoglu, Silver, Graves, Antonoglou, Wierstra, and Riedmiller}]{mnih2013playing}
Mnih, V., Kavukcuoglu, K., Silver, D., Graves, A., Antonoglou, I., Wierstra, D., and Riedmiller, M. (2013), \enquote{Playing atari with deep reinforcement learning,} \textit{arXiv preprint arXiv:1312.5602}.

\bibitem[{Mnih et~al.(2015)Mnih, Kavukcuoglu, Silver, Rusu, Veness, Bellemare, Graves, Riedmiller, Fidjeland, Ostrovski, et~al.}]{mnih2015human}
Mnih, V., Kavukcuoglu, K., Silver, D., Rusu, A.~A., Veness, J., Bellemare, M.~G., Graves, A., Riedmiller, M., Fidjeland, A.~K., Ostrovski, G., et~al. (2015), \enquote{Human-level control through deep reinforcement learning,} \textit{nature}, 518, 529--533.

\bibitem[{Mo et~al.(2021)Mo, Qi, and Liu}]{mo2021learning}
Mo, W., Qi, Z., and Liu, Y. (2021), \enquote{Learning optimal distributionally robust individualized treatment rules,} \textit{Journal of the American Statistical Association}, 116, 659--674.

\bibitem[{Munos and Szepesv{\'a}ri(2008)}]{munos2008finite}
Munos, R. and Szepesv{\'a}ri, C. (2008), \enquote{Finite-Time Bounds for Fitted Value Iteration.} \textit{Journal of Machine Learning Research}, 9.

\bibitem[{Murphy(2003)}]{Murphy}
Murphy, S.~A. (2003), \enquote{Optimal dynamic treatment regimes,} \textit{Journal of the Royal Statistical Society: Series B}, 65, 331--355.

\bibitem[{Nadler~Jr(1969)}]{nadler1969multi}
Nadler~Jr, S.~B. (1969), \enquote{Multi-valued contraction mappings.} .

\bibitem[{Namkoong et~al.(2020)Namkoong, Keramati, Yadlowsky, and Brunskill}]{namkoong2020off}
Namkoong, H., Keramati, R., Yadlowsky, S., and Brunskill, E. (2020), \enquote{Off-policy policy evaluation for sequential decisions under unobserved confounding,} \textit{Advances in Neural Information Processing Systems}, 33, 18819--18831.

\bibitem[{Nie and Wager(2021)}]{nie2021quasi}
Nie, X. and Wager, S. (2021), \enquote{Quasi-oracle estimation of heterogeneous treatment effects,} \textit{Biometrika}, 108, 299--319.

\bibitem[{OpenAI(2023)}]{openai2023gpt4}
OpenAI (2023), \enquote{GPT-4 Technical Report,} .

\bibitem[{Pearl(2009)}]{pearl2009causality}
Pearl, J. (2009), \textit{Causality}, Cambridge university press.

\bibitem[{Puterman(2014)}]{puterman2014markov}
Puterman, M.~L. (2014), \textit{Markov decision processes: discrete stochastic dynamic programming}, John Wiley \& Sons.

\bibitem[{Qi et~al.(2020)Qi, Liu, Fu, and Liu}]{qi2020multi}
Qi, Z., Liu, D., Fu, H., and Liu, Y. (2020), \enquote{Multi-armed angle-based direct learning for estimating optimal individualized treatment rules with various outcomes,} \textit{Journal of the American Statistical Association}, 115, 678--691.

\bibitem[{Qi et~al.(2022)Qi, Tang, Fang, and Shi}]{qi2022offline}
Qi, Z., Tang, J., Fang, E., and Shi, C. (2022), \enquote{Offline personalized pricing with censored demand,} in \textit{Offline Personalized Pricing with Censored Demand: Qi, Zhengling| uTang, Jingwen| uFang, Ethan| uShi, Cong}, [Sl]: SSRN.

\bibitem[{Qian and Murphy(2011)}]{qian2011performance}
Qian, M. and Murphy, S.~A. (2011), \enquote{Performance guarantees for individualized treatment rules,} \textit{Annals of statistics}, 39, 1180.

\bibitem[{Rashidinejad et~al.(2021)Rashidinejad, Zhu, Ma, Jiao, and Russell}]{rashidinejad2021bridging}
Rashidinejad, P., Zhu, B., Ma, C., Jiao, J., and Russell, S. (2021), \enquote{Bridging offline reinforcement learning and imitation learning: A tale of pessimism,} \textit{Advances in Neural Information Processing Systems}, 34, 11702--11716.

\bibitem[{Riedmiller(2005)}]{riedmiller2005neural}
Riedmiller, M. (2005), \enquote{Neural fitted Q iteration--first experiences with a data efficient neural reinforcement learning method,} in \textit{Machine Learning: ECML 2005: 16th European Conference on Machine Learning, Porto, Portugal, October 3-7, 2005. Proceedings 16}, Springer, pp. 317--328.

\bibitem[{Robins(2004)}]{gest}
Robins, J.~M. (2004), \enquote{Optimal structural nested models for optimal sequential decisions,} in \textit{Proceedings of the Second Seattle Symposium in Biostatistics}, eds. Lin, D.~Y. and Heagerty, P., Springer, New York, pp. 189--326.

\bibitem[{Shi et~al.(2018)Shi, Fan, Song, and Lu}]{shi2018high}
Shi, C., Fan, A., Song, R., and Lu, W. (2018), \enquote{High-dimensional A-learning for optimal dynamic treatment regimes,} \textit{Annals of statistics}, 46, 925.

\bibitem[{Shi et~al.(2022{\natexlab{a}})Shi, Wan, Song, Luo, Song, and Zhu}]{shi2022multi}
Shi, C., Wan, R., Song, G., Luo, S., Song, R., and Zhu, H. (2022{\natexlab{a}}), \enquote{A multi-agent reinforcement learning framework for off-policy evaluation in two-sided markets,} \textit{arXiv preprint arXiv:2202.10574}.

\bibitem[{Shi et~al.(2022{\natexlab{b}})Shi, Zhang, Lu, and Song}]{shi2022statistical}
Shi, C., Zhang, S., Lu, W., and Song, R. (2022{\natexlab{b}}), \enquote{Statistical inference of the value function for reinforcement learning in infinite-horizon settings,} \textit{Journal of the Royal Statistical Society Series B: Statistical Methodology}, 84, 765--793.

\bibitem[{Shi et~al.(2022{\natexlab{c}})Shi, Zhu, Ye, Luo, Zhu, and Song}]{shi2022off}
Shi, C., Zhu, J., Ye, S., Luo, S., Zhu, H., and Song, R. (2022{\natexlab{c}}), \enquote{Off-policy confidence interval estimation with confounded markov decision process,} \textit{Journal of the American Statistical Association}, 1--12.

\bibitem[{Shi et~al.(2022{\natexlab{d}})Shi, Li, Wei, Chen, and Chi}]{shi2022pessimistic}
Shi, L., Li, G., Wei, Y., Chen, Y., and Chi, Y. (2022{\natexlab{d}}), \enquote{Pessimistic q-learning for offline reinforcement learning: Towards optimal sample complexity,} in \textit{International Conference on Machine Learning}, PMLR, pp. 19967--20025.

\bibitem[{Siciliano et~al.(2008)Siciliano, Khatib, and Kr{\"o}ger}]{siciliano2008springer}
Siciliano, B., Khatib, O., and Kr{\"o}ger, T. (2008), \textit{Springer handbook of robotics}, vol. 200, Springer.

\bibitem[{Silver et~al.(2016)Silver, Huang, Maddison, Guez, Sifre, Van Den~Driessche, Schrittwieser, Antonoglou, Panneershelvam, Lanctot, et~al.}]{silver2016mastering}
Silver, D., Huang, A., Maddison, C.~J., Guez, A., Sifre, L., Van Den~Driessche, G., Schrittwieser, J., Antonoglou, I., Panneershelvam, V., Lanctot, M., et~al. (2016), \enquote{Mastering the game of Go with deep neural networks and tree search,} \textit{nature}, 529, 484--489.

\bibitem[{Sutton and Barto(2018)}]{sutton2018reinforcement}
Sutton, R.~S. and Barto, A.~G. (2018), \textit{Reinforcement learning: An introduction}, MIT press.

\bibitem[{Tan et~al.(2022)Tan, Lu, Kausik, Wang, and Tewari}]{tan2022offline}
Tan, K., Lu, Y., Kausik, C., Wang, Y., and Tewari, A. (2022), \enquote{Offline Policy Evaluation and Optimization under Confounding,} \textit{arXiv preprint arXiv:2211.16583}.

\bibitem[{Tang et~al.(2019)Tang, Qin, Zhang, Wang, Xu, Ma, Zhu, and Ye}]{tang2019deep}
Tang, X., Qin, Z., Zhang, F., Wang, Z., Xu, Z., Ma, Y., Zhu, H., and Ye, J. (2019), \enquote{A deep value-network based approach for multi-driver order dispatching,} in \textit{Proceedings of the 25th ACM SIGKDD international conference on knowledge discovery \& data mining}, pp. 1780--1790.

\bibitem[{Tennenholtz et~al.(2020)Tennenholtz, Shalit, and Mannor}]{tennenholtz2020off}
Tennenholtz, G., Shalit, U., and Mannor, S. (2020), \enquote{Off-policy evaluation in partially observable environments,} in \textit{Proceedings of the AAAI Conference on Artificial Intelligence}, vol.~34, pp. 10276--10283.

\bibitem[{Todorov et~al.(2012)Todorov, Erez, and Tassa}]{todorov2012mujoco}
Todorov, E., Erez, T., and Tassa, Y. (2012), \enquote{MuJoCo: A physics engine for model-based control,} in \textit{2012 IEEE/RSJ International Conference on Intelligent Robots and Systems}, IEEE, pp. 5026--5033.

\bibitem[{Tsiatis et~al.(2019)Tsiatis, Davidian, Holloway, and Laber}]{tsiatis2019dynamic}
Tsiatis, A.~A., Davidian, M., Holloway, S.~T., and Laber, E.~B. (2019), \textit{Dynamic treatment regimes: Statistical methods for precision medicine}, CRC press.

\bibitem[{Uehara et~al.(2020)Uehara, Huang, and Jiang}]{uehara2020minimax}
Uehara, M., Huang, J., and Jiang, N. (2020), \enquote{Minimax weight and q-function learning for off-policy evaluation,} in \textit{International Conference on Machine Learning}, PMLR, pp. 9659--9668.

\bibitem[{Uehara et~al.(2021)Uehara, Imaizumi, Jiang, Kallus, Sun, and Xie}]{uehara2021finite}
Uehara, M., Imaizumi, M., Jiang, N., Kallus, N., Sun, W., and Xie, T. (2021), \enquote{Finite sample analysis of minimax offline reinforcement learning: Completeness, fast rates and first-order efficiency,} \textit{arXiv preprint arXiv:2102.02981}.

\bibitem[{Uehara and Sun(2022)}]{ueharapessimistic}
Uehara, M. and Sun, W. (2022), \enquote{Pessimistic Model-based Offline Reinforcement Learning under Partial Coverage,} in \textit{International Conference on Learning Representations}.

\bibitem[{Wang et~al.(2022)Wang, Qi, and Shi}]{wang2022blessing}
Wang, J., Qi, Z., and Shi, C. (2022), \enquote{Blessing from Experts: Super Reinforcement Learning in Confounded Environments,} \textit{arXiv preprint arXiv:2209.15448}.

\bibitem[{Wang et~al.(2021)Wang, Yang, and Wang}]{wang2021provably}
Wang, L., Yang, Z., and Wang, Z. (2021), \enquote{Provably efficient causal reinforcement learning with confounded observational data,} \textit{Advances in Neural Information Processing Systems}, 34, 21164--21175.

\bibitem[{Wang et~al.(2018)Wang, Zhou, Song, and Sherwood}]{wang2018quantile}
Wang, L., Zhou, Y., Song, R., and Sherwood, B. (2018), \enquote{Quantile-optimal treatment regimes,} \textit{Journal of the American Statistical Association}, 113, 1243--1254.

\bibitem[{Xie et~al.(2021)Xie, Cheng, Jiang, Mineiro, and Agarwal}]{xie2021bellman}
Xie, T., Cheng, C.-A., Jiang, N., Mineiro, P., and Agarwal, A. (2021), \enquote{Bellman-consistent pessimism for offline reinforcement learning,} \textit{Advances in neural information processing systems}, 34, 6683--6694.

\bibitem[{Xu et~al.(2022)Xu, Zhu, Shi, Luo, and Song}]{xu2022instrumental}
Xu, Y., Zhu, J., Shi, C., Luo, S., and Song, R. (2022), \enquote{An Instrumental Variable Approach to Confounded Off-Policy Evaluation,} \textit{arXiv preprint arXiv:2212.14468}.

\bibitem[{Xu et~al.(2018)Xu, Li, Guan, Zhang, Li, Nan, Liu, Bian, and Ye}]{xu2018large}
Xu, Z., Li, Z., Guan, Q., Zhang, D., Li, Q., Nan, J., Liu, C., Bian, W., and Ye, J. (2018), \enquote{Large-scale order dispatch in on-demand ride-hailing platforms: A learning and planning approach,} in \textit{Proceedings of the 24th ACM SIGKDD international conference on knowledge discovery \& data mining}, pp. 905--913.

\bibitem[{Yan et~al.(2022{\natexlab{a}})Yan, Li, Chen, and Fan}]{yan2022efficacy}
Yan, Y., Li, G., Chen, Y., and Fan, J. (2022{\natexlab{a}}), \enquote{The efficacy of pessimism in asynchronous Q-learning,} \textit{arXiv preprint arXiv:2203.07368}.

\bibitem[{Yan et~al.(2022{\natexlab{b}})Yan, Li, Chen, and Fan}]{yan2022model}
--- (2022{\natexlab{b}}), \enquote{Model-based reinforcement learning is minimax-optimal for offline zero-sum markov games,} \textit{arXiv preprint arXiv:2206.04044}.

\bibitem[{Yang et~al.(2020)Yang, Jin, Wang, Wang, and Jordan}]{yang2020function}
Yang, Z., Jin, C., Wang, Z., Wang, M., and Jordan, M.~I. (2020), \enquote{On function approximation in reinforcement learning: Optimism in the face of large state spaces,} \textit{Advances in Neural Information Processing Systems}, 2020.

\bibitem[{Yu et~al.(2020)Yu, Thomas, Yu, Ermon, Zou, Levine, Finn, and Ma}]{yu2020mopo}
Yu, T., Thomas, G., Yu, L., Ermon, S., Zou, J.~Y., Levine, S., Finn, C., and Ma, T. (2020), \enquote{Mopo: Model-based offline policy optimization,} \textit{Advances in Neural Information Processing Systems}, 33, 14129--14142.

\bibitem[{Zhang et~al.(2013)Zhang, Tsiatis, Laber, and Davidian}]{zhang2013robust}
Zhang, B., Tsiatis, A.~A., Laber, E.~B., and Davidian, M. (2013), \enquote{Robust estimation of optimal dynamic treatment regimes for sequential treatment decisions,} \textit{Biometrika}, 100, 681--694.

\bibitem[{Zhang and Bareinboim(2016)}]{zhang2016markov}
Zhang, J. and Bareinboim, E. (2016), \enquote{Markov decision processes with unobserved confounders: A causal approach,} Tech. rep., Technical report, Technical Report R-23, Purdue AI Lab.

\bibitem[{Zhang et~al.(2018)Zhang, Laber, Davidian, and Tsiatis}]{zhang2018interpretable}
Zhang, Y., Laber, E.~B., Davidian, M., and Tsiatis, A.~A. (2018), \enquote{Interpretable dynamic treatment regimes,} \textit{Journal of the American Statistical Association}, 113, 1541--1549.

\bibitem[{Zhao et~al.(2015)Zhao, Zeng, Laber, and Kosorok}]{zhao2015new}
Zhao, Y.-Q., Zeng, D., Laber, E.~B., and Kosorok, M.~R. (2015), \enquote{New statistical learning methods for estimating optimal dynamic treatment regimes,} \textit{Journal of the American Statistical Association}, 110, 583--598.

\bibitem[{Zhou et~al.(2022)Zhou, Zhu, and Qu}]{zhou2022estimating}
Zhou, W., Zhu, R., and Qu, A. (2022), \enquote{Estimating optimal infinite horizon dynamic treatment regimes via pt-learning,} \textit{Journal of the American Statistical Association}, accepted.

\bibitem[{Zhou et~al.(2023)Zhou, Qi, Shi, and Li}]{zhou2023optimizing}
Zhou, Y., Qi, Z., Shi, C., and Li, L. (2023), \enquote{Optimizing Pessimism in Dynamic Treatment Regimes: A Bayesian Learning Approach,} in \textit{International Conference on Artificial Intelligence and Statistics}, PMLR, pp. 6704--6721.

\bibitem[{Zhu et~al.(2023)Zhu, Wan, Qi, Luo, and Shi}]{zhu2023robust}
Zhu, J., Wan, R., Qi, Z., Luo, S., and Shi, C. (2023), \enquote{Robust Offline Policy Evaluation and Optimization with Heavy-Tailed Rewards,} \textit{arXiv preprint arXiv:2310.18715}.

\end{thebibliography}

\newpage
\baselineskip=24pt
\setcounter{page}{1}
\setcounter{equation}{0}
\setcounter{section}{0}
\renewcommand{\thesection}{S.\arabic{section}}
\renewcommand{\thelemma}{S\arabic{lemma}}
\renewcommand{\theequation}{S\arabic{equation}}

\begin{center}
{\Large\bf Supplementary Materials for} \\
\medskip
{\Large\bf ``Pessimistic Causal Reinforcement Learning with Mediators for Confounded Offline Data"}  \\
\bigskip
\textbf{Danyang Wang, Chengchun Shi, Shikai Luo, Will Wei Sun}
\vspace{0.2in}
\end{center}
\bigskip

\noindent

In the supplementary, we provide detailed proofs of major theoretical results, including Theorem \ref{thm1}, Theorem \ref{thm:converge}, and Theorem \ref{thm:pess/pessm}.

\section{Proof of Theorem \ref{thm1}}\label{proof:val_iter}

The proof incorporates the Banach fixed-point theorem \citep{banach1922operations}, a technique widely used in the RL literature. For completeness, we first state the Banach fixed-point theorem, along with the definition of contraction mapping.  

\begin{definition}[Contraction mapping \citep{nadler1969multi}]
Let $(X,d)$ be a complete metric space, then a map $T:X\rightarrow X$ is called a contraction mapping on $X$ if there exists $q\in [0,1)$ such that $d(Tx,Ty)\leq qd(x,y)$ for all $x,y\in X$.
\end{definition}

\begin{lemma}[Banach fixed point theorem \citep{banach1922operations}]\label{thm:banach}
    Let $(X,d)$ be a non-empty complete metric space with a contraction mapping $T:X\rightarrow X$. Then $T$ admits a unique fixed-point $x^*$ in $X$ (e.g. $Tx^*=x^*$). Furthermore, $x^*$ can be found as follows: start with an arbitrary element $x_0\in X$ and define a sequence $(x_n)_{n\in\N}$ by $x_n=Tx_{n-1}$ for $n\geq 1$, then $\underset{n\rightarrow \infty}{\lim} x_n = x^*$.
\end{lemma}

We first show that when the Q-function is updated iteratively according to the Bellman optimality operator, it converges to the optimal Q-function $Q^*$. According to Lemma \ref{thm:banach}, it is sufficient to show that $\bello$ is a contraction mapping under the supremum norm:
\[
\|Q\|_{\infty}:=\sup \{|Q(s,a,m)|:s\in \cS, a\in \cA, m\in \cM\}.
\]

\begin{lemma}[Bellman Contraction Mapping]\label{thm:bell contrac}
The Bellman optimality operator $\bello$ is a contraction mapping. In particular, for any functions $Q_1$ and $Q_2$, we have
\begin{align*}
    \| \bello Q_1 - \bello Q_2 \|_\infty \leq \gamma\|Q_1 - Q_2  \|_\infty.
\end{align*}
\end{lemma}

\begin{proof}[Proof of Lemma \ref{thm:bell contrac}]
By definition, 
\begin{align*}
    &\| \bello Q_1 - \bello Q_2\|_\infty=\sup_{(s,a,m)}\left|\bello Q_1(s,a,m) - \bello Q_2(s,a,m)\right|\\
    &=\sup_{(s,a,m)}\Big|r(s,a,m)+\gamma \max_{a'}\sum_{s',\tilde{a}',m'}Q_1(s',\tilde{a}',m')p_s^*(s'|s,a,m)p_b(\tilde{a}'|s')p_m(m'|s',a') \\
    &-r(s,a,m)+\gamma \max_{a'}\sum_{s',\tilde{a}',m'}Q_2(s',\tilde{a}',m')p_s^*(s'|s,a,m)p_b(\tilde{a}'|s')p_m(m'|s',a') \Big|\\
    &=\gamma \sup_{(s,a,m)}\Big|\max_{a'} \sum_{s',\tilde{a}',m'}Q_1(s',\tilde{a}',m')p_s^*(s'|s,a,m)p_b(\tilde{a}'|s')p_m(m'|s',a')\\
    &-\max_{a'} Q_2(s',\tilde{a}',m')p_s^*(s'|s,a,m)p_b(\tilde{a}'|s')p_m(m'|s',a') \Big|\\
    &\le \gamma \sup_{(s,a,m,a')} \Big|\sum_{s',\tilde{a}',m'}[Q_1(s',\tilde{a}',m')-Q_2(s',\tilde{a}',m')]p_s^*(s'|s,a,m)p_b(\tilde{a}'|s')p_m(m'|s',a')\Big|\\
    &\le \gamma \sup_{(s',\tilde{a}',m')} \Big|Q_1(s',\tilde{a}',m')-Q_2(s',\tilde{a}',m')\Big|= \gamma \|Q_1 - Q_2\|_\infty,
\end{align*}
where $p_s^*$ denotes the probability density/mass function of $S_{t+1}$ given $S_t,A_t$, $M_t$, $r$ denotes the conditional mean of $R_t$ given $S_t,A_t$, $M_t$, and the first inequality is due to that $\underset{x}{\max}|f(x)-g(x)|\geq|\underset{x}{\max} ~f(x)-\underset{x}{\max}~ g(x)|$ for any $f$ and $g$. This complete the proof of Lemma \ref{thm:bell contrac}.
\end{proof}

Next, we show that the greedy policy with respect to $Q^*$ defined in \eqref{eqn:Belloptimal} is indeed the optimal policy. This proves (i) and (ii). A key observation is that, the online M2DP process can be converted into a standard MDP as follows. Starting from a given state $s$ and action $a$, the expected immediate reward and the state transition are given by $\sum_{\tilde{a},m} r(s,\tilde{a},m)p(m|s,a)p_b(\tilde{a}|s)$ and $\sum_{\tilde{a},m} p_s^*(s'|s,\tilde{a},m)p_m(m|s,a)p_b(\tilde{a}|s)$, respectively. The optimal Q-function under this MDP is given by
\begin{eqnarray*}
    \sum_{\tilde{a},m} Q^*(s,\tilde{a},m)p_m(m|s,a)p_b(\tilde{a}|s).
\end{eqnarray*}
In classical MDPs, it has been shown that the greedy policy with respect to the optimal Q-function is no worse than any other policy \citep[see e.g.,][]{puterman2014markov}. This proves the optimality of $\pi^*$ in \eqref{eqn:optimalpolicy}. The proof is hence completed.

\section{Proof of Theorem \ref{thm:converge}}\label{proof:convergence}
We focus on providing a general version of Theorem \ref{thm:converge} without the completeness assumption. In other words, we allow $\varepsilon_{\cQ}^2:=\inf_{Q_1\in \cQ}\sup_{Q_2\in \cQ} \|Q_1-\mathcal{B}^* Q_2\|_{\mu}^2>0$. Let $c_4=p_{m^{-}}^{-1}$. Where $p_{m^{-}}$ is the assigned fixed value by the M2DP process for $\forall~(s,a,m)$, when $p_m(m|a,s)\le p_{m^-}$. Specifically, it is sufficiently small, for example, $1\times10^{-5}$.
\begin{theorem}\label{thm:thm2full}
Under Assumptions \ref{asm:frontdoor_crit}, \ref{assump:coverage}(i),(ii), \ref{assump:mixing} and \ref{asm:finite}, given a dataset $\mathcal{D}$ of size $N$, for $C>0$, the learned policy at the $K$th iteration satisfies 
\begin{align*}
\E [J(\pi^*)-J(\hat{\pi}_K)]\le \frac{C\sqrt{c_1}\gamma^K V_{\max}}{1-\gamma}+\frac{C\sqrt{c_1}}{(1-\gamma)^2}\Big[\varepsilon_{\cQ}+N^{-1/2}V_{\max}\sqrt{12\lambda+(97\lambda-48)\ln(2\delta^{-1}|\cQ|^2|\cP_b||\cP_m|)}\\
    +V_{\max} \Big(\sqrt{\E_{S\sim \rho^{\cD},A\sim \hat{p}_b}\textrm{KL}(p_b(\bullet|S)\|\hat{p}_b(\bullet|S))}+\sqrt{c_2\E_{S\sim \rho^{\cD},A\sim p_b,M\sim \hat{p}_m}\textrm{KL}(p_m(\bullet|S,A)\|\hat{p}_m(\bullet|S,A))}\Big)\Big].
\end{align*}
\end{theorem}

We first provide an overview of the proof of Theorem \ref{thm:thm2full}. The proof is divided into four steps. To begin with, for any mediated Q-function $Q$, define the (unmediated) Q-function $q$ and its estimator denoted by $\bar{q}$ (by replacing the oracle $p_m$ and $p_b$ with their estimators)
\begin{eqnarray*}
    q(s,a)&=&\sum_{m,\tilde{a}} Q(s,\tilde{a},m)p_m(m|s,a)p_b(\tilde{a}|s),\\
    \bar{q}(s,a)&=&\sum_{m,\tilde{a}} Q(s,\tilde{a},m)\hat{p}_m(m|s,a)\hat{p}_b(\tilde{a}|s).
\end{eqnarray*}
In the first step, 
we bound the approximation error of $\bar{q}$ from the estimation of $p_b$ and $p_m$ in Lemma \ref{lem:totalvar}. The second step is to apply the performance difference lemma \citep[Lemma 6.1,][]{kakade2002approximately} and Lemma \ref{lem:totalvar} to upper the difference $J(\pi^*)-J(\hat{\pi}_K)$ as a function of (i) the KL divergence between $p_b$, $p_m$ and their estimators as well as (ii) the estimation error of $\hat{Q}^{(K)}$. In the third step, we apply the Bernstein's inequality \citep{bernstein1946theory} to upper bound the estimation error of $\hat{Q}^{(K)}$. Finally, we summarize all the results in the last step.

\noindent\textbf{\textit{Step 1.}} We present Lemma \ref{lem:totalvar} and its proof below. 
\begin{lemma}\label{lem:totalvar} For any mediated Q-function $Q$ whose absolute value is upper bounded by $V_{\max}$, we have
$$|\bar{q}(s,a)-q(s,a)|\leq V_{\max}[\sqrt{2\textrm{KL}(p_b(\bullet|s)\|\hat{p}_b(\bullet|s))}+\sqrt{2\textrm{KL}(p_m(\bullet|s,a)\|\hat{p}_m(\bullet|s,a))}].$$
\end{lemma}
\begin{proof}[Proof of Lemma \ref{lem:totalvar}]
By definition,
\begin{align*}
    \bar{q}(s,a)-q(s,a)=&\sum_{m,\tilde{a}} Q(s,\tilde{a},m)[\widehat{p}_m(m|s,a)\widehat{p}_b(\tilde{a}|s)-p_m(m|s,a)p_b(\tilde{a}|s)]\\
    =&\sum_{m,\tilde{a}}Q(s,\tilde{a},m)\hat{p}_m(m|s,a)[\hat{p}_b(\tilde{a}|s)-p_b(\tilde{a}|s)]\\
    +&\sum_{m,\tilde{a}}Q(s,\tilde{a},m)[\hat{p}_m(m|s,a)-p_m(m|s,a)]p_b(\tilde{a}|s)\\
    \leq &~2V_{\max}[\textrm{TV}(p_b(\bullet|s)\|\hat{p}_b(\bullet|s))+\textrm{TV}(p_m(\bullet|s,a)\|\hat{p}_m(\bullet|s,a))],
\end{align*} 
where TV$(p_b(\bullet|s)\|\hat{p}_b(\bullet|s))$ denotes the total variation distance between $p_b$ and $\hat{p}_b$ conditional on $S=s$, and TV$(p_m(\bullet|s,a)\|\hat{p}_m(\bullet|s,a))$ denotes the total variation distance between $p_m$ and $\hat{p}_m$ conditional on $S=s$, $A=a$. According to Pinsker's inequality, we obtain that
\begin{eqnarray*}
    \bar{q}(s,a)-q(s,a)\le 2V_{\max}[\sqrt{0.5\textrm{KL}(p_b(\bullet|s)\|\hat{p}_b(\bullet|s))}+\sqrt{0.5\textrm{KL}(p_m(\bullet|s,a)\|\hat{p}_m(\bullet|s,a))}].
\end{eqnarray*}
By symmetry, we can similarly show that
\begin{align*}
    q(s,a)-\bar{q}(s,a)\le V_{\max}[\sqrt{2\textrm{KL}(p_b(\bullet|s)\|\hat{p}_b(\bullet|s))}+\sqrt{2\textrm{KL}(p_m(\bullet|s,a)\|\hat{p}_m(\bullet|s,a))}]. 
\end{align*}
The proof is hence completed. 
\end{proof}

\noindent\textbf{\textit{Step 2.}} We present our findings in Lemma \ref{lem:rlobj} below, along with its proof. 
\begin{lemma}\label{lem:rlobj}
For any mediated Q-functions $Q_1$, $Q_2$, any probability density/mass function $\rho$ and any deterministic policy $\pi$, let $\|Q_1-Q_2\|_{\rho\times p_b\times p_m^{\pi}}$ denote the distance
\begin{eqnarray*}
    \Big[\E_{S\sim \rho, A\sim p_b(\bullet|S), M\sim p_m(\bullet|S,\pi(S))}|Q_1(S,A,M)-Q_2(S,A,M)|^2\Big]^{1/2}. 
\end{eqnarray*}
We have
\begin{align*}
    J(\pi^*) - J(\hat\pi_K) \leq \frac{1}{1-\gamma} \Big[\|Q^*-\hat{Q}^{(K)}\|_{\rho^{\hat{\pi}_K}\times p_b\times p_m^{\pi^*}}+\|Q^*-\hat{Q}^{(K)}\|_{\rho^{\hat{\pi}_K}\times p_b\times p_m^{\hat{\pi}_K}}\Big]\\
    +\frac{2V_{\max}}{1-\gamma} \E_{S\sim \rho^{\hat{\pi}_K}} \Big[\sqrt{2\textrm{KL}(p_b(\bullet|S)\|\hat{p}_b(\bullet|S))}+\sqrt{0.5\textrm{KL}(p_m(\bullet|S,\pi^*(S))\|\hat{p}_m(\bullet|S,\pi^*(S)))}\\
    +\sqrt{0.5\textrm{KL}(p_m(\bullet|S,\hat{\pi}_K(S))\|\hat{p}_m(\bullet|S,\hat{\pi}_K(S)))}\Big].
\end{align*}
\end{lemma}

\begin{proof}[Proof of Lemma \ref{lem:rlobj}]
Recall that $\rho^{\pi}$ denotes the $\gamma$-discounted visitation probability under $\pi$. According to the performance difference lemma (see e.g., Lemma 6.1 in \cite{kakade2002approximately}), we obtain that
\begin{eqnarray*}
    J(\pi^*) - J(\hat\pi_K)=\frac{1}{1-\gamma}\E_{S\sim \rho^{\hat{\pi}_K}} \left[q^*(S,\pi^*(S)) - q^*(S,\hat{\pi}_K(S))\right].
\end{eqnarray*}
It follows that
\begin{eqnarray*}
    J(\pi^*) - J(\hat\pi_K)\le \frac{1}{1-\gamma}\E_{S\sim \rho^{\hat{\pi}_K}} \left[q^*(S,\pi^*(S))- \bar{q}^{(K)}(S,\pi^*(S))-q^*(S,\hat{\pi}_K(S))+\bar{q}^{(K)}(S,\hat{\pi}_K(S))\right]\\
    \le \frac{1}{1-\gamma}\E_{S\sim \rho^{\hat{\pi}_K}} \left[q^*(S,\pi^*(S))- \hat{q}^{(K)}(S,\pi^*(S))-q^*(S,\hat{\pi}_K(S))+\hat{q}^{(K)}(S,\hat{\pi}_K(S))\right]\\
    +\frac{2V_{\max}}{1-\gamma} \E_{S\sim \rho^{\hat{\pi}_K}} \Big[\sqrt{2\textrm{KL}(p_b(\bullet|S)\|\hat{p}_b(\bullet|S))}+\sqrt{0.5\textrm{KL}(p_m(\bullet|S,\pi^*(S))\|\hat{p}_m(\bullet|S,\pi^*(S)))}\\
    +\sqrt{0.5\textrm{KL}(p_m(\bullet|S,\hat{\pi}_K(S))\|\hat{p}_m(\bullet|S,\hat{\pi}_K(S)))}\Big],
\end{eqnarray*}
where 
\begin{align*}
    \bar{q}^{(K)}(s,a)&=\sum_{\tilde{a},m}\hat{Q}^{(K)}(s,\tilde{a},m) \hat{p}_b(\tilde{a}|s)\hat{p}_m(m|s,a),\\
    \hat{q}^{(K)}(s,a)&=\sum_{\tilde{a},m}\hat{Q}^{(K)}(s,\tilde{a},m) p_b(\tilde{a}|s)p_m(m|s,a),
\end{align*}
and the first inequality is due to that $\bar{q}^{(K)}(s,a)\le \bar{q}^{(K)}(s,\widehat{\pi}_K(s))$ for any $s$, $a$ and the second inequality is due to Lemma \ref{lem:totalvar}.

Next, according to Cauchy-Schwarz inequality, we obtain that
\begin{eqnarray*}
    J(\pi^*)-J(\widehat{\pi}_K)\le \frac{1}{1-\gamma} \Big[\|Q^*-\hat{Q}^{(K)}\|_{\rho^{\hat{\pi}_K}\times p_b\times p_m^{\pi^*}}+\|Q^*-\hat{Q}^{(K)}\|_{\rho^{\hat{\pi}_K}\times p_b\times p_m^{\hat{\pi}_K}}\Big]\\
    +\frac{2V_{\max}}{1-\gamma} \E_{S\sim \rho^{\hat{\pi}_K}} \Big[\sqrt{2\textrm{KL}(p_b(\bullet|S)\|\hat{p}_b(\bullet|S))}+\sqrt{0.5\textrm{KL}(p_m(\bullet|S,\pi^*(S))\|\hat{p}_m(\bullet|S,\pi^*(S)))}\\
    +\sqrt{0.5\textrm{KL}(p_m(\bullet|S,\hat{\pi}_K(S))\|\hat{p}_m(\bullet|S,\hat{\pi}_K(S)))}\Big].
\end{eqnarray*}
This completes the proof.
\end{proof}

\noindent\textbf{\textit{Step 3.}} Based on the results in the first two steps, it remains to bound the estimation error between $Q^*$ and $\hat{Q}^{(K)}$ as well as the expected KL divergences. We focus on the estimation error in this step. 
Under the coverage assumptions in A4(i) and (ii), it follows from the change of measure theorem that both $\|Q^*-\hat{Q}^{(K)}\|_{\rho^{\hat{\pi}_K}\times p_b\times p_m^{\hat{\pi}_K}}$ and $\|Q^*-\hat{Q}^{(K)}\|_{\rho^{\hat{\pi}_K}\times p_b\times p_m^{\pi^*}}$ can be upper bounded by 
\begin{eqnarray*}
    \sqrt{c_1c_4}\|Q^*-\hat{Q}^{(K)}\|_{\rho^{\mathcal{D}}\times p_b\times p_m}:=\sqrt{c_1c_4} \Big[ \E_{S\sim \rho^{\cD},A\sim p_b(\bullet|S),M\sim p_m(\bullet|S,A)} |Q^*(S,A,M)-\hat{Q}^{(K)}(S,A,M)|^2 \Big]^{1/2}.
\end{eqnarray*}
Meanwhile, it follows from Lemma 15 and the proof of Theorem 11 of \cite{chen2019information} that, by rollout back recursively, $\|Q^*-\hat{Q}^{(K)}\|_{\rho^{\mathcal{D}}\times p_b\times p_m}$ can be upper bounded by
\begin{eqnarray}\label{eqn:bound1}
    \frac{1}{1-\gamma}\max_k \|\hat{Q}^{(k)}-\bello\hat{Q}^{(k-1)}\|_{\rho^{\mathcal{D}}\times p_b\times p_m}+\gamma^K V_{\max}. 
\end{eqnarray}
The following lemma provides an upper bound for $\max_k \|\hat{Q}^{(k)}-\bello\hat{Q}^{(k-1)}\|_{\rho^{\mathcal{D}}\times p_b\times p_m}$.
\begin{lemma}\label{lem:errorbound}
Let $Q^{(k)*}$ denote the best in-class Q-function that minimizes $\argmin_{Q\in \cQ} \|Q-\bello\hat{Q}^{(k-1)}\|_{\rho^{\mathcal{D}}\times p_b\times p_m}$ and $\varepsilon^{(k)}$ denote the excess squared loss $$\|\hat{Q}^{(k)}-\bello\hat{Q}^{(k-1)}\|^2_{\rho^{\mathcal{D}}\times p_b\times p_m}-\|Q^{(k)*}-\bello\hat{Q}^{(k-1)}\|^2_{\rho^{\mathcal{D}}\times p_b\times p_m}.$$  
For any $\delta>0$, with probability at least $1-\delta$, we have that
\begin{eqnarray*}
    \varepsilon^{(k)} &\le& 10\varepsilon_{\cQ}^2+ 16N^{-1}\left[(97\lambda-48)V_{\max}^2\ln(2\delta^{-1}|\cQ|^2|\cP_b||\cP_m|)+12\lambda V_{\max}^2\right] \\
    && +144\gamma^2 V_{\max}^2 \E_{S\sim \rho^{\cD},A\sim p_b}\Big[\textrm{KL}(p_b(\bullet|S)\|\hat{p}_b(\bullet|S)) +c_2 \textrm{KL}(p_m(\bullet|S,A)\|\hat{p}_m(\bullet|S,A))\Big].
\end{eqnarray*}
\end{lemma}
\begin{proof}[Proof of Lemma \ref{lem:errorbound}]
We decompose $\varepsilon^{(k)}$ into three terms $\varepsilon_1^{(k)}+\varepsilon_2^{(k)}+\varepsilon_3^{(k)}$ given by
\begin{eqnarray*}
    \varepsilon_1^{(k)}&=&\|\hat{Q}^{(k)}-\bello_{\hat{p}_b,\hat{p}_m}\hat{Q}^{(k-1)}\|^2_{\rho^{\mathcal{D}}\times p_b\times p_m}-\|Q^{(k)*}-\bello_{\hat{p}_b,\hat{p}_m}\hat{Q}^{(k-1)}\|^2_{\rho^{\mathcal{D}}\times p_b\times p_m}\\
    &-&\Big[\|\hat{Q}^{(k)}-\bello_{\hat{p}_b,\hat{p}_m}\hat{Q}^{(k-1)}\|^2_{\mathcal{D}}-\|Q^{(k)*}-\bello_{\hat{p}_b,\hat{p}_m}\hat{Q}^{(k-1)}\|^2_{\mathcal{D}}\Big],\\
    \varepsilon_2^{(k)}&=&\|\hat{Q}^{(k)}-\bello\hat{Q}^{(k-1)}\|^2_{\rho^{\mathcal{D}}\times p_b\times p_m}-\|Q^{(k)*}-\bello\hat{Q}^{(k-1)}\|^2_{\rho^{\mathcal{D}}\times p_b\times p_m}\\
    &-&\Big[\|\hat{Q}^{(k)}-\bello_{\hat{p}_b,\hat{p}_m}\hat{Q}^{(k-1)}\|^2_{\rho^{\mathcal{D}}\times p_b\times p_m}-\|Q^{(k)*}-\bello_{\hat{p}_b,\hat{p}_m}\hat{Q}^{(k-1)}\|^2_{\rho^{\mathcal{D}}\times p_b\times p_m}\Big],\\
    \varepsilon_3^{(k)}&=&\|\hat{Q}^{(k)}-\bello_{\hat{p}_b,\hat{p}_m}\hat{Q}^{(k-1)}\|^2_{\mathcal{D}}-\|Q^{(k)*}-\bello_{\hat{p}_b,\hat{p}_m}\hat{Q}^{(k-1)}\|^2_{\mathcal{D}},
\end{eqnarray*}
respectively, where $\|\bullet\|^2_{\cD}$ denotes the empirical least square loss aggregated over the data tuples in $\cD$, and $\bello_{q_b,q_m}$ denotes a version of the Bellman operator $\bello$ with $p_m$ and $p_b$ replaced by some $q_m$ and $q_b$, e.g., $\bello_{q_b,q_m} Q(s,a,m)$ equals
\begin{eqnarray*}
    \E \Big[R_t+\gamma \max_{a} \sum_{\tilde{a},m} q_m(m|S_{t+1},a) q_b(\tilde{a}|S_{t+1})Q(S_{t+1},\tilde{a},m)|S_t=s,A_t=a,M_t=m\Big].
\end{eqnarray*}

Since $Q^{(k)*}\in \cQ$, it is immediate to see that $\varepsilon_3\le 0$. It remains to upper bound $\varepsilon_1$ and $\varepsilon_2$. Consider $\varepsilon_1$ first. A key observation is that, $\varepsilon_1$ can be upper bounded by the following empirical process
\begin{eqnarray}\label{eqn:emp}
    \sup_{\substack{Q_1,Q_2\in \cQ\\q_b\in \cP_b,q_m\in\cP_m}} \frac{1}{N}\sum_{i=1}^N \psi(S_{(i)},A_{(i)},M_{(i)},S_{(i)}';Q_1,Q_2,q_b,q_m)-\E[ \psi(S,A,M,S';Q_1,Q_2,q_b,q_m)],
\end{eqnarray}
where $\psi(s,a,m,s';Q,q_b,q_m)$ is a shorthand for 
\begin{eqnarray}\nonumber
    &&(Q^{(k)*}(s,a,m)-\bello_{q_b,q_m}Q_2(s,a,m))^2-(Q_1(s,a,m)-\bello_{q_b,q_m}Q_2(s,a,m))^2\\ \label{eqn:psi}
    &=&[Q^{(k)*}(s,a,m)-Q_1(s,a,m)][Q_1(s,a,m)+Q^{(k)*}(s,a,m)-2\bello_{q_b,q_m}Q_2(s,a,m)].
\end{eqnarray}

Denote $\psi(S_{(i)},A_{(i)},M_{(i)},S_{(i)}';Q_1,Q_2,q_b,q_m)$ as $\psi_i$, and $\psi(S,A,M,S';Q_1,Q_2,q_b,q_m)$ as $\psi$. For a dataset $\cD$ collected by $K$ trajectories, each with horizon length $H_i$, $1\leq i\leq K$. We can flatten the trajectories, and write the data tuples $d\in \cD$ as $\cD=\{d_{1, 0}, \dots d_{1, H_1}, \dots d_{K, 0}, \dots d_{K, H_K}\}$, which has length $N$, where the first subscript denotes trajectory, and second subscript denotes which step in each trajectory it is collected. Based on this expression, we can rearrange the index of $\psi_i$ in the same order, which leads to the sequence $\Psi=(\psi_1, \psi_2, \dots, \psi_N)$. We partition $\Psi$ into subsequences of length $\lambda$. Let $U_i=(\psi_{i\lambda+1}, \dots, \psi_{i\lambda+\lambda})$, then the original sequence consists of $\Psi=(U_0, \dots, U_{\floor{\frac{N}{\lambda}}-1}, \psi_{\floor{\frac{N}{\lambda}}\lambda+1}, \dots, \psi_N)$. Under Assumption \ref{assump:mixing}, according to Lemma 3 in \cite{shi2022multi} and Lemma 4.1 in \cite{dedecker2002maximal}, we can construct random variable $\tilde{U}_i=(\tilde{\psi}_{i\lambda+1}, \dots, \tilde{\psi}_{i\lambda+\lambda})$, that has same distribution as $U_i$ with probability at lease $1-\beta(\lambda)$, and $\tilde{\Psi}=(\tilde{U}_0, \dots, \tilde{U}_{\floor{\frac{N}{\lambda}}-1}, \psi_{\floor{\frac{N}{\lambda}}\lambda+1}, \dots, \psi_N)$, such that $\tilde{\Psi}$ has same distribution as $\Psi$ with probability at least $1-\floor{\frac{N}{\lambda}}\beta(\lambda)$, where $\{\tilde{U}_{2i}\}_{i\geq0}$ are i.i.d., and $\{\tilde{U}_{2i+1}\}_{i\geq0}$ are also i.i.d.. Denote $\omega_i=\tilde{\psi}_{i\lambda+1}+\dots +\tilde{\psi}_{i\lambda+\lambda}$, which is the sum of all elements in $\tilde{U}_i$. Then $\{\omega_{2i}\}_{i\geq0}$ and $\{\omega_{2i+1}\}_{i\geq0}$ are also i.i.d., respectively. Under boundedness assumption of $\cQ$, $|\tilde{\psi}_i|\leq4V_{\max}^2$, therefore, $|\omega_i|\leq 4\lambda V_{\max}^2$. According to \eqref{eqn:psi}, variance of $\psi_i$ can be upper bounded by
\begin{eqnarray}\label{eqn:variancebound}
\begin{split}
    &\Var(\psi_i)\le \E[(Q_1-Q^{(k)*})^2(Q_1+Q^{(k)*}-2\cB_{q_b,q_m}^*Q_2)^2]\\
    \leq&16V_{\max}^2 \|Q_1-Q^{(k)*}\|^2_{\rho^{\cD}\times p_b\times p_m} \\
    \le& 32V_{\max}^2 \Big[\|Q_1-\bello Q_2\|_{\rho^{\cD}\times p_b\times p_m}^2+
   \|Q^{(k)*}-\bello Q_2\|_{\rho^{\cD}\times p_b\times p_m}^2\Big] \\
   =&32V_{\max}^2 \Big[\|Q_1-\bello Q_2\|_{\rho^{\cD}\times p_b\times p_m}^2-\|Q^{(k)*}-\bello Q_2\|_{\rho^{\cD}\times p_b\times p_m}^2+
   2\|Q^{(k)*}-\bello Q_2\|_{\rho^{\cD}\times p_b\times p_m}^2\Big] \\
    \le& 32V_{\max}^2 \varepsilon^{(k)}+64V_{\max}^2\varepsilon_{\cQ}^2
   :=\sigma_{\psi}^2.
\end{split}
\end{eqnarray}
The last inequality is based on the definition of $\varepsilon^{(k)}$ and $ \varepsilon_{\cQ}^2$. Besides, for $i,j\in[1,N]$, we denote $(Q_1-Q^{(k)*})(S_{(i)},A_{(i)},M_{(i)})$ as $Q_{1,i}-Q^{(k)*}_i$, following same prodecure as in \ref{eqn:variancebound}, 
\begin{align*}
    &cov(\psi_i, \psi_j)\le 16V_{\max}^2 \E\left[(Q_{1,i}-Q^{(k)*}_i)(Q_{1,j}-Q^{(k)*}_j)\right]\\
    \le& 16V_{\max}^2\sqrt{\E[(Q_{1,i}-Q^{(k)*}_i)^2]\E[(Q_{1,j}-Q^{(k)*}_j)^2]}\\
    \le& 32V_{\max}^2 \varepsilon^{(k)}+64V_{\max}^2\varepsilon_{\cQ}^2 =\sigma_{\psi}^2.
\end{align*}
Therefore, for a $\omega_i$ with fixed $i$, e.g., $i=0$, 
\begin{align*}
    &\Var(\omega_0)=\sum_{i=1}^\lambda \Var(\tilde{\psi}_i)+2\sum_{1\le i<j\le \lambda}cov(\tilde{\psi}_i,\tilde{\psi}_j)\\
    \le& \lambda \sigma_{\psi}^2+2\lambda(\lambda-1)\sigma_{\psi}^2\\
    =&\lambda(2\lambda-1)\sigma_{\psi}^2\\
    =&\lambda(2\lambda-1)(32V_{\max}^2 \varepsilon^{(k)}+64V_{\max}^2\varepsilon_{\cQ}^2):=\sigma^2.
\end{align*}
The approximation of $\tilde{\psi_i}$ follows from the fact that $\tilde{U}_i$ and $U_i$ are identically distributed given a probability lower bound. Note that this upper bound for variance applies to all $\omega_i$.

Since all the function classes are finite hypothesis classes, we apply the Bernstein's inequality \citep{bernstein1946theory} to upper bound $\omega_i$, which is then further applied in bounding \eqref{eqn:emp}. Let number of $\omega_i$'s where $i$ is odd be $n_1$, and number of $\omega_i$'s where $i$ is even be $n_2$, then $n_1+n_2=\floor{\frac{N}{\lambda}}$. It follows from \cite{bernstein1946theory} that for $\epsilon_1, \epsilon_2 \geq 0$,
\begin{align*}
    \mathbb{P}\left(\sum_{i=0}^{n_1-1} \left[\omega_{2i+1} - \E(\omega_{2i+1})\right]>\epsilon_1\right)&\le \exp\left(\frac{-\epsilon_1^2}{2(n_1\sigma^2+4\lambda V^2_{\max}\epsilon_1/3)}\right),\\
    \text{and }\mathbb{P}\left(\sum_{i=0}^{n_2-1} \left[\omega_{2i} - \E(\omega_{2i})\right]>\epsilon_2\right)&\le \exp\left(\frac{-\epsilon_2^2}{2(n_2\sigma^2+4\lambda V^2_{\max}\epsilon_2/3)}\right).
\end{align*}
Applying union bound on $\cQ, \cP_b, \cP_m$, and by definiton of $\omega_i$, we have that for $\delta_1,\delta_2\ge 0$, with probability at least $1-\left(\delta_1+\delta_2+\floor{\frac{N}{\lambda}}\beta(\lambda)\right)$,
\begin{align*}
    &\sum_{i=1}^{\floor{\frac{N}{\lambda}}\lambda}\psi_i-\floor{\frac{N}{\lambda}}\lambda\E(\psi) \\
    \leq & \frac{4\lambda V_{\max}^2}{3}\left[\ln\left(\frac{|\cQ|^2|\cP_b||\cP_m|}{\delta_1}\right)+\ln\left(\frac{|\cQ|^2|\cP_b||\cP_m|}{\delta_2}\right)\right]
\\
&+\sqrt{2n_1\sigma^2\ln\left(\frac{|\cQ|^2|\cP_b||\cP_m|}{\delta_1}\right)}+\sqrt{2n_2\sigma^2\ln\left(\frac{|\cQ|^2|\cP_b||\cP_m|}{\delta_2}\right)}.
\end{align*}
Similarly, with probability at least $1-\left(\delta_1+\delta_2+\floor{\frac{N}{\lambda}}\beta(\lambda)\right)$, \eqref{eqn:emp} can be bounded as follows:
\begin{align*}
    &\frac{1}{N}\sum_{i=1}^N \psi_i-\E(\psi)\\ 
    =&\frac{\floor{\frac{N}{\lambda}}\lambda}{N}\frac{1}{\floor{\frac{N}{\lambda}}\lambda}\left(\sum_{i=1}^{\floor{\frac{N}{\lambda}}\lambda}\psi_i-\floor{\frac{N}{\lambda}}\lambda\E(\psi)\right)+ \frac{1}{N}\left(\sum_{i=\floor{\frac{N}{\lambda}}+1}^N\psi_i\right)\\
    \le &\frac{4\lambda V_{\max}^2}{3N}\left[\ln\left(\frac{|\cQ|^2|\cP_b||\cP_m|}{\delta_1}\right)+\ln\left(\frac{|\cQ|^2|\cP_b||\cP_m|}{\delta_2}\right)\right]
\\
&+\frac{1}{N}\left[\sqrt{2n_1\sigma^2\ln\left(\frac{|\cQ|^2|\cP_b||\cP_m|}{\delta_1}\right)}+\sqrt{2n_2\sigma^2\ln\left(\frac{|\cQ|^2|\cP_b||\cP_m|}{\delta_2}\right)}\right]+\frac{4\lambda V_{\max}^2}{N}\\
\le&\frac{4\lambda V_{\max}^2}{3N}\left[\ln\left(\frac{|\cQ|^2|\cP_b||\cP_m|}{\delta_1}\right)+\ln\left(\frac{|\cQ|^2|\cP_b||\cP_m|}{\delta_2}\right)\right]
\tag{*}\\
&+\sqrt{\frac{128(2\lambda-1)V_{\max}^2(\varepsilon^{(k)}+2\varepsilon_{\cQ}^2)}{N}\left[\ln\left(\frac{|\cQ|^2|\cP_b||\cP_m|}{\delta_1}\right)+\ln\left(\frac{|\cQ|^2|\cP_b||\cP_m|}{\delta_2}\right)\right]}+\frac{4\lambda V_{\max}^2}{N}.
\end{align*}

Where (*) follows from $n_1,n_2\le \frac{N}{\lambda}$ and $\sqrt{a}+\sqrt{b}\le \sqrt{2(a+b)}$. Without loss of generality, we let $\delta_1=\delta_2=\frac{\delta}{2}$. Then
\begin{align*}
    (*)=&\sqrt{\frac{256(2\lambda-1)V_{\max}^2(\varepsilon^{(k)}+2\varepsilon_{\cQ}^2)\ln(2\delta^{-1}|\cQ|^2|\cP_b||\cP_m|)}{N}}+\frac{8\lambda V_{\max}^2\ln(2\delta^{-1}|\cQ|^2|\cP_b||\cP_m|)+12\lambda V_{\max}^2}{3N}\\
    \le& \frac{\varepsilon^{(k)}}{2}+\varepsilon_{\cQ}^2+\frac{(776\lambda-384)V_{\max}^2\ln(2\delta^{-1}|\cQ|^2|\cP_b||\cP_m|)+12\lambda V_{\max}^2}{3N}.
\end{align*}
Where the last inequality follows from $\sqrt{ab}\le \frac{a+b}{2}$. Using Bonferroni's inequality, we obtain that the above inequality holds with probability at least $1-\left(\delta+\floor{\frac{N}{\lambda}}\beta(\lambda)\right)$ uniformly for any $Q_1,Q_2\in \cQ$, $q_b\in\cP_b$ and $q_m\in \cP_m$. This yields that with probability at least $1-\left(\delta+\floor{\frac{N}{\lambda}}\beta(\lambda)\right)$,
\begin{eqnarray}\label{eqn:varepsilon1k}
    \varepsilon_1^{(k)}\le \frac{\varepsilon^{(k)}}{2}+\varepsilon_{\cQ}^2+\frac{(776\lambda-384)V_{\max}^2\ln(2\delta^{-1}|\cQ|^2|\cP_b||\cP_m|)+12\lambda V_{\max}^2}{3N}.
\end{eqnarray}

\noindent Next, consider $\varepsilon_2^{(k)}$. We have
\begin{eqnarray*}
    \varepsilon_2^{(k)}=2\E_{\substack{S\sim \rho^{\cD},A\sim p_b\\ M\sim p_m}} [\hat{Q}^{(k)}(S,A,M)-Q^{(k)*}(S,A,M)][\bello_{\hat{p}_b,\hat{p}_m}\hat{Q}^{(k-1)}(S,A,M)-\bello\hat{Q}^{(k-1)}(S,A,M)]\\
    =2\gamma\E_{\substack{S\sim \rho^{\cD},A\sim p_b\\ M\sim p_m}} [\hat{Q}^{(k)}(S,A,M)-Q^{(k)*}(S,A,M)]\Big[\max_{a} \sum_{\tilde{a},m} \hat{p}_m(m|S',a) \hat{p}_b(\tilde{a}|S')\hat{Q}^{(k-1)}(S',\tilde{a},m)
    \\-\max_{a} \sum_{\tilde{a},m} p_m(m|S',a) p_b(\tilde{a}|S')\hat{Q}^{(k-1)}(S',\tilde{a},m)\Big]
    \le 2\gamma\E_{\substack{S\sim \rho^{\cD},A\sim p_b\\ M\sim p_m}} [\hat{Q}^{(k)}(S,A,M)-Q^{(k)*}(S,A,M)]\\
    \times V_{\max}\max_a \sum_{\tilde{a},m}|\hat{p}_m(m|S',a)\hat{p}_b(\tilde{a}|S')-p_m(m|S',a)p_b(\tilde{a}|S')|\\
    \le     \frac{\|\widehat{Q}^{(k)}-Q^{(k)*}\|^2_{\rho^{\cD}\times p_b\times p_m}}{6}
    +6\gamma^2V_{\max}^2\E_{S'\sim \rho^\cD}\max_a \Big[\sum_{\tilde{a},m}|\hat{p}_m(m|S',a)\hat{p}_b(\tilde{a}|S')-p_m(m|S',a)p_b(\tilde{a}|S')|\Big]^2,
\end{eqnarray*}
where the last inequality is due to the stationarity assumption which requires $S'$ to have the same distribution function as $S$, and Cauchy-Schwarz inequality. 

Similar to \eqref{eqn:variancebound}, the first term in the last line can be upper bounded by $\varepsilon^{(k)}/3+2\varepsilon_{\cQ}^2/3$. Meanwhile, using similar arguments in the proof of Lemma \ref{lem:totalvar}, the second term in the last line can be upper bounded by
\begin{eqnarray*}
    &&6\gamma^2V_{\max}^2\E_{S\sim \rho^{\cD}}\max_a \Big[ 2\textrm{TV}(p_b(\bullet|S)\|\hat{p}_b(\bullet|S))+2\textrm{TV}(p_m(\bullet|S,a)\|\hat{p}_m(\bullet|S,a)) \Big]^2\\
    &\le& 12\gamma^2V_{\max}^2\E_{S\sim \rho^{\cD}}\max_a \Big[ \sqrt{\textrm{KL}(p_b(\bullet|S)\|\hat{p}_b(\bullet|S))}+\sqrt{\textrm{KL}(p_m(\bullet|S,a)\|\hat{p}_m(\bullet|S,a))} \Big]^2\\
    &\le& 24\gamma^2V_{\max}^2\E_{S\sim \rho^{\cD}} \max_a \Big[\textrm{KL}(p_b(\bullet|S)\|\hat{p}_b(\bullet|S))+\textrm{KL}(p_m(\bullet|S,a)\|\hat{p}_m(\bullet|S,a))\Big],
\end{eqnarray*}
where the last inequality follows from the Cauchy-Schwarz inequality. Under the coverage assumption, it can be further upper bounded by
\begin{eqnarray*}
    24\gamma^2V_{\max}^2\E_{S\sim \rho^{\cD},A\sim p_b} \Big[\textrm{KL}(p_b(\bullet|S)\|\hat{p}_b(\bullet|S))+c_2 \textrm{KL}(p_m(\bullet|S,A)\|\hat{p}_m(\bullet|S,A))\Big].
\end{eqnarray*}
This together with \eqref{eqn:varepsilon1k} yields that
\begin{eqnarray*}
    \varepsilon^{(k)} &\le& 10\varepsilon_{\cQ}^2+ 16N^{-1}\left[(97\lambda-48)V_{\max}^2\ln(2\delta^{-1}|\cQ|^2|\cP_b||\cP_m|)+12\lambda V_{\max}^2\right] \\
    && +144\gamma^2 V_{\max}^2 \E_{S\sim \rho^{\cD},A\sim p_b}\Big[\textrm{KL}(p_b(\bullet|S)\|\hat{p}_b(\bullet|S)) +c_2 \textrm{KL}(p_m(\bullet|S,A)\|\hat{p}_m(\bullet|S,A))\Big].
\end{eqnarray*}
The proof of Lemma \ref{lem:errorbound} is hence completed. 
\end{proof}
Combined with \eqref{eqn:bound1}, we obtain that
\begin{eqnarray*}
    &&\max\Big(\|Q^*-\hat{Q}^{(K)}\|_{\rho^{\hat{\pi}_K}\times p_b\times p_m^{\pi^*}}, \|Q^*-\hat{Q}^{(K)}\|_{\rho^{\hat{\pi}_K}\times p_b\times p_m^{\hat{\pi}^{(K)}}}\Big)\le \sqrt{c_1c_4}\gamma^K V_{\max}\\
    &+&\frac{\sqrt{c_1c_4}}{1-\gamma}\Big[4\varepsilon_{\cQ}+\frac{4V_{\max}}{\sqrt{N}}\sqrt{12\lambda+(97\lambda-48)\ln(2\delta^{-1}|\cQ|^2|\cP_b||\cP_m|)}\\
    &+&12\gamma V_{\max} \Big(\sqrt{\E_{S\sim \rho^{\cD},A\sim p_b}\textrm{KL}(p_b(\bullet|S)\|\hat{p}_b(\bullet|S))}+\sqrt{c_2^{-1}\E_{S\sim \rho^{\cD},A\sim p_b}\textrm{KL}(p_m(\bullet|S,A)\|\hat{p}_m(\bullet|S,A))}\Big)\Big],
\end{eqnarray*}
with probability at least $1-\left(\delta+\floor{\frac{N}{\lambda}}\beta(\lambda)\right)$. 

\noindent\textbf{\textit{Step 4.}} Based on the results in Steps 2 \& 3, we obtain with probability at least $1-\left(\delta+\floor{\frac{N}{\lambda}}\beta(\lambda)\right)$ that
\begin{eqnarray*}
    J(\pi^*)-J(\hat{\pi}_K)\le \frac{2\sqrt{c_1c_4}\gamma^K V_{\max}}{1-\gamma}+\frac{2\sqrt{c_1c_4}}{(1-\gamma)^2}\Big[4\varepsilon_{\cQ}+\frac{4V_{\max}}{\sqrt{N}}\sqrt{12\lambda+(97\lambda-48)\ln(2\delta^{-1}|\cQ|^2|\cP_b||\cP_m|)}\\
    +12\gamma V_{\max} \Big(\sqrt{\E_{S\sim \rho^{\cD}}\textrm{KL}(p_b(\bullet|S)\|\hat{p}_b(\bullet|S))}+\sqrt{c_2\E_{S\sim \rho^{\cD},A\sim p_b}\textrm{KL}(p_m(\bullet|S,A)\|\hat{p}_m(\bullet|S,A))}\Big)\Big]\\
    +\frac{2V_{\max}}{1-\gamma} \E_{S\sim \rho^{\hat{\pi}_K}} \Big[\sqrt{2\textrm{KL}(p_b(\bullet|S)\|\hat{p}_b(\bullet|S))}+\sqrt{0.5\textrm{KL}(p_m(\bullet|S,\pi^*(S))\|\hat{p}_m(\bullet|S,\pi^*(S)))}\\
    +\sqrt{0.5\textrm{KL}(p_m(\bullet|S,\hat{\pi}_K(S))\|\hat{p}_m(\bullet|S,\hat{\pi}_K(S)))}\Big],
\end{eqnarray*}
Under the coverage assumption, it follows from the Cauchy-Schwarz inequality that the last two lines can be upper bounded by 
\begin{eqnarray*}
    \frac{2V_{\max}}{1-\gamma}  \Big[\sqrt{2c_1\E_{S\sim \rho^{\cD}}\textrm{KL}(p_b(\bullet|S)\|\hat{p}_b(\bullet|S))}+\sqrt{0.5c_1c_2\E_{S\sim \rho^{\cD},A\sim p_b}\textrm{KL}(p_m(\bullet|S,A)\|\hat{p}_m(\bullet|S,A))}\\
    +\sqrt{0.5c_1c_2\E_{S\sim \rho^{\cD},A\sim p_b}\textrm{KL}(p_m(\bullet|S,A)\|\hat{p}_m(\bullet|S,A))}\Big].
\end{eqnarray*}
Notice that $c_1,c_4\ge 1$, we obtain that
\begin{eqnarray}\label{eqn:finalstep}
\begin{split}
    J(\pi^*)-J(\hat{\pi}_K)\le \frac{2\sqrt{c_1c_4}\gamma^K V_{\max}}{1-\gamma}+\frac{2\sqrt{c_1c_4}}{(1-\gamma)^2}\Big[4\varepsilon_{\cQ}+\frac{4V_{\max}}{\sqrt{N}}\sqrt{12\lambda+(97\lambda-48)\ln(2\delta^{-1}|\cQ|^2|\cP_b||\cP_m|)}\\
    +O(1) V_{\max} \Big(\sqrt{\E_{S\sim \rho^{\cD}}\textrm{KL}(p_b(\bullet|S)\|\hat{p}_b(\bullet|S))}+\sqrt{c_2\E_{S\sim \rho^{\cD},A\sim p_b}\textrm{KL}(p_m(\bullet|S,A)\|\hat{p}_m(\bullet|S,A))}\Big)\Big],
\end{split}
\end{eqnarray}
for some constant $O(1)>0$, with probability at least $1-\left(\delta+\floor{\frac{N}{\lambda}}\beta(\lambda)\right)$. According to Assumption \ref{assump:mixing}, $\beta(\lambda)$ converges to 0 with an exponential rate. Therefore, we can denote it as $1-\delta$ with a a non-degenerate probability. Let $\mathcal{I}$ denote such an event. The probability that $\mathcal{I}$ does not occur is upper bounded by $\delta$. Consequently, for any $\delta>0$, 
\begin{eqnarray*}
    \E[J(\pi^*)-J(\hat{\pi}_K)]=\E[J(\pi^*)-J(\hat{\pi}_K)]\mathbb{I}(\mathcal{I})+\E[J(\pi^*)-J(\hat{\pi}_K)]\mathbb{I}(\mathcal{I}^c).
\end{eqnarray*}
The first term on the RHS is upper bounded by the RHS of \eqref{eqn:finalstep} whereas the second term is upper bounded by $\delta V_{\max}$ by definition. Therefore, we finished proving Theorem \ref{thm:thm2full}. The proof of Theorem \ref{thm:converge} presented in the main text is thus completed by setting $\delta$ to $N^{-1}$ and discarding negative terms.

\section{Proof of Theorem \ref{thm:pess/pessm}}\label{proof:pessim}

\begin{proof}
The proof of Theorem \ref{thm:pess/pessm} is based on that of Theorem \ref{thm:converge}. Let $\widetilde{Q}^*$
denote a transformed version of $Q^*$, given by $\widetilde{Q}^*(s,a,m)=Q^*(s,a,m)-\inf_{(s',a',m')}\widehat{Q}(s',a',m')$. Similarly, let $\widetilde{q}^*(s,a)$ denote the corresponding unmediated Q-function associated with $\widetilde{Q}^*(s,a,m)$. According to the performance difference lemma \citep{kakade2002approximately}, using similar arguments in the proof of Theorem \ref{thm:converge}, we obtain that
\begin{eqnarray}\label{eqn:proofthm31}
    \E [J(\pi^*)-J(\widetilde{\pi})]&\le&\frac{1}{1-\gamma}\E\E_{S\sim \rho^{\widetilde{\pi}}} \Big[\widetilde{q}^*(S,\pi^*(S))-\widetilde{q}(S,\pi^*(S))\Big]\\\label{eqn:proofthm32}
    &+&\frac{1}{1-\gamma}\E\E_{S\sim \rho^{\widetilde{\pi}}}\Big[\widetilde{q}(S,\widetilde{\pi}(S))-\widetilde{q}^*(S,\widetilde{\pi}(S)) \Big],
\end{eqnarray}
where
\begin{eqnarray*}
    \widetilde{q}(s,a)=\sum_{\tilde{a},m} \{\widehat{p}_m(m|s,a)-\Delta(s,a,m)\}\widehat{p}_b(\tilde{a}|s)\tilde{Q}(s,\tilde{a},m).
\end{eqnarray*}
In the following, we bound \eqref{eqn:proofthm31} and \eqref{eqn:proofthm32} separately. 

Consider \eqref{eqn:proofthm31} first. By definition, \eqref{eqn:proofthm31} equals
\begin{eqnarray*}
    \frac{1}{1-\gamma}\E\E_{S\sim \rho^{\widetilde{\pi}}} \sum_{\tilde{a},m}\Big[p(m|S,\pi^*(S))p_b(\tilde{a}|S)Q^*(S,\tilde{a},m)- \widehat{p}_m(m|S,\pi^*(S))\widehat{p}_b(\tilde{a}|S)\widehat{Q}(S,\tilde{a},m) \Big]\\
    +\frac{\E\E_{S\sim \rho^{\widetilde{\pi}}}}{1-\gamma} \sum_{\tilde{a},m}\Delta(S,\pi^*(S),m)\widehat{p}_b(\tilde{a}|S)\widehat{Q}(S,\tilde{a},m) \\
    +\frac{\E\E_{S\sim \rho^{\widetilde{\pi}}}}{1-\gamma}\inf \widehat{Q}\sum_{\tilde{a},m}\Big[\{\widehat{p}_m(m|S,\pi^*(S))-\Delta(S,\pi^*(S),m)\}\widehat{p}_b(\tilde{a}|S) -p(m|S,\pi^*(S))p_b(\tilde{a}|S)\Big].
\end{eqnarray*}
Notice that the last line equals $-(1-\gamma)^{-1}\E\E_{S\sim \rho^{\widetilde{\pi}}} \sum_m\Delta(S,\pi^*(S),m) \inf\widehat{Q}$ which is upper bounded by $(1-\gamma)^{-1}V_{\max} \sum_m \sqrt{c_1c_3\E\E_{S\sim \rho^{\mathcal{D}}, A\sim p_b}\Delta^2(S,A,m)}$ under Assumptions \ref{assump:coverage}(i), (iii). Similarly, the second line can be upper bounded by $\frac{V_{\max}}{1-\gamma} \sum_m \sqrt{c_1c_3\E\E_{S\sim \rho^{\mathcal{D}}, A\sim p_b}\Delta^2(S,A,m)}$ as well. Next, consider the first line. It can be decomposed into the sum of the following two terms,
\begin{eqnarray*}
    &&\frac{\E\E_{S\sim \rho^{\widetilde{\pi}}}}{1-\gamma} \sum_{\tilde{a},m} p(m|S,\pi^*(S))p_b(\tilde{a}|S)[Q^*(S,\tilde{a},m)-\widehat{Q}(S,\tilde{a},m)]\\
    &+&\frac{\E\E_{S\sim \rho^{\widetilde{\pi}}}}{1-\gamma} \sum_{\tilde{a},m} \widehat{Q}(S,\tilde{a},m)[p(m|S,\pi^*(S))p_b(\tilde{a}|S)-\hat{p}(m|S,\pi^*(S))\hat{p}_b(\tilde{a}|S)].
\end{eqnarray*}
Using similar arguments in Step 2 of the proof of Theorem \ref{thm:converge}, we can show that they can be upper bounded by
\begin{eqnarray}\label{eqn:proofthm33}
\begin{split}
    \frac{1}{1-\gamma}\sqrt{\E\|Q^*-\hat{Q}\|^2_{\rho^{\hat{\pi}}\times p_b\times p_m^{\pi^*}}}+\frac{V_{\max}}{1-\gamma}\Big[\sqrt{2\E\E_{S\sim \rho^{\tilde{\pi}}}\textrm{KL}(p_b(\bullet|S)\|\hat{p}_b(\bullet|S))}\\+2\E\E_{S\sim \rho^{\tilde{\pi}}}\textrm{TV}(p_m(\bullet|S,\pi^*(S))\|\hat{p}_m(\bullet|S,\pi^*(S)))\Big].
\end{split}
\end{eqnarray}
Let $\mathcal{E}(s,a,m)$ denote the event that $|p_m(m|s,a)-\hat{p}_m(m|s,a)|\le \Delta(s,a,m)$. Under Assumptions \ref{asump:uncertainty} and \ref{assump:coverage}, the expected total variation distance can be upper bounded by
\begin{eqnarray*}
    &&\E\E_{S\sim \rho^{\tilde{\pi}}}\textrm{TV}(p_m(\bullet|S,\pi^*(S))\|\hat{p}_m(\bullet|S,\pi^*(S)))\\
    &=&\frac{1}{2}\E\E_{S\sim \rho^{\tilde{\pi}}}\sum_m |p_m(m|S,\pi^*(S))-\hat{p}_m(m|S,\pi^*(S))|\mathbb{I}(\mathcal{E}(S,\pi^*(S),m))\\
    &+&\frac{1}{2}\E\E_{S\sim \rho^{\tilde{\pi}}}\sum_m |p_m(m|S,\pi^*(S))-\hat{p}_m(m|S,\pi^*(S))|\mathbb{I}(\mathcal{E}^c(S,\pi^*(S),m))\\
    &\le&\frac{1}{2}\sum_m \E\E_{S\sim \rho^{\tilde{\pi}}} \Delta(S,\pi^*(S),m)+\frac{\alpha}{2}|\mathcal{M}|\le \frac{1}{2}\sum_m \sqrt{c_1c_3\E\E_{S\sim \rho^{\mathcal{D}},A\sim p_b} \Delta^2(S,A,m)}+\frac{\alpha}{2}|\mathcal{M}|.
\end{eqnarray*}
Consequently, \eqref{eqn:proofthm33} is upper bounded by
\begin{eqnarray*}
    \frac{1}{1-\gamma}\sqrt{c_1c_4\E\|Q^*-\hat{Q}\|^2_{\rho^{\mathcal{D}}\times p_b\times p_m}}+\frac{V_{\max}}{1-\gamma}\sqrt{2c_1\E\E_{S\sim \rho^{\mathcal{D}}}\textrm{KL}(p_b(\bullet|S)\|\hat{p}_b(\bullet|S))}\\
    +\frac{V_{\max}}{1-\gamma}\sum_m \sqrt{c_1c_3\E\E_{S\sim \rho^{\mathcal{D}},A\sim p_b} \Delta^2(S,A,m)}+\frac{\alpha V_{\max}}{1-\gamma}|\mathcal{M}|. 
\end{eqnarray*}
To summarize, \eqref{eqn:proofthm31} is upper bounded by
\begin{eqnarray}\label{eqn:proofthm34}
\begin{split}
    \frac{1}{1-\gamma}\sqrt{c_1c_4\E\|Q^*-\hat{Q}\|^2_{\rho^{\mathcal{D}}\times p_b\times p_m}}+\frac{V_{\max}}{1-\gamma}\sqrt{2c_1\E\E_{S\sim \rho^{\mathcal{D}}}\textrm{KL}(p_b(\bullet|S)\|\hat{p}_b(\bullet|S))}\\
    +\frac{3V_{\max}}{1-\gamma}\sum_m \sqrt{c_1c_3\E\E_{S\sim \rho^{\mathcal{D}},A\sim p_b} \Delta^2(S,A,m)}+\frac{\alpha V_{\max}}{1-\gamma}|\mathcal{M}|. 
\end{split}
\end{eqnarray}
We next consider \eqref{eqn:proofthm32}. By definition, it can be decomposed into the following,
\begin{eqnarray}\label{eqn:proofthm3twolines}
\begin{split}
    &&\frac{\E\E_{S\sim \rho^{\widetilde{\pi}}}}{1-\gamma}\sum_{\tilde{a},m}\Big[ p_m(m|S,\tilde{\pi}(S))\{\widehat{p}_b(\tilde{a}|S)\tilde{Q}(S,\tilde{a},m)-p_b(\tilde{a}|S)\tilde{Q}^{*}(S,\tilde{a},m)\} \Big]\\
    &+&\frac{\E\E_{S\sim \rho^{\widetilde{\pi}}}}{1-\gamma}\sum_{\tilde{a},m}\Big[ \{\hat{p}_m(m|S,\tilde{\pi}(S))-\Delta(S,\tilde{\pi}(S),m)-p_m(m|S,\tilde{\pi}(S))\}\hat{p}_b(\tilde{a}|S) \tilde{Q}(S,\tilde{a},m) \Big].
\end{split}
\end{eqnarray}
The first line in \eqref{eqn:proofthm3twolines} equals 
\begin{eqnarray*}
    \frac{\E\E_{S\sim \rho^{\widetilde{\pi}}}}{1-\gamma}\sum_{\tilde{a},m}\Big[ p_m(m|S,\tilde{\pi}_K(S)\{\widehat{p}_b(\tilde{a}|S)\hat{Q}(S,\tilde{a},m)-p_b(\tilde{a}|S)\hat{Q}^{*}(S,\tilde{a},m)\} \Big].
\end{eqnarray*}
Using similar arguments in bounding \eqref{eqn:proofthm31}, we obtain that it can be upper bounded by
\begin{eqnarray*}
    \frac{1}{1-\gamma}\sqrt{c_1c_4\E\|Q^*-\hat{Q}\|^2_{\rho^{\mathcal{D}}\times p_b\times p_m}}+\frac{V_{\max}}{1-\gamma}\sqrt{2c_1\E\E_{S\sim \rho^{\mathcal{D}}}\textrm{KL}(p_b(\bullet|S)\|\hat{p}_b(\bullet|S))}.
\end{eqnarray*}
Now consider the second line in \eqref{eqn:proofthm3twolines}. It can be further decomposed into the sum of 
\begin{eqnarray}\label{eqn:proofthm35}
\begin{split}
    \frac{\E\E_{S\sim \rho^{\widetilde{\pi}}}}{1-\gamma}\sum_{\tilde{a},m}\Big[ \{\hat{p}_m(m|S,\tilde{\pi}(S))-\Delta(S,\tilde{\pi}(S),m)-p_m(m|S,\tilde{\pi}(S))\}\hat{p}_b(\tilde{a}|S) \tilde{Q}(S,\tilde{a},m) \Big]\\
    \times \mathbb{I}(\mathcal{E}(S,\tilde{\pi}(S),m))
\end{split}
\end{eqnarray}
and
\begin{eqnarray}\label{eqn:proofthm36}
\begin{split}
    \frac{\E\E_{S\sim \rho^{\widetilde{\pi}}}}{1-\gamma}\sum_{\tilde{a},m}\Big[ \{\hat{p}_m(m|S,\tilde{\pi}(S))-\Delta(S,\tilde{\pi}(S),m)-p_m(m|S,\tilde{\pi}(S))\}\hat{p}_b(\tilde{a}|S) \tilde{Q}(S,\tilde{a},m) \Big]\\
    \times \mathbb{I}(\mathcal{E}^c(S,\tilde{\pi}(S),m)).
\end{split}
\end{eqnarray}
By the definition of $\mathcal{E}$ and the nonnegativeness of $\tilde{Q}$, \eqref{eqn:proofthm35} is nonpositive. As for \eqref{eqn:proofthm36}, with some calculations, we obtain that is can be upper bounded by
\begin{eqnarray*}
    &&\frac{\E\E_{S\sim \rho^{\widetilde{\pi}}}}{1-\gamma}\sum_{\tilde{a},m}\hat{p}_m(m|S,\tilde{\pi}(S))\hat{p}_b(\tilde{a}|S) \tilde{Q}(S,\tilde{a},m)\mathbb{I}(\mathcal{E}^c(S,\tilde{\pi}(S),m))\\
    &=&\frac{\E\E_{S\sim \rho^{\widetilde{\pi}}}}{1-\gamma}\sum_{\tilde{a},m}\hat{p}_m(m|S,\tilde{\pi}(S))\hat{p}_b(\tilde{a}|S) [\hat{Q}(S,\tilde{a},m)-\inf_{(s',a',m')}\hat{Q}(s',a',m')]\mathbb{I}(\mathcal{E}^c(S,\tilde{\pi}(S),m))\\
    &\le&2V_{\max} \frac{\E\E_{S\sim \rho^{\widetilde{\pi}}}}{1-\gamma}\sum_{m}\hat{p}_m(m|S,\tilde{\pi}(S))\mathbb{I}(\mathcal{E}^c(S,\tilde{\pi}(S),m))\\
    &\le&2V_{\max} \frac{\E\E_{S\sim \rho^{\widetilde{\pi}}}}{1-\gamma}\sum_{a,m}\hat{p}_m(m|S,a)\mathbb{I}(\mathcal{E}^c(S,a,m))\le \frac{2\alpha V_{\max}|\mathcal{A}|}{1-\gamma}.
\end{eqnarray*}
Consequently, \eqref{eqn:proofthm32} is upper bounded by
\begin{eqnarray*}
    \frac{1}{1-\gamma}\sqrt{c_1c_4\E\|Q^*-\hat{Q}\|^2_{\rho^{\mathcal{D}}\times p_b\times p_m}}+\frac{V_{\max}}{1-\gamma}\sqrt{2c_1\E\E_{S\sim \rho^{\mathcal{D}}}\textrm{KL}(p_b(\bullet|S)\|\hat{p}_b(\bullet|S))}+\frac{2\alpha V_{\max}|\mathcal{A}|}{1-\gamma}.
\end{eqnarray*}
This together with \eqref{eqn:proofthm34} yields the desired result in Theorem \ref{thm:pess/pessm}, omitting constant terms. 
\end{proof}

\end{document}